\documentclass[manuscript,screen]{acmart}

\usepackage{pifont}
\usepackage{enumitem}
\usepackage{wrapfig}
\usepackage{threeparttable}
\usepackage{multirow}

\newcommand{\cmark}{\ding{51}}%
\newcommand{\xmark}{\ding{55}}%

\AtBeginDocument{%
  }

\setcopyright{acmlicensed}
\copyrightyear{2024}
\acmYear{2024}
\acmDOI{XXXXXXX.XXXXXXX}

\acmConference[Conference acronym 'XX]{Make sure to enter the correct
  conference title from your rights confirmation emai}{June 03--05,
  2018}{Woodstock, NY}
\acmISBN{978-1-4503-XXXX-X/18/06}

\begin{document}

\title{Dive into Time-Series Anomaly Detection: A Decade Review}


\author{Paul Boniol}
\affiliation{%
  \institution{Inria, DI ENS, PSL, CNRS}
  \city{Paris}
  \country{France}}
\email{paul.boniol@inria.fr}

\author{Qinghua Liu}
\affiliation{%
  \institution{The Ohio State University}
  \city{Columbus}
  \country{USA}}
\email{liu.11085@osu.edu}

\author{Mingyi Huang}
\affiliation{%
  \institution{The Ohio State University}
  \city{Columbus}
  \country{USA}}
\email{huang.5749@osu.edu}

\author{Themis Palpanas}
\affiliation{%
  \institution{Université Paris Cité; IUF}
  \city{Paris}
  \country{France}}
\email{themis@mi.parisdescartes.fr}

\author{John Paparrizos}
\affiliation{%
  \institution{The Ohio State University}
  \city{Columbus}
  \country{USA}}
\email{paparrizos.1@osu.edu}

\renewcommand{\shortauthors}{Paparrizos et al.}

\begin{abstract}
Recent advances in data collection technology, accompanied by the ever-rising volume and velocity of streaming data, underscore the vital need for time series analytics. In this regard, time-series anomaly detection has been an important activity, entailing various applications in fields such as cyber security, financial markets, law enforcement, and health care. While traditional literature on anomaly detection is centered on statistical measures, the increasing number of machine learning algorithms in recent years call for a structured, general characterization of the research methods for time-series anomaly detection. This survey groups and summarizes anomaly detection existing solutions under a process-centric taxonomy in the time series context. In addition to giving an original categorization of anomaly detection methods, we also perform a meta-analysis of the literature and outline general trends in time-series anomaly detection research.
\end{abstract}

\maketitle

\section{Introduction}

A wide range of cost-effective sensing, networking, storage, and processing solutions enable the collection of enormous amounts of measurements over time \cite{jeung2010effective,liu2023amir,paparrizos2021vergedb,jiang2020pids,jiang2021good,liu2021decomposed,paparrizos2018fastthesis,krishnan2019artificial,paparrizos2022fast,liu2024time,liu2024adaedge}. Recording these measurements results in an ordered sequence of real-valued data points commonly referred to as {\em time series}. More generic terms, such as {\em data series} or {\em data sequences}, have also been used to refer to cases where the ordering of data relies on a dimension other than time (e.g., the angle in data from astronomy, the mass in data from spectrometry, or the position in data from biology) \cite{Palpanas2015}. Analytical tasks over time series data are necessary virtually in every scientific discipline and their corresponding industries \cite{paparrizos2016screening,paparrizos2016detecting,paparrizos2023accelerating,paparrizos2019grail,dziedzic2019band,goel2016social,mckeown2016predicting,bariya2021k,qiu2024tfb,d2024beyond}, including in astronomy \cite{huijse2014computational,wachman2009kernels,alam2015eleventh}, biology \cite{bar2003continuous,ernst2006stem,bar2004analyzing,bar2012studying}, economics \cite{lutkepohl2004applied,tsay2014financial,brockwell2016introduction,shasha1999tuning, gavrilov2000mining, mantegna1999hierarchical, ruiz2012correlating}, energy sciences \cite{bach2017flexible,alvarez2010energy,martinez2015survey}, engineering \cite{uehara2002extraction,williams2003modeling,kashino1999time,mirylenka2016characterizing,raza2015practical}, environmental sciences \cite{goddard2003geospatial,webster2005changes,hoegh2007coral,morales2010pattern,rong2018locality,honda2002mining,grover2015deep}, medicine \cite{richman2000physiological,costa2002multiscale,peng1995quantification}, neuroscience \cite{biswal2010toward,knieling2017online}, and social sciences \cite{mccleary1980applied, brockwell2016introduction}. The analysis of time series has become increasingly prevalent for understanding a multitude of natural or human-made processes \cite{paparrizos2020debunking,paparrizosDEB23}. Unfortunately, inherent complexities in the data generation of these processes, combined with imperfections in the measurement systems as well as interactions with malicious actors, often result in abnormal phenomena. Such abnormal events appear subsequently in the collected data as anomalies. Considering that the volume of the produced time series will continue to rise due to the explosion of Internet-of-Things (IoT) applications \cite{mahdavinejad2017machine,iotstats,hung2017leading}, an abundance of anomalies is expected in time series collections.  

The detection of anomalies in time series has received ample academic and industrial attention for over six decades \cite{page1957problems,fox1972outliers,tsay1988outliers,Abdul-Aziz2010Propulsion,MorariuBorangiu2018Time,BuiEtAl2018Time,mason2002multivariate,krensky2018hype,boniol2023new,paparrizos2022tsb,paparrizos2022volume,msad-sylligardos23,boniol2022theseus,boniol2021sandinaction,liu2024elephant,boniol2024adecimo,boniol2024interactive}. With the term {\em anomalies} we refer to data points or groups of data points that do not conform to some notion of normality or an expected behavior based on previously observed data \cite{barnett1984outliers,hawkins_identification_1980,goldstein2016comparative,izenman2008modern,chandola2009anomaly}. In the literature, alternative terms such as outliers, novelties, exceptions, peculiarities, aberrations, deviants, or discords often appear to describe the occurrences of such rare, unusual, often hard-to-explain anomalous patterns \cite{carreno2020analyzing,aggarwal2017introduction,esling2012time}. Depending on the application, anomalies can constitute \cite{aggarwal2017introduction}: (i) noise or erroneous data, which hinders the data analysis; or (ii) actual data of interest. In the former case, the anomalies are unwanted data that are removed or corrected. In the latter case, the anomalies may identify meaningful events, such as failures or changes in behavior, which are the basis for subsequent analysis.

Regardless of the purpose of the time series and the semantic meaning of anomalies, {\em anomaly detection} describes the process of analyzing a time series for identifying unusual patterns, which is a challenging task because many types of anomalies exist. They appear in different sizes and shapes. According to Foorthuis~\cite{foorthuis2020nature}, research on general-purpose anomaly detection dates back to 1777, where Bernoulli's work seems to be the first addressing issues of accepting or rejecting extreme cases of observations \cite{bernoulli1961most}. Robust theory in that area was developed during the 1800s (e.g., method of least squares in 1805 \cite{smith2012source}) \cite{peirce1852criterion,glaisher1873rejection,edgeworth1887xli,stone1873rejection,gould1855peirce} and 1900s \cite{irwin1925criterion,pearson1936efficiency,grubbs1950sample,dixon1950analysis,rousseeuw2005robust} but it was not until the 1950s when the first works focused specifically in time series data \cite{page1957problems}. In 1972, Fox conducted one of the first studies to examine anomalous behaviors across time and defined two types of anomalies: (i) an anomaly affecting a single observation; and (ii) an anomaly affecting an observation and subsequent observations \cite{fox1972outliers}. In 1988, Tsay extended these definitions into four types for univariate time series \cite{tsay1988outliers} and subsequently for multivariate time series \cite{tsay2000outliers}. In the same time frame, the first few approaches appear for detecting anomalies in time series, with a focus on utilizing statistical tests such as the Likelihood-ratio test \cite{tukey1977exploratory,chang1988estimation}

\begin{figure}[t]
 \centering
 \includegraphics[scale=0.87]{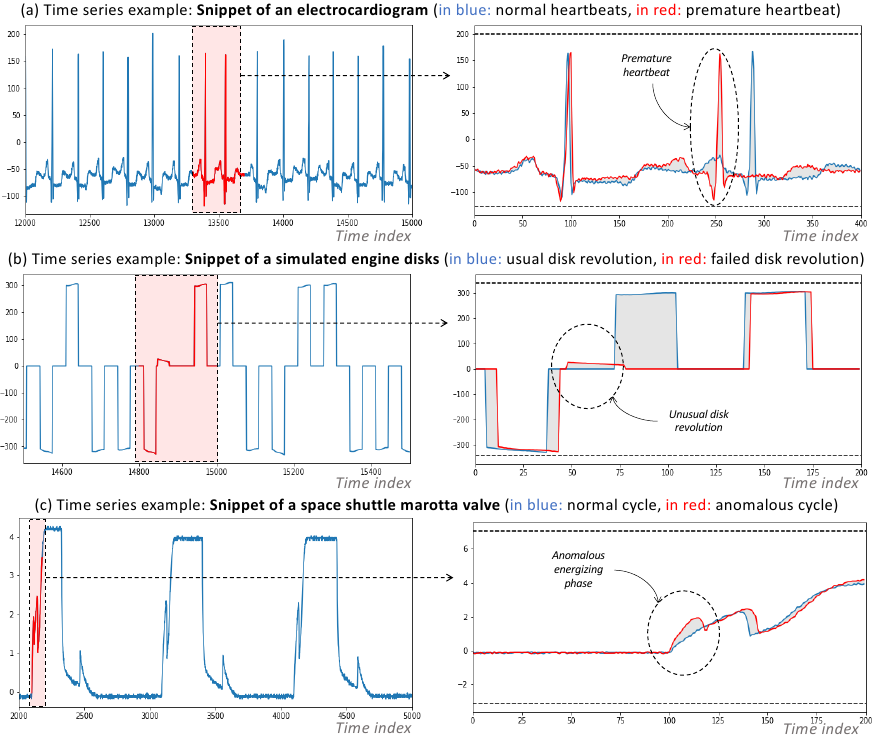}
 \caption{Examples of different time series applications and types of anomalies.}
 \label{fig:introexamples}
\end{figure}

Since then, a large number of works have appeared in this area, which continues to expand at a rapid pace. Additionally, numerous surveys have been published to provide an overview of the current advancements in this field
\cite{Blazquez-GarciaEtAl2020review,BraeiWagner2020Anomaly,ChalapathyChawla2019Deep,ChandolaEtAl2012Anomaly,ChandolaEtAl2009Anomaly,ChoudharyEtAl2017RuntimeEfficacy,CookEtAl2020Anomaly,GuptaEtAl2014Outlier,HodgeAustin2004Survey}. Unfortunately, the majority of the surveys focus on general-purpose anomaly detection methods and only a portion of them briefly review methods for time-series anomaly detection. Even though traditional anomaly detection methods may treat time series as any other high-dimensional vector and attempt to detect anomalies, our focus is on approaches that are specifically designed to consider characteristics of time series. To illustrate the importance of this point, in Figure \ref{fig:introexamples}, we present three examples of anomalies in time series applications where the temporal ordering and the collective consideration of points enable the detection of anomalies. For example, in \ref{fig:introexamples}(a), considering each point in isolation cannot reveal the underlying anomaly in data. 

Therefore, time-ordering features are important to be considered in the anomaly detection pipeline. Depending on the research community, multiple solutions have been proposed to tackle the above-mentioned challenge.
For example, a group of methods proposed in the literature will propose a transformation step that converts time information into a relevant vector space and then apply traditional outlier detection techniques. In addition, other groups of methods will consider distances (or similarity measures dedicated to time series) to identify unusual time series or subsequences. Then, methods proposed in the deep learning community will benefit from specific architectures that embed time information (such as recurrent neural networks or convolutional-based approaches).
 
Unfortunately, these areas remain mostly disconnected, using different datasets, baselines, and evaluation measures. 
New algorithms are evaluated only against non-representative subsets of approaches and it is virtually impossible to find the best state-of-the-art approach for a concrete use case. To remedy this issue, this survey presents a novel, comprehensive, process-centric taxonomy for the multiple detection methods in each category. We collected a comprehensive range of algorithms in the literature and grouped them into families of algorithms with similar approaches. 
In addition, to identify research trends, we also provide statistics on the type and area of proposed approaches over time.

Then, we also list existing benchmarks that can be used as a common ground on which new proposed methods (regardless of the community) should be evaluated. Finally, we enumerate existing evaluation measures usually used to evaluate anomaly detection methods while discussing their limitation and benefits when applied to the specific case of time series.

\section{Time-Series Anomaly Detection Overview}
\label{sec:overview}
In this section, we discuss the problem formulation for time-series anomaly detection algorithms and motivate our process-centric taxonomy. 

\subsection{On the Definition of Anomalies in Time Series}

Attesting to the challenging nature of the problem at hand, we observe that there does not exist a single, universal, precise definition of anomalies or outliers. Traditionally, anomalies are observations that deviate considerably from the majority of other samples in a distribution. The anomalous points raise suspicions that a mechanism different from those of the other data generated the specific observation. Such a mechanism usually represents either an erroneous data measurement procedure or an unusual event.

In cases of errors in the data measurement procedure, the anomalous observations are marked as ``noise" -- unwanted data that are not attractive to the analyst and should be removed in the data cleaning process. Many pieces of literature have been dedicated to this type of problem, particularly in the sensor setting, where the main objective is to eliminate transmission error and render accurate predictions. In time-series anomaly detection, however, recent literature begins to center on detecting anomalous events, which indicate ``novelty" -- unusual but interesting phenomena that originate from an inherent variability in the domain of the data. A natural example of this type of problem is fraud detection for credit information, where the principal aim is to detect and analyze the fraud itself. 

The detection of these two types of anomalies (anomalies and outliers can be used interchangeably) can be achieved via expert knowledge. By knowing exactly how the system works, the experts can set the parameter to fit a distribution of values that represent the healthy functioning state.  Anomalies are then detected by marking points of more than three standard deviations away from the mean of data distribution estimated by the experts. To validate the detection process, we also need to perform extensive tests to test the distribution (and its parameters) against the dataset.

Nevertheless, in several real-world problems, we do not know precisely the data distribution that has generated a set of points (and all the different artifacts that played a role). Besides, the data distributions that we encounter in practice, are almost always rather complex and very hard to identify or approximate effectively. Consequently, defining and identifying anomalies using their distance from a mean value defined by experts is sometimes hardly practical.

Despite the challenge of estimating distribution parameters by experts, recent developments in computational power have liberated us from an alternative approach to analyzing data distribution from the data itself. Using a variety of learning methods, researchers may apply computer algorithms to analyze raw data, estimate a fair distribution, and detect anomalies without expert knowledge. Even though being strongly dependent on the quality and the context of the datasets, these methods seem to show strong results in achieving relatively complex tasks. In this paper, we focus on this type of method.


\subsection{Types of Anomalies in Time Series}

\begin{figure}[t]
 \centering
 \includegraphics[scale=0.535]{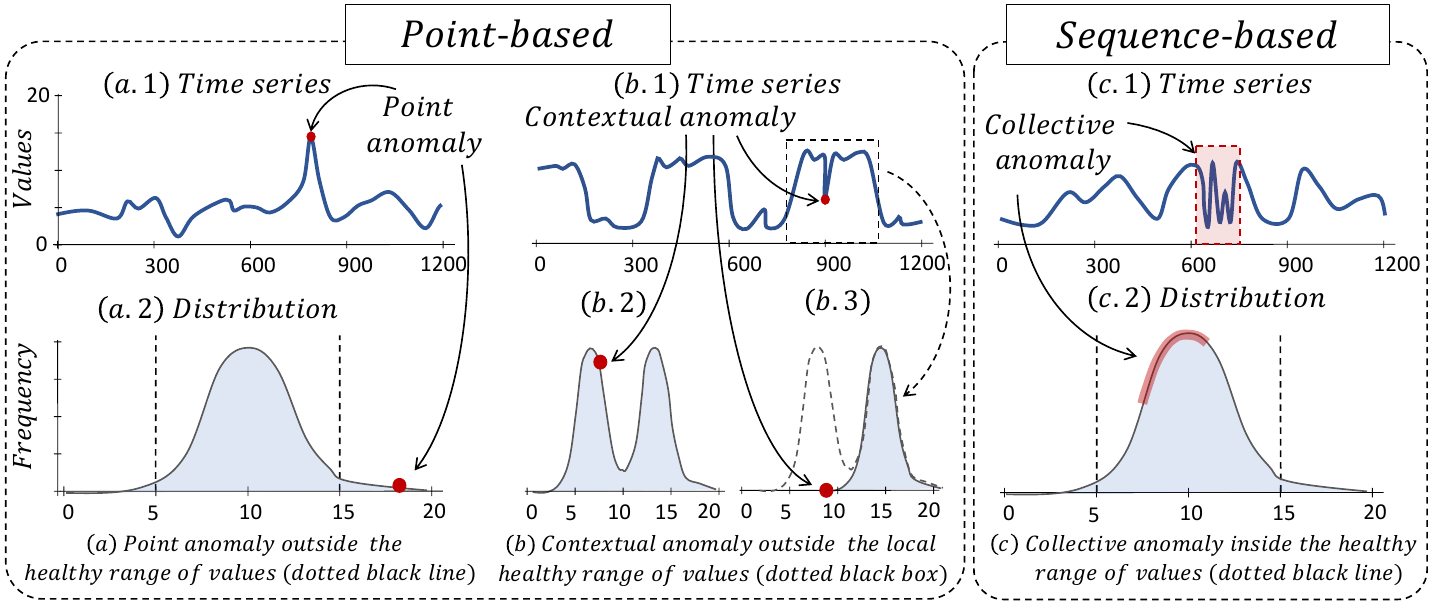}
 \caption{Synthetic illustration of the three time series anomaly types: (a) point; (b) contextual; and (c) collective anomalies.}
 \label{fig:anomaly_type}
\end{figure}

There is a further complication in time-series anomaly detection. Due to the temporality of the data, anomalies can occur in the form of a single value or collectively in the form of sub-sequences. In the specific context of point, we are interested in finding points that are far from the usual distribution of values that correspond to {\it healthy} states. In the specific context of sequences, we are interested in identifying anomalous sub-sequences, which are usually not outliers but exhibit rare and, hence, anomalous patterns. In real-world applications, such a distinction between points and sub-sequences becomes crucial because even though individual points might seem normal against their neighboring points, the shape generated by the sequence of these points may be anomalous. 

Formally, we define three types of time series anomalies: {\em point}, {\em contextual}, and {\em collective} anomalies. {\em Point} anomalies refer to data points that deviate remarkably from the rest of the data. Figure~\ref{fig:anomaly_type}(a) depicts a synthetic time series with a point anomaly: the value of the anomaly is outside the expected range of normal values. {\em Contextual} anomalies refer to data points within the expected range of the distribution (in contrast to point anomalies) but deviate from the expected data distribution, given a specific context (e.g., a window). Figure~\ref{fig:anomaly_type}(b) illustrates a time series with a contextual anomaly: the anomaly is within the usual range of values (left distribution plot of Figure~\ref{fig:anomaly_type}(b)) but outside the normal range of values for a local window (right distribution plot of Figure~\ref{fig:anomaly_type}(b)). {\em Collective} anomalies refer to sequences of points that do not repeat a typical (previously observed) pattern. Figure~\ref{fig:anomaly_type}(c) depicts a synthetic collective anomaly. The first two categories, namely, point and contextual anomalies, are referred to as {\em point-based} anomalies. whereas, {\em collective} anomalies are referred to as {\em sequence-based} anomalies.

As an additional note, there is another case of sub-sequence anomaly detection referred to as whole-sequence detection, relative to the point detection. In this case, the period of the sub-sequence is that of the entire time series, and the entire time series is evaluated for anomaly detection as a whole. This is typically the case in the sensor cleaning environment where researchers are interested in finding an abnormal sensor among all the functioning sensors.

\subsection{Univariate versus Multivariate}

\begin{figure}[t]
 \centering
 \includegraphics[scale=0.52]{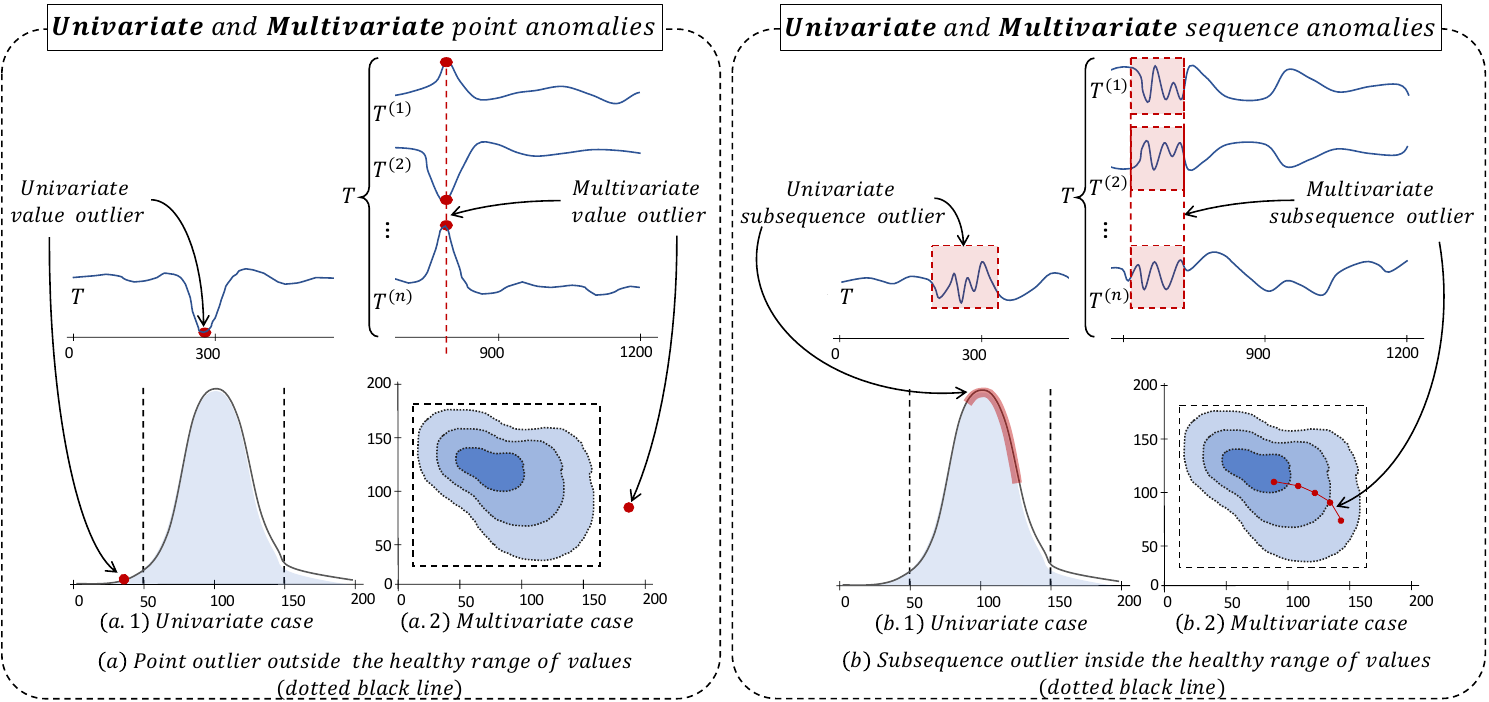}
 \caption{Synthetic example comparing anomalies in univariate and multivariate time series for (a) a point outlier and (b) a sequence outlier.}
 \label{fig:point_vs_sequence}
\end{figure}

Another characteristic of time-series anomaly detection algorithms comes from the dimensionality of the data. {\em Univariate} time series consists of an ordered sequence of real values on a single dimension, and the anomalies are detected based on one single feature. In this case, as illustrated in Figure~\ref{fig:point_vs_sequence}(b.1), a subsequence can be represented as a vector. 
On the other hand, {\em Multivariate} time series is either a set of ordered sequences of real values (with each ordered sequence having the same length) or an ordered sequence of vectors composed of real values. In this specific case, as illustrated in Figure~\ref{fig:point_vs_sequence}(b.2), a subsequence is a matrix in which each line corresponds to a subsequence of one single dimension. 
Instances of anomalies are detected according to multiple features, where values of one feature, when singled out, may look normal despite the abnormality of the sequence as a whole.

\subsection{Unsupervised, Semi-supervised versus Supervised}
This task can be divided into three distinct cases:
(i) experts do not have information on what anomalies to detect; 
(ii) experts only have information on the expected normal behaviors;
(iii) experts have precise examples of which anomalies they have to detect (and have a collection of known anomalies). This gives rise to the distinction among (i) unsupervised, (ii) semi-supervised, and (iii) supervised methods. 

\vspace{.2cm}
\noindent{\textit{Unsupervised:}}
In case (i), one can decide to adopt a fully unsupervised method. These methods have the benefit of working without the need for a large collection of known anomalies and can detect unknown abnormal behavior automatically. Such methods can be used either to monitor the health state or to mine the historical time series of a system (to build a collection of abnormal behaviors that can then be used on a supervised framework). 

\vspace{.2cm}
\noindent{\textit{Semi-supervised:}}
In case (ii), methods can learn to detect anomalies based on annotated examples of normal sequences provided by the experts. 
This is the classical case for most of the anomaly detection methods in the literature. 
One should note that this category is often defined as Unsupervised. However, we consider it unfair to group such methods with the category mentioned above, knowing that they require much more prior knowledge than the previous one.

\vspace{.2cm}
\noindent{\textit{Supervised:}}
While in case (i) and (ii) anomalies were considered unknown, in case (iii), we consider that the experts know precisely, what type of pattern(s) they want to detect, and that a collection of time series with labeled anomalies is available. In that case, we have a database of anomalies at our disposal. In a supervised setting, one may be interested in predicting the abnormal sub-sequence by its prior sub-sequences. Such sub-sequences can be called {\it precursors} of anomalies.

\subsection{Anomaly Detection Pipelines}

\begin{figure}[t]
 \centering
 \includegraphics[scale=0.60]{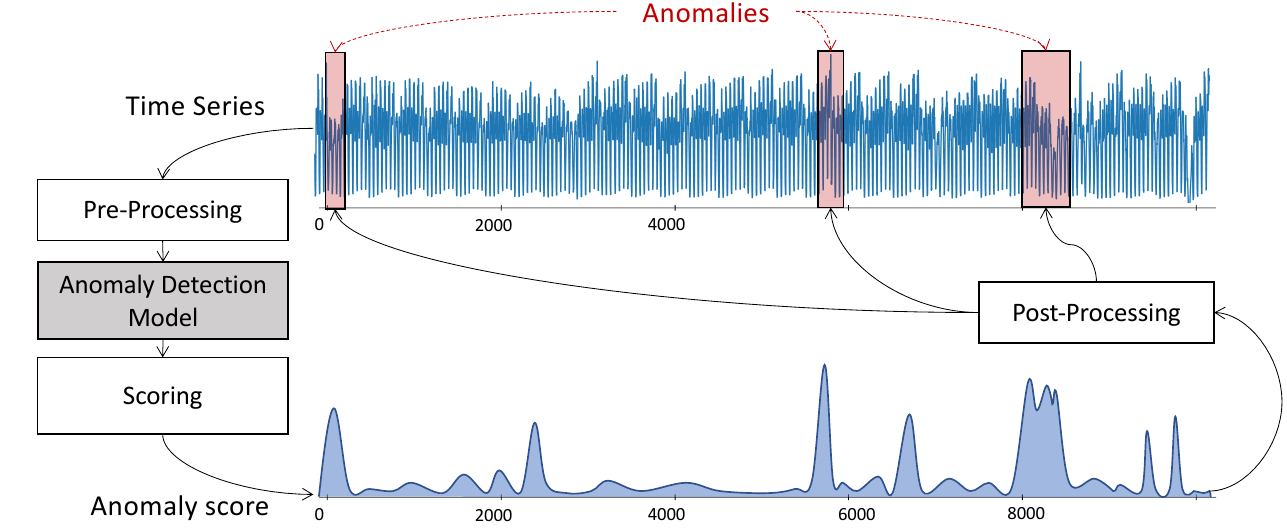}
 \caption{Time series anomaly detection pipeline.}
 \label{fig:pipeline1}
\end{figure}

Upon summarizing the various different algorithms on different domains, we realized a common pipeline for time-series anomaly detection algorithms. The pipeline consists of four parts: {\it data pre-processing}, {\it detection method}, {\it scoring}, and {\it post-processing}. Figure \ref{fig:pipeline1} illustrates the process. The decomposition of the general anomaly detection process into small steps of a pipeline is beneficial for comparing different algorithms on various dimensions. Understanding of algorithms' function in the pre-processing step helps interpret its treatment of time series data specifically. 

The {\it data processing} step represents how the anomaly detection method processes the time series data at the initial step. We have noticed all the anomaly detection models are somehow based on a windowed approach initially - converting the time series data into a matrix with rows of sliding window slices of the original time series. The pre-processing step consists of the additional processing procedure besides the window transformation, which varies from statistical feature extraction to fitting a machine learning model and building a neural network.

After the data is processed, different {\it detection methods} are implemented on the dataset, which might be simply calculating distances among the processed sub-sequences, fitting a classification hyper-plane, or using the processed model to generate new sub-sequences and comparing them with original sub-sequences. The detection methods are usually traditional outlier detection methods for vector data. 

Then, during the {\it scoring} process, the results derived in the detection methods will be converted to an anomaly score that assesses the abnormality of individual sub-sequences by a real value (such as a probability of being an anomaly). The scores will be further used to infer the score of the individual point. The resulting score is a time series of the same length as the initial time series.

Lastly, during the {\it post-processing} step, the anomaly score time series is processed to extract the anomalous points or intervals. Usually, a threshold value will be determined, where the points with anomaly scores surpassing the threshold will be marked as the anomaly.

This categorization of time-series anomaly detection pipelines inspires our process-centric taxonomy of the detection methods, which will be discussed thoroughly in the next section.

\section{Anomaly Detection Taxonomy}

\begin{figure}[t]
 \centering
 \includegraphics[scale=0.56]{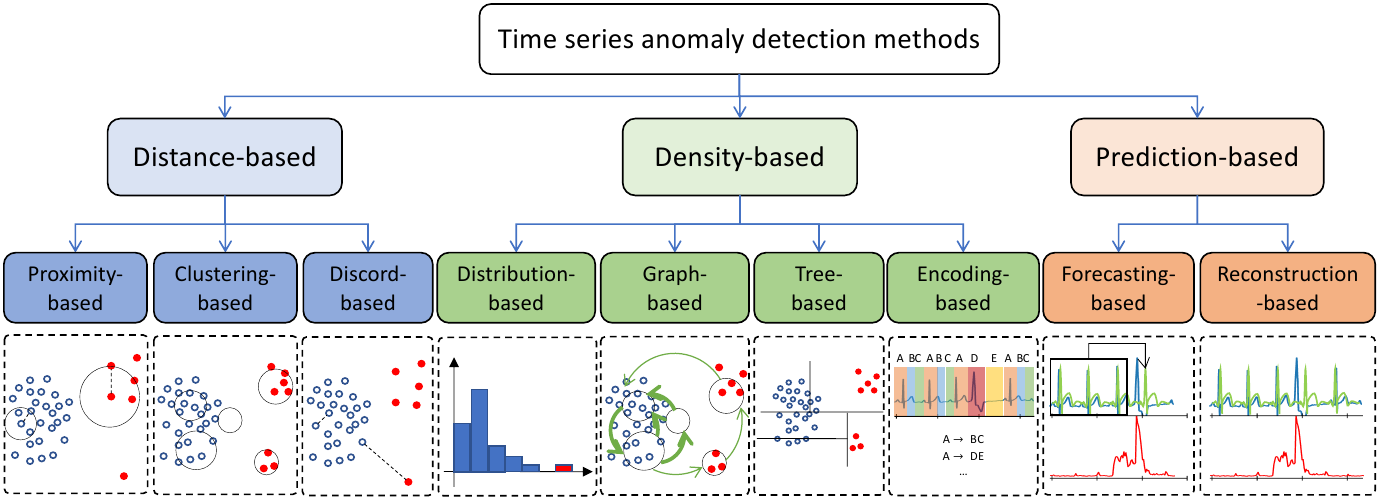}
 \caption{Process-centric anomaly detection taxonomy.}
 \label{fig:taxanomy}
\end{figure}

In this section, we describe our proposed process-centric taxonomy of the detection methods. We divide methods into three categories: (i) {\it distance-based}, (ii) {\it density-based}, and (iii) {\it prediction-based}. The first family contains methods that focus on the analysis of sub-sequences for the purpose of detecting anomalies in time series, mainly by utilizing distances to a given model. Second, instead of measuring nearest-neighbor distances, density-based methods focus on detecting globally normal or isolated behaviors. Third, prediction-based methods aim to train a model (on anomaly-free time series or with very few anomalies) to reconstruct the normal data or predict the future expected normal points. In the following sections, we break down each category into process-centric subcategories. Figure~\ref{fig:taxanomy} illustrates our proposed process-centric taxonomy.

Note that the second-level categorization is not mutually exclusive. A model might compress the time series data while adopting a discord-based identification strategy. In this case, the model falls within two different sub-categories. In the table of methods, only one of the second-level will be listed to give a clearer representation. 

\subsubsection{{\bf Distance-based}}

As its name suggests, the distance-based method detects anomalies purely from the raw time series using distance measures. Given two sequences (or univariate time series), $A \in \mathbb{R}^{\ell}$ and $B \in \mathbb{R}^{\ell}$, of the same length, $\ell$, we define the distance between $A$ and $B$ as $d(A,B) \in \mathbb{R}$, such as $d(A,B)=0$ when $A$ and $B$ are the same. There exist different definitions of $d$ in the literature. The classical distance widely used is the Euclidean distance or the Z-normalized Euclidean distance (Euclidean distance with sequences of mean values equal to 0 and standard deviations equal to 1). Then, Dynamic Time Warping (DTW) is commonly used to cope with misalignment issues.
Overall, the distance-based algorithms merely treat the numerical value of time series as it is, without further modifications such as removing seasonality or introducing a new structure built on the data. Within the distance-based models, there come three second-level categories: proximity-based, clustering-based, and discord-based models.

\begin{enumerate}

\item The \textit{{\bf proximity-based}} model measures proximity by calculating the distance of a given sub-sequence to its close neighborhood. The isolation of a sub-sequence with regards to its closest neighbors is the main criterion to consider if this sub-sequence is an anomaly or not.  This notion of isolation with regard to a given neighborhood has been proposed for non-time series data. Thus, the methods contained in this category have been introduced for the general case of multi-dimensional outlier detection. In our specific case, the sub-sequence of a time series can be considered a multi-dimensional point with the number of dimensions equal to the length of the sub-sequence.

\item The \textit{{\bf clustering-based}} model infers anomalies from a cluster partition of the time series sub-sequences. In practice, the anomaly score is calculated by the non-membership of a sub-sequence of each of the clusters learned by the model. Other considerations, such as cluster distance and cluster capacity, can also be considered. The clustering issue is related to the anomaly detection problem in that points may either belong to a cluster or be deemed anomalies. In practice, the fact that a sub-sequence belongs or not to a cluster is assessed by the computation of the distance between this sub-sequence and the cluster centroid or medoid.

\item  The \textit{{\bf discord-based}} model tries to identify efficiently specific types of sub-sequences in the time series named discord. Formally, a sub-sequence $A$ (or a given length $\ell$) is a discord, if the distance between its nearest neighbor is the largest among all the nearest neighbors' distances computed between sub-sequences of length $\ell$ in the time series. Overall, similarly to proximity-based methods, The isolation of a sub-sequence with regards to its closest neighbors is the main criterion to consider if this sub-sequence is an anomaly or not. However, on contrary to proximity-based methods, discord-based methods have been introduced for the specific case of anomaly detection in time series. Thus, as such methods introduced efficient processes for time series distance computation specifically, we group them into one different sub-category.

\end{enumerate}

\subsubsection{{\bf Density-based}}

The density-based does not treat the time series as simple numerical values but imbues them with more complex architecture. The density-based method processes time series data on top of a representation of the time series that aims to measure the density of the points or sub-sequence space. Such representation varies from graphs, trees, and histograms to a grammar induction rule. The density-based models have four second-level categories: distribution-based, graph-based, tree-based, and encoding-based. 

\begin{enumerate}
    
\item  The \textit{{\bf distribution-based}} anomaly detection approach is building a distribution from statistical features of the points or sub-sequences of the time series. By examining the distributions of features of the normal sub-sequences, it tries to recover relevant statistical models and then uses them to infer the abnormality of the data.

\item A \textit{{\bf graph-based}} method represents the time series and the corresponding sub-sequences as a graph. The nodes and edges represent the different types of sub-sequences (or representative features) and their evolution in time. For instance, the nodes can be sets of similar sub-sequences (using a predefined distance measure), and the edge weights can be the number of times a sub-sequence of a given node has been followed by a sub-sequence of another node. The detection of anomalies is then achieved using characteristics of the graph, such as the node and edge weights, but also the degree of the nodes.

\item  A \textit{{\bf tree-based}} approach aims to divide the point or sub-sequence of a time series using trees. For instance, such trees can be used to split different points or sub-sequences based on their similarity. The detection of anomalies is then based on the statistics and characteristics of the tree, such as its depth.

\item  A \textit{{\bf encoding-based}} anomaly detection model compresses or represents the time series into different forms of symbols. The encoding-based model suggests that a time series can be interpreted as a sequence of context-free, discrete symbols or states. For instance, anomalies can be detected by using grammar rules with the symbols extracted from the time series. It should be noted that an encoding-based model is not exclusive to itself; it may even be based on a graph-based or tree-based model.

\end{enumerate}

\color{black}

\subsubsection{{\bf Prediction-based}}

Prediction-based methods aim to detect anomalies by predicting the expected normal behaviors based on a training set of time series or sub-sequences (containing anomalies or not). For instance, some methods will be trained to predict the next value or sub-sequence based on the previous one. The prediction is then post-processed to detect abnormal points or sub-sequences. Then, the prediction error is used as an anomaly score. The underlying assumption of prediction-based methods is that normal data are easier to predict, while anomalies are unexpected, leading to higher prediction errors. Such assumptions are valid when the training set contains no or few time series with anomalies. Therefore, prediction-based methods are usually more optimal under semi-supervised settings. 

\begin{enumerate}

\item The \textit{{\bf forecasting-based}} method is a model that, for a given index or timestamp, takes as input points or sub-sequences before this given timestamp and predicts its corresponding value or sub-sequence. In other words, such methods use past values as input to predict the following one. The forecasting error (the difference between the predicted and the real value) is used as an anomaly score. Indeed, such forecasting error is representative of the expectation of the current value based on the previous points or sub-sequences. The larger the error, the more unexpected the value, and thus, potentially abnormal. Forecasting-based approaches assume that the training data (past values or sub-sequences) is almost anomaly-free. Thus, such methods are mostly semi-supervised.

\item The \textit{{\bf reconstruction-based}} method corresponds to a model that compresses the input time series (or sub-sequence) into a latent space (smaller than the input size) and is trained to reconstruct the input time series from the latent space. The difference between the input time series and the reconstructed one (named the reconstruction error) is used as an anomaly score. As for forecasting-based methods, the larger the error, the more unexpected the value, and thus, potentially abnormal. Moreover, as the reconstruction error is likely to be small for the time series used to train the model, such reconstruction methods are optimal in semi-supervised settings. 

\end{enumerate}

\subsection{Scoring Process}

\begin{figure}[t]
 \centering
 \includegraphics[scale=0.74]{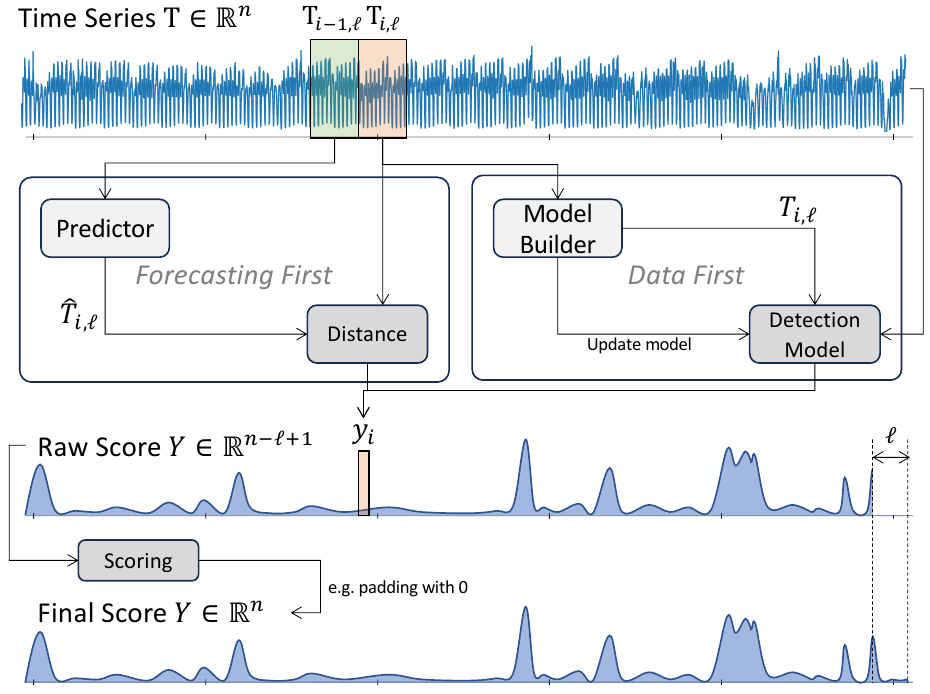}
 \caption{The scoring process.}
 \label{fig:pipeline}
\end{figure}

As summarized in the pipeline, anomaly detection algorithms distinguish outliers by inference on the ``anomaly scores" of each temporal data point. The anomalies are marked by points whose scores exceed the threshold value. Due to the special features of time series data, a further general categorization can be provided based on the algorithm's strategy of scoring the anomalies. We include this generalization as a complement to our taxonomy.

A \textit{forecasting-first} approach first infers the values of interested time series, without knowing the actual values of the dataset, and then determines if the coming data points are anomalies (based on their distance to the inferred values). This gives possibilities for streaming data anomaly detection. A \textit{data-first approach}, on the other hand, reads the data first to update the model. Then, the entire training data samples will be used to compare with the arriving data via the detection model. Figure \ref{fig:pipeline} gives an illustration of the two.

\begin{figure}[t]
	\centering
	\includegraphics[scale=0.55]{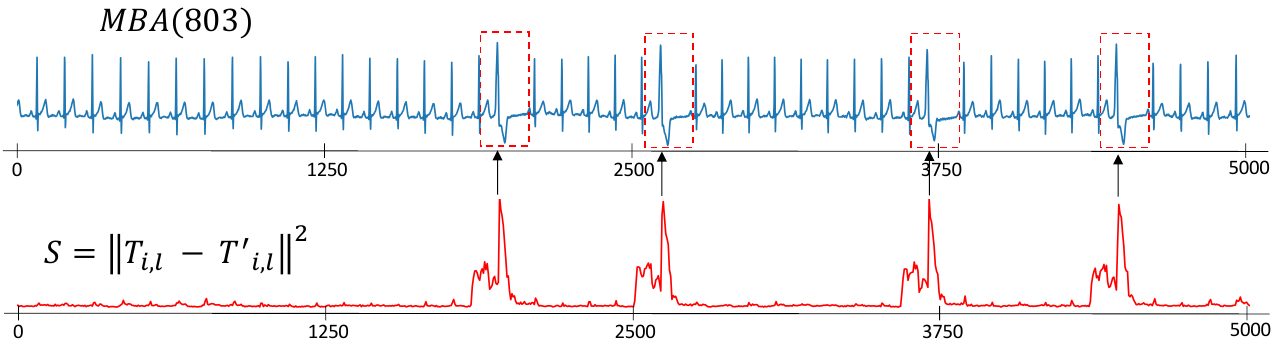}
	\caption{Result using $|\mathcal{Z}| = 16$ for a autoencoder Encoder($Conv(64,3)$-$Relu()$-$Dense()$-$Tanh()$), Decoder($DeConv(64,3)$-$Relu()$-$Dense()$-$Tanh()$). Top plot: Input time series snippet. Bottom plot: $S$ (using mean square error) for all the sub-sequences of length $80$.}
	\label{autoencoder_result_example}
\end{figure}

Just like forecasting-first algorithms, data-first algorithms may also be capable of generating new sub-sequences to compare with original sequences. Figure \ref{autoencoder_result_example} gives such an example, where the autoencoder reconstructs an estimated sequence $T_{i,l}^\prime$ to calculate the error $S$ on the ECG data. Although it behaves like a forecasting-first method by trying to ``forecast" the original sub-sequences, it is technically data-first because it requires the arrival of new data to make valid comparisons.

\section{Survey Organization}

\begin{figure}[t]
 \centering
 \includegraphics[width=\linewidth]{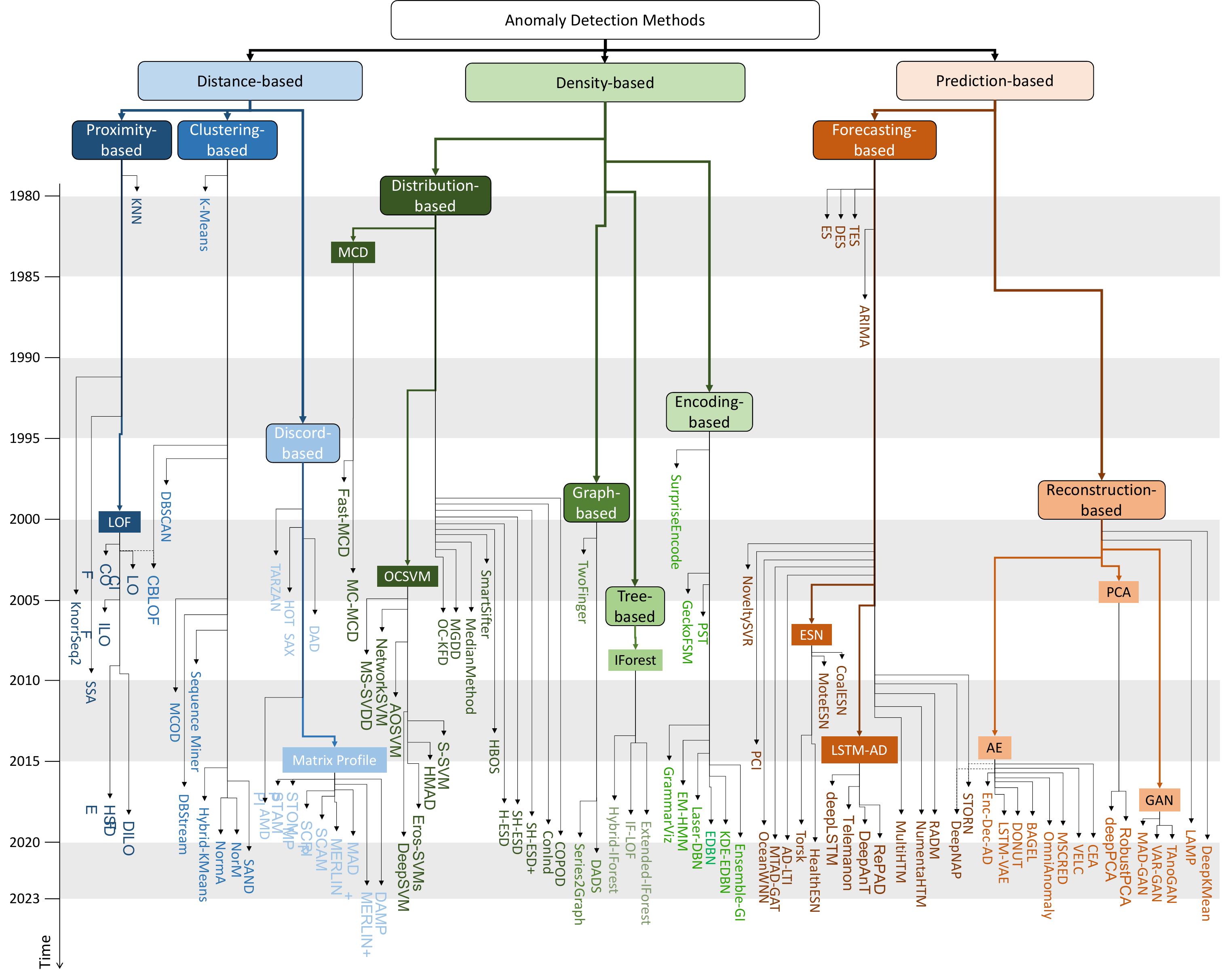}
 \caption{Illustration of Anomaly Detection Taxonomy for all methods.}
 \label{fig:bigtaxanomy}
\end{figure}

In the following sections, we will present the State-Of-The-Art (SOTA) in the three major categories and elaborate on the specific variants of the SOTA proposed in the past literature. Figure~\ref{fig:bigtaxanomy} illustrates our detailed proposed taxonomy listing all the methods discussed in this paper. Note that, in  Figure~\ref{fig:bigtaxanomy}, the names of the methods (the first letter) are positioned on the y-axis based on their publication date. Even though some concepts might be anterior to the date indicated in Figure~\ref{fig:bigtaxanomy} (for instance the concept of k-means was introduced in 1967), the dates correspond to the first paper discussing the concepts applied to anomaly detection.
The rest of this paper is organized as follows:

\begin{itemize} 
\item We first present {\bf distance-based} methods that perform anomaly detection using distance computation and comparisons on points or sub-sequences of the time series. 

\item Next we enumerate the {\bf density-based} methods. These approaches process time series data on top of a representation that aims to measure the density of the points or sub-sequences within the time series space.

\item We furnish with two groups of {\bf prediction-based} methods that aim to predict the expected normal behaviors from a training set of time series. These two groups are forecasting-based methods (that use the forecasting error as anomaly score) and reconstruction-based methods (that are trained to reconstruct an input time series and use the reconstruction error as anomaly score).

\item We will also include a table of all the methods in each section to reveal their other characteristics (such as the requirement for supervision, the capability of handling streaming data, etc) as complements to our taxonomy.

\item After briefly describing all the methods, we will discuss a meta-analysis of the time-series anomaly detection community by examining the evolution and the trends of each category (distance-based, density-based, prediction-based). We will also have a closer look at the evolution in time of proposed methods that are semi-supervised, unsupervised, and able to handle multivariate time series.

\item We will conclude this survey with the evaluation of such methods. We will first enumerate existing benchmarks proposed in the literature, as well as different evaluation measures and their limitations when applied to time-series anomaly detection.

\end{itemize}



\section{Time Series Notations}

We now introduce some formal notations related to time series. 
We define a time series $T \in \mathbb{R}^n $ as a sequence of real-valued numbers $T_i\in\mathbb{R}$ $[T_0,T_2,...,T_{n-1}]$, where $n=|T|$ is the length of $T$, and $T_i$ is the $i^{th}$ point of $T$.

A multivariate, or $D$-dimensional time series $\mathbf{T} \in \mathbb{R}^{(D,n)}$ is a set of $D$ univariate time series of length $n$. 
We note $\mathbf{T} = [T^{(0)}, ... , T^{(D-1)}]$ and for $j \in [0,D-1]$, we note the univariate time series $T^{(j)} = [T^{(j)}_{0}, T^{(j)}_{1}, ... ,T^{(j)}_{n-1}]$.
A subsequence $T^{(j)}_{i,\ell} \in \mathbb{R}^{\ell}$ of the dimension $T^{(j)}$ of the multivariate time series $T$ is a subset of contiguous values from $T^{(j)}$ of length $\ell$ (usually $\ell \ll n$) starting at position $i$; formally, $T^{(j)}_{i,\ell} = [T^{(j)}_{i}, T^{(j)}_{i+1},...,T^{(j)}_{i+\ell-1}]$. The multivariate subsequence is defined as $T_{i,\ell} = [T^{(0)}_{i,\ell}, ... ,T^{(D-1)}_{i,\ell}]$.
For a given univariate time series $T$, the set of all subsequences in $T$ of length $\ell$ is defined as $\mathbb{T}_{\ell} = \{T_{0,\ell},T_{1,\ell},...,T_{|T| - \ell,\ell}\}$.

\section{Distance-based Methods}

In this section, various distance-based anomaly detection methods are introduced. We enumerate the methods in three categories described in the following section. We enumerate all the mentioned methods in Table~\ref{distance-based_table}.

\begin{table}[ht]
\footnotesize

\caption{Summary of the distance-based anomaly detection methods.}
\vspace{-0.3cm}
\begin{tabular}[t]{lccccc}
\toprule
&Second Level &Prototype  &Dim &Method & Stream\\

\toprule
KNN \cite{hawkins_identification_1980}          & Proximity-based& Nearest Neighbor &M & U &\xmark \\
KnorrSeq2 \cite{Palshikar2005DistanceBased}     & Proximity-based& Nearest Neighbor &M & U &\xmark \\
LOF \cite{BreunigEtAl2000LOF}                   & Proximity-based &LOF &M & U &\xmark \\
COF \cite{TangEtAl2002Enhancing}                & Proximity-based &LOF &M & U &\xmark \\
LOCI \cite{PapadimitriouEtAl2003LOCI}           & Proximity-based &LOF &M & U &\cmark \\
ILOF \cite{PokrajacEtAl2007Incremental}         & Proximity-based &LOF &M & U &\cmark \\
DILOF \cite{NaEtAl2018DILOF}                    & Proximity-based &LOF &M & U &\cmark \\
HSDE \cite{LiEtAl2017LocalityBased}             & Proximity-based & LOF & I& U & \xmark \\
\hline
k-means \cite{hawkins_identification_1980}      & Clustering-based& k-means &M & U &\xmark \\
Hybrid-k-means \cite{Song2017}                  & Clustering-based& k-means &M & U &\xmark \\
DeepkMeans \cite{MORADIFARD2020185}             & Clustering-based & k-means & M & Se & \xmark \\
DBSCAN \cite{Sander1998}                        & Clustering-based&DBSCAN &M & U &\xmark \\
DBStream \cite{10.1109/TKDE.2016.2522412}       & Clustering-based&DBSCAN &M & U &\cmark \\
MCOD \cite{KontakiEtAl2011Continuous}           & Clustering-based& - &I & U &\xmark \\
CBLOF \cite{HeEtAl2003Discovering}              & Clustering-based &LOF &M & U &\xmark\\
sequenceMiner \cite{BudalakotiEtAl2009Anomaly}  & Clustering-based & - &I & U &\xmark \\
NorM (SAD) \cite{BoniolEtAl2020Automated}       & Clustering-based&NormA &I & U &\xmark \\
NormA \cite{boniol_unsupervised_2021}           & Clustering-based&NormA &I & U &\xmark \\
SAND \cite{boniol2021sand}                      & Clustering-based&NormA &I & U &\cmark \\
\hline
TARZAN\cite{KeoghEtAl2002Finding}               & Discord-based& - &I & S &\xmark \\
HOT SAX \cite{KeoghEtAl2005HOT}                 & Discord-based& - &I & U &\xmark \\
DAD \cite{YankovEtAl2007Disk}                   & Discord-based& - &I & U &\xmark \\
AMD \cite{YangLiao2017Adjacent}                 & Discord-based& - &I & U &\xmark \\
STAMPI \cite{YehEtAl2016Matrix}                 & Discord-based&Matrix Profile &M & U &\cmark \\
STOMP \cite{ZhuEtAl2016Matrix}                  & Discord-based&Matrix Profile &M & U &\xmark \\
MERLIN~\cite{DBLP:conf/icdm/NakamuraIMK20}      & Discord-based & Matrix Profile & I & U & \xmark  \\
MERLIN++~\cite{MERLINplusplus}                  & Discord-based & Matrix Profile  & I & U  &  \xmark  \\
SCRIMP \cite{ZhuEtAl2018Matrix}                 & Discord-based&Matrix Profile &I & U &\xmark \\
SCAMP \cite{ZimmermanEtAl2019Matrix}            & Discord-based&Matrix Profile &I & U &\xmark \\
VALMOD \cite{LinardiEtAl2020Matrix}             & Discord-based&Matrix Profile &I & U &\cmark \\
DAMP \cite{10.1145/3534678.3539271}             & Discord-based&Matrix Profile &I & U &\cmark \\
LAMP \cite{ZimmermanEtAl2019Matrixa}            & Discord-based & Matrix Profile & I & Se & \cmark \\

\toprule
\end{tabular}
    \begin{tablenotes}
      \scriptsize
      \centering
      \item  I: Univariate; M: Multivariate // S: Supervised; Se: Semi-Supervised U: Unsupervised
     \end{tablenotes}
\label{distance-based_table}
\vspace{-0.5cm}
\end{table}%

\subsection{Proximity-based Methods}

Proximity-based methods use distance to close neighborhoods as the main step to detect anomalies. We detail below two method types of proximity-based methods.

\subsubsection{Kth Nearest Neighbor}
One of the first distance-based and proximity-based methods introduced in the literature for anomaly detection is using K-th Nearest Neighbor (KNN) principle~\cite{hawkins_identification_1980}.
KNN-type methods utilize a metric among neighboring sub-sequences to infer the abnormality scores of the time series' sub-sequences.
A distance measure $d(\cdot, \cdot)$ (also called dissimilarity measure) is used to find the nearest neighbors for each subsequence.
Common distance measures are Euclidean, Manhatten, or in general, Minkowski distances.
The k-anomaly score $\mathcal{A}_k : \{T_{i,\ell}\}_{i \in \mathcal{I}} \to \mathbb{R}$ for the set of time series' sub-sequences $\{T_{i,\ell}\}_{i \in \mathcal{I}}$ is calculated based on each sub-sequences' $k^{th}$ nearest neighbors using a variable aggregation function $agg: \mathbb{R}^k \to \mathbb{R}$:
\begin{equation}
    \mathcal{A}_k (T_{i,\ell}) = \inf_{\mathcal{J} \subset \mathcal{I}, |\mathcal{J}|=k+1} agg_{j \in \mathcal{J}} d(T_{i,\ell}, T_{j,\ell})
\end{equation}
\noindent
In the above equation, $\ell$ is the fixed length of the sliding window, and $k+1$ accounts for trivial matches.
Following the intuition of~\cite{10.1145/335191.335437} that the anomaly score for a subsequence $T_{i,\ell}$ is the distance to its $k^{th}$ nearest neighbor, we can use $agg = \sum$.
Alternative proposals for $\mathcal{A}_k$ may utilize other aggregation methods, such as median, minimum, or other functions that pool the distances into scalar scores.
With different aggregation functions and distance metrics choices, one can propose different KNN-type models that are appropriate for various anomaly detection problems.

In addition to the standard KNN technique~\cite{10.1145/335191.335437}, where the maximum distance to the $k^{th}$ nearest neighbor is used as anomaly score, other variants of KNNs have been suggested by researchers.
For instance, KnorrSeq and KnorrSeq2 are also two variants of KNNs proposed in the litterature~\cite{Palshikar2005DistanceBased}.
The first algorithm KnorrSeq is based on a tumbling window and discovers only global outliers by marking those sub-sequences for which at least $p\%$ of the other subsequence are further away than a threshold $D$.
Their second algorithm KnorrSeq2 is an implementation of KNN that detects sub-sequences as outliers if at least $p\%$ of the $k$ preceding or $k$ succeeding sub-sequences are further away than a threshold $D$.
The anomaly score is calculated using $\sum$ as the aggregation function:
\begin{equation}
    \mathcal{A}_k (T_{i,\ell}) =  \inf_{\mathcal{J} \subset \mathcal{I}, |\mathcal{J}|=2k+1,\forall j \in \mathcal{J}, |j-i| \leq k} \sum_{j \in \mathcal{J}} \begin{cases}
        1, & \text{if } d(T_{i,\ell}, T_{j,\ell}) > D \\
        0, & \text{otherwise}
    \end{cases}
\end{equation}
\noindent
The anomalous sub-sequences are selected using a threshold $\tau = pk$ on the anomaly scores: $\mathcal{A}_k (T_{i,\ell}) <= \tau$.

\subsubsection{Local Outlier Factor}
The most commonly used proximity-based approach is the Local Outlier Factor (LOF)~\cite{Breunig:2000:LID:342009.335388}, which measures the degree of being an outlier for each instance. Unlike the previous proximity-based models, which directly compute the distance of sub-sequences, LOF depends on how the instance is isolated to the surrounding neighborhood. This method aims to solve the outlier detection task where an outlier is considered as {\it ``an observation that deviates so much from other observations as to arouse suspicion that it was generated by a different mechanism"} (Hawkins definition \cite{hawkins_identification_1980}). This definition is coherent with the anomaly detection task in time series where the {\it different mechanism} can be either an arrhythmia in an electrocardiogram or a failure in the components of an industrial machine. 

First, let's consider $T_{i,\ell}$ and $T_{j,\ell}$ two sub-sequences of the same time series. In the following paragraphs, we note these two sub-sequences, A and B, respectively. Given $k$-$distance(A)$ the distance between $A$ and its $k^{th}$ nearest neighbor ($N_k(A)$ the set of these $k$ neighbors), LOF is based on the following reachability distance definition:

\begin{figure}[t]
 \centering
 \includegraphics[scale=0.50]{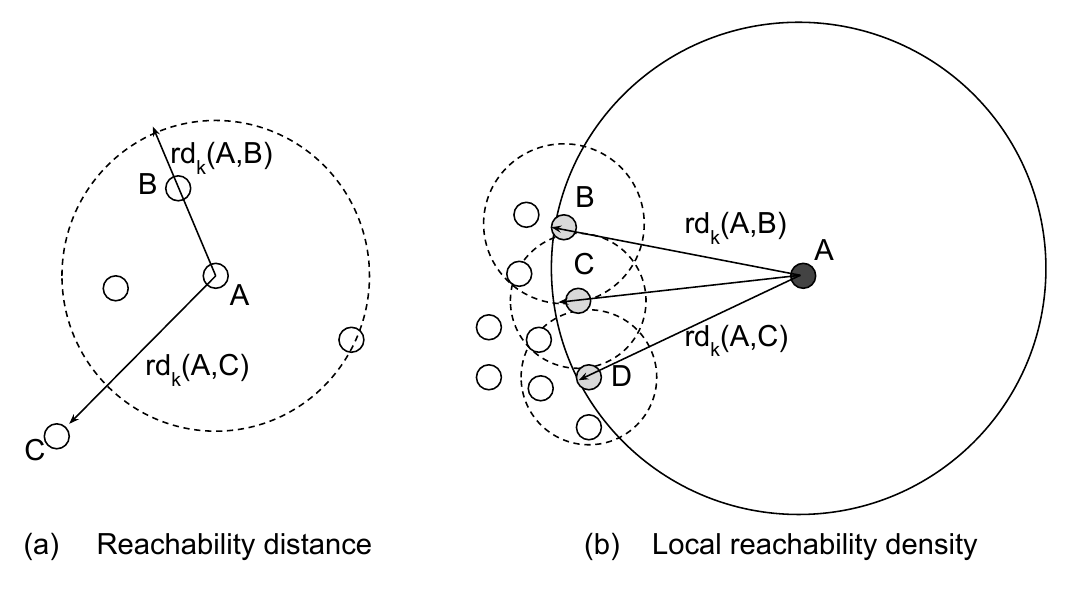}
 \caption{(a) Reachability distance between A and B, A and C for $k=4$. (b) Difference between $rd_k(A,X), X \in N_k(A)$, when A is an anomaly and B, C, and D are regular instances. }
 \label{fig:LOF_exp}
\end{figure}

\begin{equation}
rd_k(A,B) = max(k\text{-}distance(B),d(A,B)) 
\end{equation}

As illustrated in Figure~\ref{fig:LOF_exp}, the main concept behind this distance definition is to stress out the homogeneity of distances between instances within the $k$-neighborhood (i.e., the $k$-neighborhood will have the same distance between each other). Thus the local reachability can be defined as follow:

\begin{equation}
lrd_k(A) = \frac{|N_k(A)|}{\sum_{B \in N_k(A)} rd_k(A,B)}
\end{equation}

Given an instance, $A$, $lrd_k(A)$ is the inverse of the average reachability of A from its neighborhood, i.e., the average distance at which $A$ can be reached from its neighbors. Therefore, LOF is given by:

\begin{equation}
LOF_k(A) = \frac{\sum_{B \in N_k(A)} \frac{lrd_k(B)}{lrd_k(A)}}{|N_k(A)|} = \frac{\frac{\sum_{B \in N_k(A)} lrd_k(B)}{|N_k(A)|}}{lrd_k(A)}
\end{equation}

Intuitively,  the $LOF_k$ of an instance is the average of the local reachability density of the neighbors divided by its own reachability density. Therefore, if we set sub-sequences of length $\ell$ as the length of the sub-sequence, this factor can be used as an anomaly score.

In the past decade, researchers also suggested many variants of the LOF method~\cite{Breunig:2000:LID:342009.335388}. COF~\cite{TangEtAl2002Enhancing}, for example, is a connectivity-based variant of LOF. It indicates how far away a point shifts from a pattern, adjusting the notion of isolation to not depend on the density of data clouds. LOCI~\cite{PapadimitriouEtAl2003LOCI} is another LOF-like algorithm that utilizes different statistics (correlation integral and MDEF) to infer individual points' isolation. Other LOF variants are the ILOF~\cite{PokrajacEtAl2007Incremental} and DILOF~\cite{NaEtAl2018DILOF} method, which adopts a faster algorithm and detects anomalies incrementally. Finally, the hierarchy-segmentation-based data extraction method (HSDE)~\cite{LiEtAl2017LocalityBased} is inspired by the strategy of LOF to detect abnormal points in time series.

\subsection{Discord-based Methods}

A practical modification to the KNN-type model is to use the discord, which evolves from comparing distances to the nearest neighbor to comparing distances to the $k^{th}$ neighbor.  Such adaptations assist in edge conditions where a small number of anomalies are clustered along with limited distances, and the conventional KNN approach struggles to recognize them. The following gives the specific definitions of discord:

 \begin{definition}[Top-k $m^{th}$-discord] ~\cite{DBLP:conf/icdm/YehZUBDDSMK16,DBLP:conf/edbt/Senin0WOGBCF15,Keogh2007,Liu2009,DBLP:conf/adma/FuLKL06,DBLP:conf/sdm/BuLFKPM07,Parameter-Free_Discord,boniol_unsupervised_2021}
\label{def:k-m-discord}
Suppose the window is of length $\ell$. Given a collection of sub-sequences $\{T_{j,\ell}\}_{j \in \mathcal{I}}$, let $f_m$ denote $m^{th}$-discord function measuring the distance to $m^{th}$ nearest neighborhood so that $f_m(T_{j,l}) = \min_{j \in \mathcal{I} \setminus \{i\}}^m d(T_{i,l}, T_{j,l})$. A sub-sequence $T_{i,\ell}$ is a Top-k $m^{th}$-discord if $f_m(T_{i,\ell})$ is the $k^{th}$ maximum among the set $\{T_{j,\ell}\}_{j \in \mathcal{I}}$.
\end{definition}

Note that the $m^{th}$-discord is the special case of Top-k $m^{th}$-discord when $k = 1$, and discord is the special case of $m^{th}$-discord when $m = 1$.

\begin{figure}[t]
	\centering
	\includegraphics[width=0.5\linewidth]{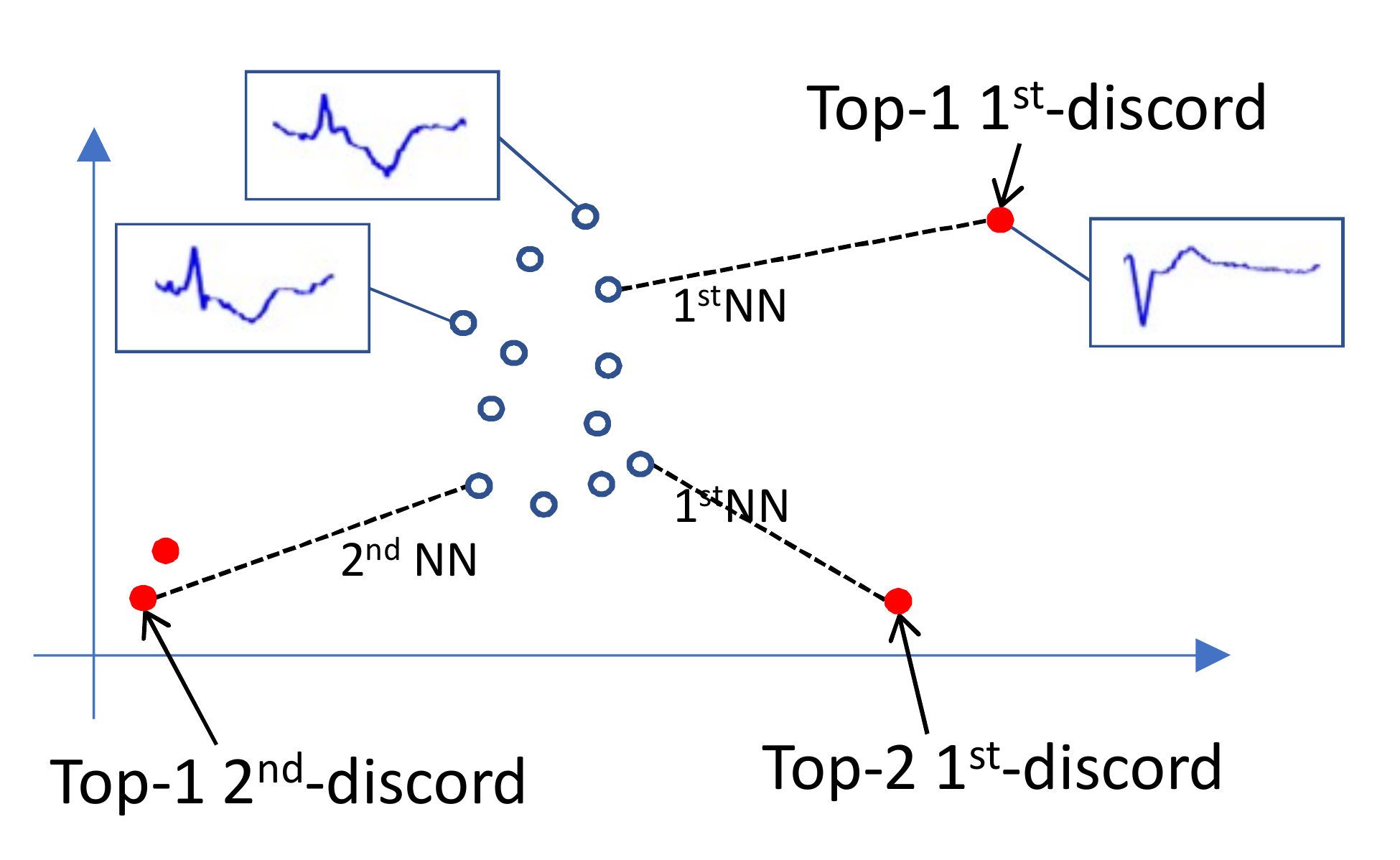}
\caption{A dataset with 16 sub-sequences (of the same length $\ell$) depicted as points in 2-dimensional space; 12 sub-sequences are normal (hollow points), and 4 are anomalous (solid, red points).}
	\label{Discord_representation}
\end{figure}

Figure~\ref{Discord_representation} illustrates these notions with an example, where for ease of exposition, we represent each sub-sequence as a point in 2-dimensional space.
The figure depicts two $1^{st}$-discords: the discord in the top-right ($Top$-$1$) has its 1-NN at a larger distance than the discord in the bottom-right ($Top$-$2$).
The figure also shows a group of two anomalous sub-sequences: one of them is the $Top$-$1$ $2^{nd}$-discord, and the other sub-sequence is its 1-NN (also a discord).

There exist several studies that have proposed fast and scalable discord discovery algorithms in various settings~\cite{DBLP:conf/edbt/Senin0WOGBCF15,Keogh2007,Liu2009,DBLP:conf/adma/FuLKL06,DBLP:conf/icdm/YehZUBDDSMK16,DBLP:conf/sdm/BuLFKPM07,DBLP:conf/icdm/YankovKR07,Parameter-Free_Discord}, including simple and $m^{th}$-discords\footnote{The authors of these papers define the problem as $k^{th}$-discord discovery.}, in-memory and disk-aware techniques, exact and approximate algorithms, using either their Symbolic Aggregate Approximation (SAX) \cite{Keogh2007,DBLP:conf/edbt/Senin0WOGBCF15} or Haar wavelets \cite{DBLP:conf/sdm/BuLFKPM07,DBLP:conf/adma/FuLKL06}. In the following sections, we present the state-of-the-art solutions to the sub-sequence anomaly detection problem. Note that in this discussion, we focus on the $Top$-$k$ anomalies (using instead a threshold $\epsilon$ to detect anomalies would be a straightforward modification of the solution).

Disk Aware Discord Discovery method (DAD)~\cite{DBLP:conf/icdm/YankovKR07}  is a method that proposes a new exact algorithm to discover discord requiring only two linear scans of the disk thought for managing very large datasets. The algorithm uses the raw sequences directly. First, it addresses the simpler problem of detecting {\it range discord}, then generalizes the problem to detect the $Top$-$k$ discord.

Other than DAD, several other discord-like anomaly identification approaches have also been proposed. TARZAN~\cite{KeoghEtAl2002Finding} is a discord method via SAX discretization through the sliding window. The approach processes data by building a suffix tree and calculating its anomaly score by applying inferences on the discord. Like TARZAN, HOT SAX~\cite{KeoghEtAl2005HOT} also adopts SAX discretization throughout the processing step; it then measures the distance to the nearest non-self match for sub-sequences to identify abnormalities. AMD~\cite{YangLiao2017Adjacent} further improves HOT SAX, which performs dynamic segmentation to vary the window length.

As a final note, we observe that in situations with multiple similar anomalies, we should either use a method that supports $m^{th}$-discords, or use a simple discord (i.e., $1^{st}$-discord) method as follows. 
Starting at the beginning of the series and proceeding to the right, we apply the discord method by only considering the points to the left of the current position, and every time an anomaly is detected, we search the entire series for similar anomalies (this will reveal all occurrences of the multiple similar anomalies). 
As we proceed to the right, the discord method will detect only the first occurrence of each set of similar anomalies (the rest being detected by the similarity search operation mentioned above).
Note that this solution implies that we have accumulated enough data at the beginning of the series for the first execution of the discord method.



\subsubsection{Matrix Profile}

Matrix Profile \cite{DBLP:conf/icdm/YehZUBDDSMK16,DBLP:conf/icdm/ZhuZSYFMBK16} is a discord-based method that represents time series as a matrix of closest neighbor distances. Compared to its predecessor, Matrix Profile proposed a new metadata time series computed effectively, capable of providing various valuable details about the examined time series, such as discords. \color{black} For simplicity, we can call this metadata series matrix profile, and we can define it as follows:

\begin{definition}[Matrix Profile]
\label{def:MatrixProfile}
	A matrix profile $MP$ of a time series $T$ of length $n$ is a time series of length $n-\ell-1$ where the $i^{th}$ element of $MP$ contains the Euclidean normalized distance of the sub-sequence of length $\ell$ of $T$ starting at $i$ to its nearest neighbor.
\end{definition}

However, the latter definition does not tell us where that neighbor is located. This information is recorded in the matrix profile index:

\begin{definition}[Matrix Profile Index]
\label{def:MatrixProfileIndex}
	A matrix profile index $I_{MP}$ is a vector of the index where $I_{MP} [i] = j$ and $j$ is the index of the nearest neighbor of $i$.
\end{definition}

Two general definitions of Join matrix computation can be inferred. The first, called {\it Self-Join}, corresponds exactly to the matrix profile. The second, called {\it Join}, corresponds to the same operation for two different time series. Formally we have the following:

\begin{figure}[t]
	\centering
	\includegraphics[scale=0.45]{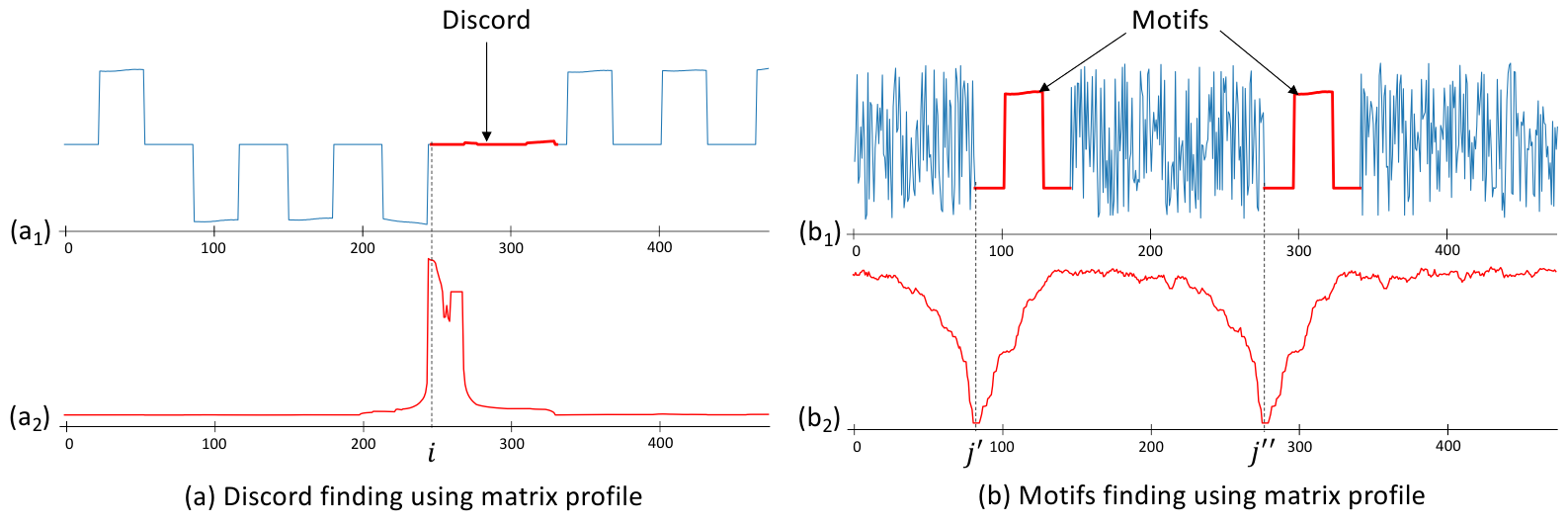}
	\caption{(a) Matrix profile ($a_2$) applied on the SED (Nasa disk failure datasets) time series snippet ($a_1$). The highest value in the matrix profile ($a_1$) points to the discord of the SED time series. (b) Matrix profile ($b_2$) applied on a synthetic time series ($b_1$). The smallest values in the matrix profile ($a_1$) point to a motif pair in the time series.}
	\label{Discord_example}
\end{figure}

\begin{definition}[Time Series Self-Join]  
	Given a time series $T$, the self-join of $T$ with sub-sequence length $\ell$, denoted by $T \mkern-5.8mu\bowtie_\ell T$, is a meta time series, where:
	$|T \mkern-5.8mu\bowtie_\ell T| = |T|-\ell+1$ and $\forall i. 1\le i \le |T \mkern-5.8mu\bowtie_{\ell} T|$, $(T \mkern-5.8mu\bowtie_{\ell} T)_{i,1} = dist (T_{i,\ell}$, $1^{st} NN$ of $T_{i,\ell}$).
	\label{SelfJoin}
\end{definition}

\begin{definition}[Time Series Join]  
	Given two time series $A$ and $B$, and a sub-sequence length $\ell$, the $Join$ between $A$ and $B$ denoted by $(A \mkern-5.8mu\bowtie_\ell B)$, is a meta time series, where:
	$|A \mkern-5.8mu\bowtie_\ell B| = |B|-\ell+1$ and $\forall i. 1\le i \le |A \mkern-5.8mu\bowtie_\ell B|$, $(A \mkern-5.8mu\bowtie_\ell B)_{i,1} = min(dist(B_{i,\ell},A_{1,\ell}),...,dist(B_{i,\ell},A_{|A|-\ell+1,\ell}))$. 
\end{definition}

These metadata are computed using Mueen’s ultra-fast Algorithm for Similarity Search (MASS) \cite{FastestSimilaritySearch} that requires just $O(n*log(n))$ time by exploiting the Fast Fourier Transform (FFT) to calculate the dot products between the query and all the sub-sequences of the time series. Once these metadata are generated, retrieving the $Top$-$k$ discord is possible by considering the maximum value of the Matrix Profile and ordering it, excluding the trivial matches (overlapping sub-sequences). Retrieving the sub-sequences with the shortest distance to their nearest neighbor (called $motifs$) is also possible. These sub-sequences correspond to a recurrent motif in the time series and can be useful in the anomaly search.

Figure~\ref{Discord_example} shows an example of the Matrix Profile metadata. On the one hand, Figure~\ref{Discord_example} (a) shows that the identified discord corresponds to a sub-sequence that deviates significantly from the normal cycles. On the other hand, Figure~\ref{Discord_example} (b) shows that the singular shapes are well-identified as motifs.

A family of Matrix Profile anomaly detection methods has also been proposed in the last decade. STAMP~\cite{YehEtAl2016Matrix} proposed an algorithm that can provide an accurate answer at any time during the full computation with time complexity of $O(n^2 log(n))$.  STAMPI~\cite{YehEtAl2016Matrix} not only performs the standard all-pairs-similarity-join of sub-sequences for matrix profile methods but also adapts the method incrementally to accommodate streaming purposes. STOMP~\cite{ZhuEtAl2016Matrix}, based on STAMP, developed a faster algorithm taking advantage of the sub-sequences order and achieving the computation with time complexity of $O(n^2)$. Moreover, a GPU implementation of STOMP has been proposed.  Like STOMP, SCAMP~\cite{ZimmermanEtAl2019Matrix} also adopts GPU for the matrix profile anomaly detection process. The SCRIMP method~\cite{ZhuEtAl2018Matrix} combines the STAMP algorithm (anytime) and STOMP (ordered) to make a hybrid approach. 
Moreover, the LAMP approach~\cite{ZimmermanEtAl2019Matrixa} is able to compute a constant time approximation of the MP value given a newly arriving time series subsequence.
While every aforementioned method can extract discords of a predefined length, VALMOD~\cite{LinardiEtAl2018Matrix} and MAD~\cite{valmodjournal} have been proposed to extract discords of variable length within a predefined length interval. Moreover, MERLIN~\cite{DBLP:conf/icdm/NakamuraIMK20} and MERLIN++~\cite{MERLINplusplus} have been proposed to identify discords of arbitrary length.
Finally, DAMP~\cite{10.1145/3534678.3539271}, a discord-based method, is able to work on online settings, and scale to fast-arriving streams.

\subsection{Clustering-based Methods}

Approaches based on clustering strategies have been proposed for the anomaly detection task. The main idea behind these methods is to partition the sub-sequence space and then evaluate how one sub-sequence fits into the partition.

\subsubsection{K-means Method}

The k-means clustering algorithm is a widely used unsupervised learning technique in data mining and machine learning. Its main objective is to partition a given dataset into $k$ distinct clusters, where each data point belongs to the cluster with the closest mean value. The algorithm operates iteratively, starting with an initial random assignment of $k$ centroids. 
For the specific case of time series, the Euclidean or the DTW distance is commonly used. 
K-means algorithm can also be used for anomaly detection in time series~\cite{hawkins_identification_1980}. The computational steps are the following:

\begin{itemize}
    \item All the sub-sequences of a given length $\ell$ (provided by the user) are clustered using the k-means algorithm. The Euclidean distance is used, and the number of clusters has to be provided by the user. 
    \item Once the partition is learned. We compute the anomaly scores of each sub-sequence based on the distance to the centroid of its assigned cluster.
    \item The larger the distance, the more abnormal the time series will be considered.
\end{itemize}

Such a method is straightforward but has been shown to be very effective for the specific case of multivariate time series~\cite{10.14778/3538598.3538602}. Moreover, extensions such as Hybrid-k-means~\cite{Song2017} can be used for anomaly detection as well. 
Finally, the k-means method can be used on top of other pre-processing and representation steps. 
For instance, DeepKMeans~\cite{MORADIFARD2020185} uses an Autoencoder to learn a latent representation of the time series and applies k-means on top of the latent space to identify anomalies.

\subsubsection{DBSCAN}

Another commonly used clustering-based method is the Density-Based Spatial Clustering of Application with Noise algorithm (DBSCAN)~\cite{Sander1998}. When identifying anomalies, DBSCAN marks data points into three categories: (i) core points, (ii) border points, and (iii) anomalies. To classify the points, two non-parametric parameters are important to detect the potential anomalies using DBSCAN: the radius $\epsilon$ of neighbors of the analyzed point and the minimum number $\mu$ of points in each normal cluster. Given these parameters, one can identify the anomalous following the categorization of the sub-sequences as follows:

\begin{itemize}
    \item A $\epsilon-$neighborhood of $T_{i,\ell}$ is $B_{\phi}(T_{i,\ell}, \epsilon) \cap T$, where $T$ is the training data $\{T_{i,\ell}\}_{i \in I}$. And $B_{\phi}(T_{i,\ell}, \epsilon)$ is the ball of radius $\epsilon$ centered at $T_{i,\ell}$ with respect to the norm $\phi(\cdot, \cdot)$.
    \item $T_{i,\ell}$ is a \textit{core point} if the size of the $\epsilon-$neighborhood of $T_{i,\ell}$ is  greater than $\mu$. 
    \item $T_{i,\ell}$ is a \textit{border point} if $T_{i,\ell}$ contains a core point in its $\epsilon-$neighborhood.
    \item $T_{i,\ell}$ is identified as an anomalous sub-sequence if $T_{i,\ell}$ is neither a border nor a core point. 
\end{itemize}

DBSCAN has been applied for anomaly detection on a univariate time series that contains observations with average daily temperature over 33 years~\cite{Celik2011AnomalyDI}. The processing step is to first convert the dataset into sub-sequences set with a sliding window, which are further z-normalized. After the processing step, DBSCAN is applied to the sub-sequences, and the anomalies are detected accordingly, as the method above prescribes. Similar clustering approaches such as DBStream~\cite{10.1109/TKDE.2016.2522412} can be used for anomaly detection.

\subsubsection{Other Clustering-based Methods}

Another clustering-based time-series anomaly detection method is the MCOD~\cite{KontakiEtAl2011Continuous} method. 
MCOD maintains a set of micro-clusters containing only normal objects (in our case: sub-sequences) to efficiently and robustly detect outliers in the event stream.
MCOD determines an object $x$ as an outlier if there are less than $k$ clusters at a distance of $R$ from $x$.
We can represent this binary decision using the following product function:
\begin{equation}
    \mathcal{A}_k (T_{i,\ell}) =  \inf_{\mathcal{J} \subset \mathcal{I}, |\mathcal{J}|=k+1} \prod_{j \in \mathcal{J}} \begin{cases}
        1, & \text{if } d(T_{i,\ell}, T_{j,\ell}) > R \\
        0, & \text{otherwise}
    \end{cases}
\end{equation}
\noindent

The function above returns discrete values $1$ and $0$ only.
So $\mathcal{A}_k$ is $1$ if and only if all $k$ nearest neighbors are at least $R$ distance apart from the considered subsequence.
Moreover, due to its similarity with KNN-based methods, it is important to note that MCOD can also be associated with proximity-based approaches. 

Another clustering-based approach is CBLOF~\cite{HeEtAl2003Discovering}, a LOF-based clustering algorithm, which first clusters the data and then assigns the CBLOF factor to each entry to measure both the size and relative of and among the individual clusters. 

Then, Sequenceminer~\cite{BudalakotiEtAl2009Anomaly} is an approach proposed by NASA. It clusters the sequences using the longest common sub-sequence (LCS) metric and ranks cluster members based on LCS, and selects the top $p\%$ as anomalies. The anomalies are identified by the parts of the sequence that differ the most and characterizes anomalous edit.

More recently, NormA~\cite{boniol_unsupervised_2021,BoniolEtAl2020SAD,BoniolEtAl2020Automated} is a clustering-based algorithm that summarizes the time series with a weighted set of sub-sequences. The normal set (weighted collection of sub-sequences to feature the training dataset) results from a clustering algorithm (Hierarchical), and the weights are derived from cluster properties (cardinality, extra-distance clustering, time coverage). An extension of NormA, called SAND~\cite{boniol2021sand}, has been proposed for streaming time series. 
The main difference between NormA and SAND lies in the approach used to update the weight in a streaming manner. Additionally, the clustering step in SAND is performed using the k-Shape method~\cite{paparrizos_k-shape_2016,paparrizos2017fast,paparrizos2023odyssey}, whereas NormA employs a hierarchical clustering method.

\section{Density-based Methods}

Unlike distance-based approaches, the density-based approach does not treat the time series as simple numerical values but imbues them with more complex representations. The density-based method processes time series data on top of a representation of the time series that aims to measure the density of the points or sub-sequence space. Such representation varies from graphs, trees, and histograms to a grammar induction rule. The density-based models have four second-level categories: distribution-based, graph-based, tree-based, and encoding-based. We enumerate all the mentioned methods in Table~\ref{density-based_table}.

\begin{table}[ht]
\footnotesize

\caption{Summary of the density-based anomaly detection methods.}
\vspace{-0.3cm}
\begin{tabular}[t]{lccccc}
\toprule
&Second Level &Prototype  &Dim &Method & Stream\\

\toprule
FAST-MCD \cite{Fast_MCD_1999}& Distribution-based & MCD & M & Se & \xmark \\
MC-MCD \cite{HARDIN2004625}& Distribution-based & MCD & M & Se & \xmark \\

OCSVM \cite{MaPerkins2003Timeseries}                & Distribution-based & SVM & M & Se & \xmark \\
AOSVM \cite{Gomez-VerdejoEtAl2011Adaptive}          & Distribution-based & SVM & M & U & \cmark \\
Eros-SVMs \cite{LamriniEtAl2018Anomaly}             & Distribution-based & SVM & M & Se & \xmark \\
S-SVM \cite{BhargavaRaghuvanshi2013Anomaly}         & Distribution-based & SVM & I & Se & \xmark \\
MS-SVDD \cite{XiaoEtAl2009Multisphere}              & Distribution-based & SVM & M & Se & \xmark \\
NetworkSVM \cite{ZhangEtAl2007One}                  & Distribution-based & SVM & M & Se & \xmark \\
HMAD \cite{GornitzEtAl2015Hidden}                   & Distribution-based & SVM & I & Se & \xmark \\
DeepSVM \cite{deepocsvm}                            & Distribution-based & SVM & M & U & \xmark \\

HBOS \cite{Goldstein_histogram-basedoutlier}        & Distribution-based & - & M & U & \xmark \\
COPOD \cite{li2020copod}                            & Distribution-based & - & M & U & \xmark \\
ConInd \cite{AntoniBorghesani2019statistical}       & Distribution-based & - & M & Se & \xmark \\
MGDD \cite{SubramaniamEtAl2006Online}               & Distribution-based & - & M & U & \cmark \\
OC-KFD \cite{Roth2006Kernel}                        & Distribution-based & - & M & U & \xmark \\
SmartSifter \cite{YamanishiEtAl2004OnLIne}          & Distribution-based & - & M & U & \cmark \\
MedianMethod \cite{BasuMeckesheimer2007Automatic}   & Distribution-based & - & I & U & \cmark \\
S-ESD \cite{HochenbaumEtAl2017Automatic}            & Distribution-based & ESD & I & U & \xmark \\
S-H-ESD \cite{HochenbaumEtAl2017Automatic}          & Distribution-based & ESD & I & U & \xmark \\
SH-ESD+ \cite{VieiraEtAl2018Enhanced}               & Distribution-based & ESD & I & U & \xmark \\

\hline
TwoFinger \cite{Marceau2000Characterizing}          & Graph-based & - & I & Se & \xmark \\
GeckoFSM \cite{SalvadorChan2005Learning}            & Graph-based & - & M & S & \xmark \\
Series2Graph \cite{BoniolPalpanas2020Series2Graph}& Graph-based & Series2Graph & I & U & \xmark \\
DADS \cite{10.1007/s00778-021-00657-6} & Graph-based & Series2Graph & I & U & \xmark \\

\hline
IForest \cite{LiuEtAl2008Isolation}                 & Tree-based & IForest & M & U & \xmark \\
IF-LOF \cite{ChengEtAl2019Outlier}                  & Tree-based & IForest/LOF & M & U & \xmark \\
Extended IForest \cite{HaririEtAl2019Extended}      & Tree-based & IForest & M & U & \xmark \\
Hybrid IForest \cite{MarteauEtAl2017Hybrid}         & Tree-based & IForest & M & Se & \xmark \\

\hline
SurpriseEncode \cite{ChakrabartiEtAl1998Mining}     & Encoding-based & - & M & U & \xmark \\
GranmmarViz \cite{SeninEtAl2015Time}                & Encoding-based & - & I & U & \xmark \\
Ensemble GI \cite{GaoEtAl2020Ensemble}              & Encoding-based & - & I & U & \xmark \\
PST \cite{SunEtAl2006Mining}                        & Encoding-based & Markov Ch. & M & U & \xmark \\
EM-HMM \cite{ParkEtAl2016Multimodal}                & Encoding-based & Markov Ch. & M & Se & \cmark \\
LaserDBN \cite{OgbechieEtAl2017Dynamic}             & Encoding-based & Bayseian Net. & M & Se & \xmark \\
EDBN \cite{PauwelsCalders2019anomaly}               & Encoding-based & Bayseian Net. & M & Se & \xmark \\
KDE-EDBN \cite{PauwelsCalders2019Detecting}         & Encoding-based & Bayseian Net. & M & Se & \xmark \\
PCA \cite{pcal_2003}                                & Encoding-based & PCA & M & Se & \xmark \\
RobustPCA \cite{PaffenrothEtAl2018Robust}           & Encoding-based & PCA & M & U & \xmark \\
DeepPCA \cite{deepPCA}                              & Encoding-based & PCA & M & Se & \xmark \\
POLY \cite{10.1016/j.peva.2010.08.018}              & Encoding-based & - & I & U & \xmark \\
SSA \cite{10.1016/j.peva.2010.08.018}               & Encoding-based & - & I & U & \xmark \\

\toprule
\end{tabular}
    \begin{tablenotes}
      \scriptsize
      \centering
      \item  I: Univariate; M: Multivariate // S: Supervised; Se: Semi-Supervised U: Unsupervised
     \end{tablenotes}
\label{density-based_table}
\end{table}%

\subsection{Distribution-based Methods}

The first category of density-based approaches is distribution-based anomaly detection approaches. 
Distribution-based methods involve building a distribution from statistical features of the points or sub-sequences of the time series. 
By examining the distributions of features of the normal sub-sequences, the distribution-based approach tries to recover relevant statistical models. It uses them to infer the abnormality of the data. In the following sections, we describe important anomaly detection methods belonging to this category.

\subsubsection{Minimum Covariance Determinant}

The Minimum Covariance Determinant (MCD) is a common distribution-based statistic in use~\cite{MCDfirst}. The algorithm seeks to find a subset (of a given size $h$) of all the sequences to estimate $\mu$ (mean of the subset) and $S$ (covariance matrix of the subset) with minimal determinant. In other words, the objective is to find the subset that is the least likely to include anomalies. Once the estimation is done, Mahalanobis distance is utilized to calculate the distance from sub-sequences to the mean, which is regarded as the anomaly score. 

FAST-MCD~\cite{Fast_MCD_1999} is a faster version of the MCD algorithm. Within small datasets, the result of the FAST-MCD algorithm usually aligns with that of the exact MCD. In contrast, faster and more accurate results are derived through the FAST-MCD rather than the classical MCD for large time series. Finally, MC-MCD~\cite{HARDIN2004625}, an extension of MCD, has been proposed for the multi-cluster setting.

\subsubsection{One-Class SVM}

One-Class Support Vector Machine (OCSVM) is a typical distribution-based example, which aims to separate the instances from an origin and maximize the distance from the hyperplane separation \cite{NIPS1999_1723} or spherical separation \cite{tax2004support}. The anomalies are identified with points of high decision score, i.e., far away from the separation hyper-plane. This method is a variant of the classical Support Vector Machine for classification tasks \cite{708428}.
Mathematically, given $\ell$-dimensional training data points $x_0,... x_n \in \mathcal{X}$, a. non-linear function $\phi$ that map the feature space $\mathcal{X}$ to a dot product space $F$, a kernel $k(x,y) = (\phi(x),\phi(y))$ (usually set to a Gaussian kernel $k(x,y) = e^{-||x-y||^2/c}$), the quadratic program to solve using a hyperplane is the following:

\begin{align*}
\min_{\omega \in F,\xi \in \mathbb{R},\rho \in \mathbb{R}} & \frac{1}{2}||w||^2 + \frac{1}{\nu \ell}\sum_{i} \xi_i - \rho \\
\text{subject to: } & (\omega.\phi(x_i)) \geq \rho - \xi_i, \\
  & \xi_i \geq 0.
\end{align*}

For a given new instance $x$, by deriving the dual problem, the decision function can be defined as follow:

\[
f(x) = sgn(\sum_{i} \alpha_i k(x_i,x) - \rho)
\]

\begin{figure}[t]
 \centering
 \includegraphics[scale=0.8]{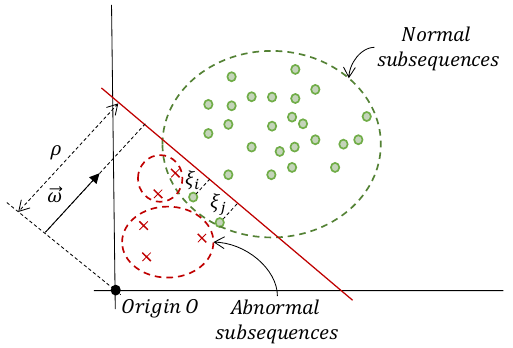}
 \caption{OCSVM illustration in which a point corresponds to a subsequence and only the green points are provided for the training step.}
 \label{fig:svm}
\end{figure}

Assuming that the optimization problem above can be solved, we can use such a decision function as an anomaly score. To be able to do it, one has to ensure to train the OCSVM model on a normal section of the time series only (those have to be annotated by a knowledge expert and therefore require extra work to be used). Figure~\ref{fig:svm} illustrates the anomaly detection process. It is also important to note that OCSVM is very similar to Support vector Data Description (SVDD) that has also been used for anomaly detection~\cite{XiaoEtAl2009Multisphere}.

In recent decades, an array of SVM variants have been applied in the time series setting. AOSVM~\cite{Gomez-VerdejoEtAl2011Adaptive} is an efficient streaming anomaly detection algorithm based on SVM that accommodates SVM to an online detection. The model is also adaptive, i.e., it forgets old data, featuring low computational and memory requirements. Eros-SVMs~\cite{LamriniEtAl2018Anomaly} is another variant of the SVM algorithm. It adopts a semi-supervised approach to train the model in the normal data. The algorithm then measures time windows' metrics as features fed into multiple OCSVMs, which are further selected based on the EROS similarity metric.
Moreover, multiple other methods based on OCSVM have been proposed in the literature. These methods either propose processing steps before applying OCSVM (such as NetworkSVM~\cite{ZhangEtAl2007One} or S-SVM~\cite{BhargavaRaghuvanshi2013Anomaly} that proposed seasonality decomposition or Stockwell transformation) or combine OCSVM with other methods (such as HMAD~\cite{GornitzEtAl2015Hidden} that uses hidden Markov chain to represent the time series into a latent space, on which OCSVM is applied).
Finally, DeepSVM~\cite{deepocsvm} proposes to use an Autoencoder architecture to extract meaningful features that use OCSVM on top of the learned latent space to detect anomalies.

\subsubsection{Histogram-based Outlier Score}

Histogram-based Outlier Score (HBOS)~\cite{Goldstein_histogram-basedoutlier} is another distribution-based algorithm for anomaly detection. For every single dimension (i.e., timestamps of a sub-sequence for univariate time series or values across multiple dimensions for multivariate time series), a univariate histogram is constructed with $k$ equal width bins. Each histogram is further normalized so that the height is 1. For a given multivariate point $p$ (or univariate sub-sequence $T_{i,\ell}$), we count the bin that contains $p$ for each dimension and multiply together the inverse of the frequency of bins where $p$ belongs for all dimensions. The algorithm assumes mutual Independence among the time series' dimensions (or the timestamps for univariate sub-sequence anomaly detection). Moreover, HBOS suits the particular case of highly unbalanced time series (i.e., very few anomalies) and unknown distribution.

\subsubsection{Extreme Studentized Deviate}

Various statistics, such as Extreme Studentized Deviate (ESD), are useful for inferring time series abnormality. The latter computes the statistical test for $k$ extreme values by $C_k = \max_k|x_k - \bar{x}|/s$, where $\bar{x}$ and $s$ denote the mean and the variance. The test is then compared with other critical values to determine if a value is abnormal. If so, then the value is removed, and the statistical test is re-calculated iteratively. Built on ESD, the S-ESD and SH-ESD~\cite{HochenbaumEtAl2017Automatic} methods remove the seasonal component using Seasonal-Trend decomposition (STL) and subtract the robust median. The pure, normalized data is then applied with ESD to detect anomalies. SH-ESD+~\cite{VieiraEtAl2018Enhanced}, on a further step than SH-ESD, applies the STL decomposition using a Loess regression and then generalizes the ESD on the residual part of seasonal decomposition to detect anomalies.

\subsubsection{Other Distribution-based Methods}

Besides the distribution-based algorithms presented above, many other methods are proposed using different models. First, the MedianMethod~\cite{BasuMeckesheimer2007Automatic} is a simple anomaly detection method proposed initially to filter outliers. The main idea is to measure the deviation from the median of the previous points and the median of their successive differences.
Moreover, SmartSifter~\cite{YamanishiEtAl2004OnLIne} aims to build a histogram for categorical values and a finite mixture model for continuous data. The proposed method aims to update those histograms/density as new points arrive in an unsupervised manner and then compute a score that estimates how the new point updated has changed the histogram/density. 
Then, OC-KFD~\cite{Roth2006Kernel} utilizes linear quantile models. The model is selected via cross-validated likelihood, from which a linear quantile model is fitted, and outliers are detected by considering confidence intervals. 
Moreover, MGDD~\cite{SubramaniamEtAl2006Online} utilizes kernel density estimation on sliding windows of the original time series. The estimated distribution is then used to identify the anomalies. 
Then, COPOD~\cite{li2020copod} is a copula-based anomaly detection method and an ideal choice for multivariate data.
Finally, unlike the previous models, ConInd~\cite{AntoniBorghesani2019statistical} is an algorithm based on domain knowledge. The model can detect only the known anomalies, where multiple statistical anomaly indicators (condition indicators) are proposed based on different distribution assumptions. 

\subsection{Graph-based Methods}

Then, graph-based methods are another category of density-based approaches. Graph-based methods represent the time series and the corresponding sub-sequences as a graph. The nodes and edges represent the different types of sub-sequences (or representative features) and their evolution in time. In this section, we describe the most important approaches in this category.

\subsubsection{Finite State Machine}

Finally, Finite State Machine (FSM) is a general categorization of machine learning algorithms that can be only in exactly one of a finite number of states at any given time. In reaction to any inputs, the FSM will shift from one state to another; such changes between states are called a transition. In anomaly detection, input time series will, upon certain machine-learned rules, change the state of the algorithms. If the state turns into an anomaly, the input is identified as anomalous. 

\begin{figure}[t]
	\centering
	\includegraphics[scale=0.6]{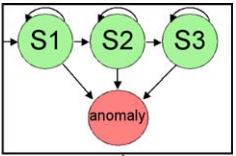}
	\caption{Illustration of the Finite State Machine (FSM).}
	\label{FSM}
\end{figure}

Figure~\ref{FSM} gives an illustration of such a process. In many ways, FSM is similar to a dynamic Bayesian network, which also uses Directed Acyclic Graph (DAG). However, the transition rule between FSM states is usually parametric, and thus the entire process is machine learning based. However, Finite State Machine is a general categorization where particular methods might be vastly different, each of which is unique to its own specific rules of the learning algorithm.

TwoFingers~\cite{Marceau2000Characterizing}, for example, builds a database of normal behavior by constructing a suffix tree for variable-length N-grams from the training data. The trees are transformed within the finite state machine and are further compacted to a DAG. Finally, the Finite State Machine, endowed with the architecture of DAG, matches the new series to detect anomalies. GeckFSM~\cite{SalvadorChan2005Learning}, however, is vastly different from TwoFingers, despite also following a finite-state machine structure. The proposed approach, GeckoFSM, aims to cluster the points (based on their slope) in the univariate time series and then extract some non-overlapping sub-sequences. A slope-based cluster merging operation then finds an optimal number of clusters, where transition human-readable rules of FSM for each cluster are further computed using the RIPPER algorithm~\cite{COHEN1995115}. Anomalies are identified as points that derive significantly from these rules.

\subsubsection{Graph Representation of Sub-sequences}

A second approach is to convert the time series into a directed graph with nodes representing the usual types of subsequences and edges representing the frequency of the transitions between types of subsequences. Series2Graph~\cite{BoniolPalpanas2020Series2Graph} is building such kinds of graphs. Moreover, an extension of Series2Graph proposed in the literature, named DADS~\cite{10.1007/s00778-021-00657-6}, proposes a distributed implementation and, therefore, a much more scalable method for large time series.

\begin{figure}[t]
 \centering
 \includegraphics[scale=1.1]{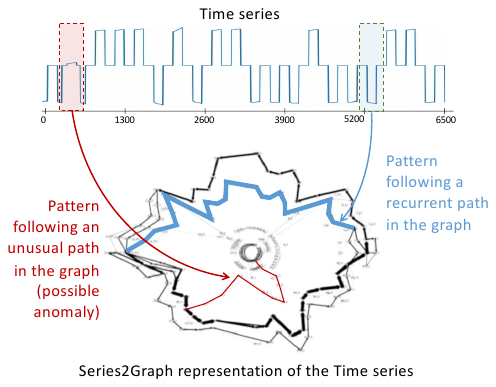}
 \caption{Example of Series2graph representation.}
 \label{fig:s2g}
\end{figure}

For a given data series $T$, the overall process of Series2Graph is divided into the following four main steps.

\begin{enumerate}
\item {\bf Subsequence Embedding}: Project all the 
subsequences (of a given length $\ell$) of $T$ in a two-dimensional space, where 
shape similarity is preserved. 
\item {\bf Node Creation}: Create a node for each one of the densest parts of the above two-dimensional space. These nodes can be seen as a summarization of all the major patterns of length $\ell$ that occurred in $T$. 
\item {\bf Edge Creation}: Retrieve all transitions between pairs of subsequences represented by two different nodes: each transition corresponds to a pair of subsequences, where one occurs immediately after the other in the input data series $T$. 
We represent transitions with an edge between the corresponding nodes. 
The weights of the edges are set to the number of times the corresponding pair of subsequences was observed in $T$.
\item {\bf Subsequence Scoring}: Compute the normality (or anomaly) score of a subsequence of length $\ell_q \geq \ell$ (within or outside of $T$), based on the previously computed edges/nodes and their weights/degrees.
\end{enumerate}

Figure~\ref{fig:s2g} depicts the resulting graph returned by Series2Graph. The unusual path in the graph (with edges with low weights and nodes with low degrees) corresponds to anomalies in the time series.

\subsection{Tree-based Methods}

Instead of modeling the time series into a graph, the different points or sub-sequences can also be organized in trees to highlight potential isolated instances that could correspond to anomalies.
Isolation Forest (IForest) is density-based and the most famous tree-based approach for anomaly detection. IForest tries to isolate the outlier from the rest \cite{Liu:2008:IF:1510528.1511387}.
The key idea remains on the fact that, in a normal distribution, anomalies are more likely to be isolated (i.e., requiring fewer random partitions to be isolated) than normal instances. 
If we assume the latter statement, we only have to produce a partitioning process that indicates well the isolation degree (i.e., anomalous degree) of instances.

\begin{figure}[t]
 \centering
 \includegraphics[scale=0.20]{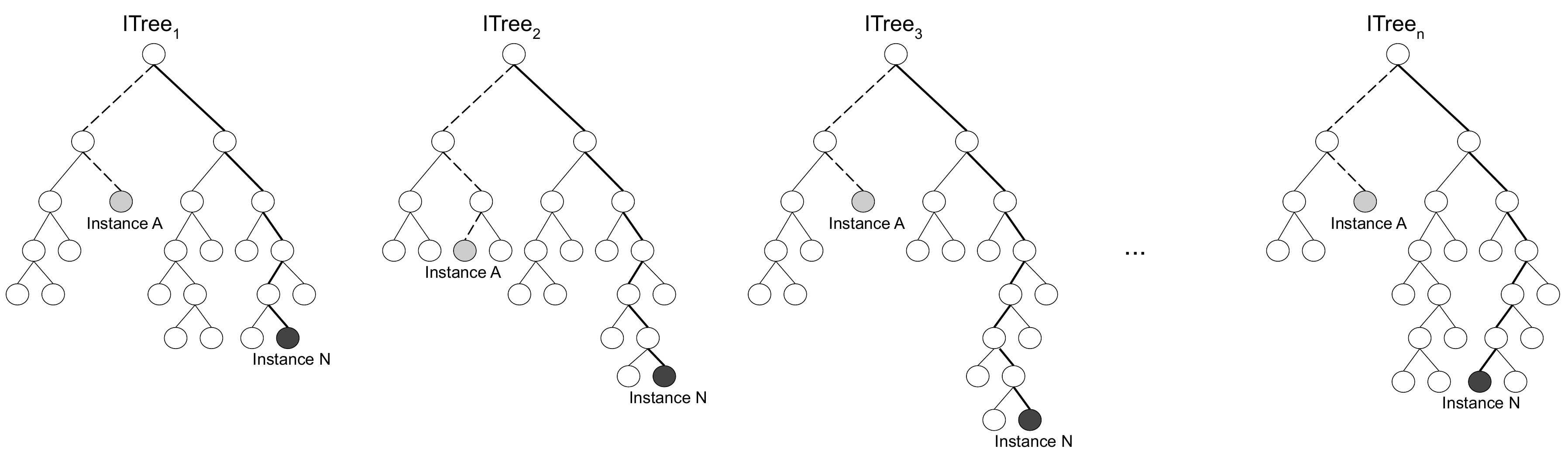}
 \caption{Set of isolation trees that randomly partition a dataset. On average, instance N has a longer path to the root than instance A. Thus, instance A's anomaly score will be higher.}
 \label{fig:isolation_forest_set}
\end{figure}

Let first define the concept of Isolation Tree as stated in \cite{Liu:2008:IF:1510528.1511387}. 
Let be $Tr$ a binary tree where each node has zero or two children and a test that consists of an attribute $q$ and a split $p$ such that $p<q$ divides data points into the two children. $Tr$ is built by dividing recursively the training dataset $T = \{T_{i,\ell},T_{i+1,\ell},...,T_{|T| - \ell,\ell}\}$ randomly selecting $p$ and $q$ until, the maximal depth of the tree is reached, or the number of different instances is equal to 1. Figure \ref{fig:isolation_forest_set} depicts an example of isolation trees.

Using that kind of data structure, the path length into the tree $Tr$ to reach an instance $T_{i,\ell}$ is directly correlated to the anomaly degree of that instance. Therefore, we can define the anomaly score as follow:

\begin{equation}
S(x,n) = 2^{s}
s = \frac{\sum_{Tr \in \mathcal{T}} h(x,Tr) }{|\mathcal{T}|c(n)}, c(n) = H(n-1) - \frac{2(n-1)}{n}
\end{equation}

With $h(x,Tr)$ the length of the path to reach $x$ in the tree $Tr$, $\mathcal{T}$ a set of different isolation trees built, $n$ the number of instances in the training set, and H is the harmonic number. It can be simply but surely estimated using the Euler constant.

Other IForest-based algorithms have also been proposed recently. Extended IForest~\cite{HaririEtAl2019Extended} is an extension of the traditional Isolation Forest algorithm, which removes the branching bias using hyperplanes with random slopes. The random sloping hyperplanes enable an unbiased selection of features free of the branching structure within the dataset. Hybrid IForest~\cite{MarteauEtAl2017Hybrid} is another improvement on IForest, enabling a supervised setting and eliminating the dataset's potential confounding due to unbalanced clusters. 
Finally, IF-LOF~\cite{ChengEtAl2019Outlier} combines IForest and LOF by applying IForest and then utilizes LOF to refine the results, which speeds up the process.

\subsection{Encoding-based Methods}

Encoding-based methods represent the sub-sequences of a time series into a low-dimensional latent space or data structure. The anomaly score is directly from the latent space representations. More specifically, the anomaly scores are attributed to the points that correspond to the encoded sub-sequences in the latent space.

\subsubsection{Principal Component Analysis}

The first encoding-based approach is to encode and represent the time series with its principal components. Principal Components Analysis (PCA) investigates the major components of the time series that contribute the most to the covariance structure. The anomaly score is measured by the sub-sequences distance from $0$ along the principal components weighted by their eigenvalues. A standard routine is to pick $q$ significant components that can explain 50\% variations of the time series and $r$ minor components that explain less than 20\% variations. A point is an anomaly if its values of major principles components, $y_1, y_2...$, have a weighted sum exceeding the threshold its minor one has. So $x$ (or a sub-sequence $T_{i,\ell}$ of a given time series) is an anomaly if:

\begin{equation}
\sum_1^q \frac{y_i}{\lambda_i} > c_1 \text{ or,}  \sum_{p-r+1}^p \frac{y_i}{\lambda_i} > c_2 
\end{equation}

In the equation above, $c_1$ and $c_2$ are predefined threshold values, and $\lambda$s are the eigenvalues. RobustPCA~\cite{PaffenrothEtAl2018Robust} aims to recover the principal matrix $L_0$ by decomposing the original covariance matrix into $M = L_0 + S_0$ to minimize the rank of $L_0$. The residual term $S_0$ helps separate the anomalous subsets and makes the algorithm applicable to time series containing many anomalies. 
Finally, deepPCA~\cite{deepPCA} is similar to robustPCA but with an autoencoder preprocessing step first. The autoencoder maps the time series into a latent space, and then the PCA (described above) is used to identify anomalies.

\subsubsection{Grammar and Itemset Representations}

Another approach is to represent the time series into a set of symbols associated with rules. GrammarViz~\cite{DBLP:conf/edbt/Senin0WOGBCF15} adopts an approach to find anomalies based on the concept of Kolmogorov complexity where the randomness in a sequence is a function of its algorithmic incompressibility. The main idea is that it is possible to represent a time series as context-free grammar (a set of symbols associated with rules), and the sections of the time series that match a few grammar rules are considered anomalies. In addition, A feature of this algorithm is also centered on its ability to find anomalies of different lengths.

More precisely, the algorithm can be divided into different phases. First, the whole time series is summarized using Symbolic Aggregate Approximation (SAX) to have discrete values and not continuous ones. Next, context-free grammar is built using Sequitur, a linear space and time algorithm able to derive context-free grammar from a string incrementally. Finally, a rule density curve is built, which is the metadata that allows the detection of anomalies. It is possible to obtain a rule density curve by iterating over all grammar rules and incrementing a counter for each time series that points to the rule spans. Once the rule density curve is obtained, it is possible to discover anomalies by picking the minimum values of the curve. Otherwise, it is possible to discover the least frequent sub-sequences (and possible anomalies) by applying the Rare Rule Anomaly (RRA) algorithm.

Other grammar-based methods have been proposed in the literature. First, Ensemble GI~\cite{GaoEtAl2020Ensemble} is an extension of the GrammarViz algorithm, which further implements grammar induction on an ensemble approach that obtains the anomaly detection result based on the ensemble detection of models with different parameter values. Unlike the previous two, SupriseEncode~\cite{ChakrabartiEtAl1998Mining} adopts a distinct compression-based method representing the record as an itemset. The compression ratio of segments (code-length encoding) derived from each sub-sequence is compared among the training data set to derive the anomaly score.

\subsubsection{Hidden Markov Model}

Another type of Encoder-based method is Hidden Markov Model (HMM). The latter assumes the existence of a Markov process $X$ such that the time series data observed is dependent on that $X$. The goal is to derive $X$ by observing the data. The anomalies are detected by measuring the ability of the encoding to represent the time series. For instance, EM-HMM~\cite{ParkEtAl2016Multimodal} is a time-series anomaly detection method based on HMM.

More precisely, PST~\cite{SunEtAl2006Mining} is another detection method based on HMM. It proposes an efficient algorithm for computing the Probabilistic Suffix Tree (PST), a compact variable-order Markov chain. In practice, the algorithm embeds possible chains of values (and their probability) into the trees and infers the anomaly score by computing the probabilities of the chains of values.

\subsubsection{Bayesian Networks}

Bayesian Network builds a graph denoting the relationship between random variables in terms of Directed Acyclic Graph (DAG). Each node in the DAG stands for a random variable, and the edges represent the probabilistic relationships among the variables. Dynamic Bayesian Network, or temporal Bayesian Network, generalizes the Bayesian Network graph model to the time series setting. The model is capable of modeling the temporal relationship for different random variables with first-order assumption. 

\begin{figure}[H]
	\centering
	\includegraphics[scale=0.15]{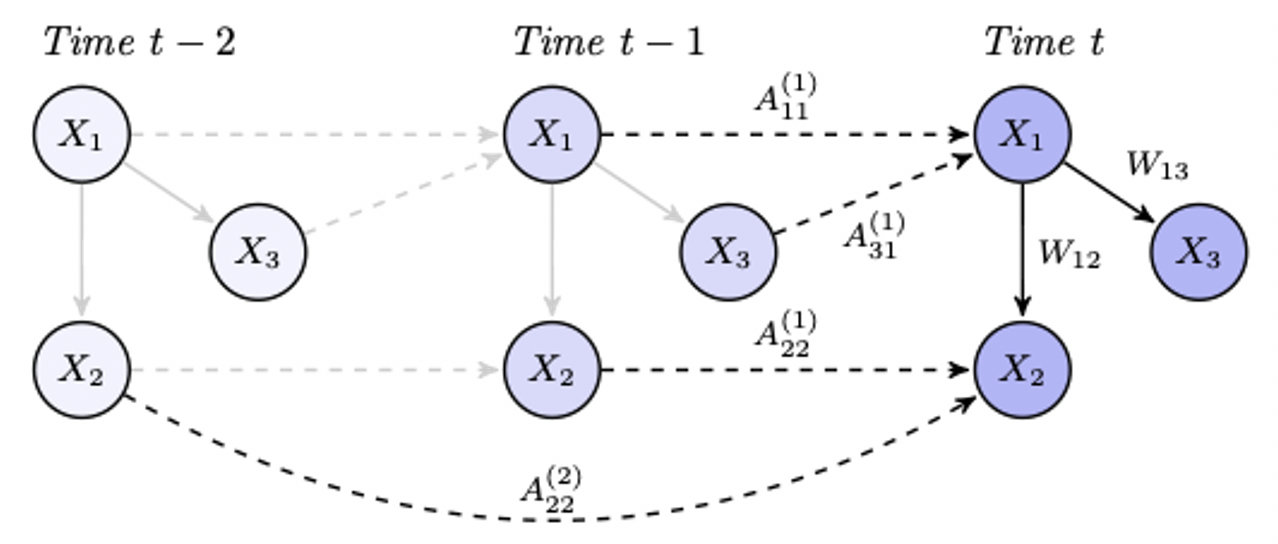}
	\caption{Illustration of Dynamic Bayesian Network (DBN).}
	\label{DBN}
\end{figure}

As displayed in Figure~\ref{DBN}, the random variables at different timestamps are connected by probabilistic edges, which are referenced by the model to characterize the temporal change of the random variables in the dataset. The joint probability distribution of a given DBN variable $X_{i}$ is given by the following equation:

\begin{equation}
    P(X_{i:t,1}, X_{i:t,2}, X_{i:t,R-1} ... X_{i:t,R}) = \prod_{t \leq T, r \leq R} P(X_{i:t,r}| Par(X_{i:t,r}))
\end{equation}

$Par(X_{i:t,r})$ denotes the parent of $X_{i:t,r}$, which are either inside the previous timestamps or just the parent inside the same timestamps, due to the first order assumption. 

Various time-series anomaly detection methods have implemented DBN in their algorithms. LaserDBN~\cite{OgbechieEtAl2017Dynamic} is a method proposed for image time series. The data is first preprocessed by k-means clustering to perform feature selection, and then Dynamic Bayesian Network is implemented to compute the probabilities of individual sub-sequences. EDBN~\cite{PauwelsCalders2019anomaly} proposes an extension of DBN to suit the particular case of textual time-series anomaly detection (specifically for business processes and logs). Finally, KDE-EDBN~\cite{PauwelsCalders2019Detecting} is an extension of EDBN that uses Kernel Density Estimation (KDE) to handle numerical attributes in logs.

\subsubsection{Other Encoding-based Methods}

In addition to all the methods described above, several more anomaly detection methods could be grouped in the encoding-based category, such as polynomial approximation methods to detect anomalies, like POLY~\cite{10.1007/978-3-540-71701-0_17} or SSA~\cite{10.1016/j.peva.2010.08.018}. The latter is training multiple polynomial approximation models for each time series (or sub-sequence in the time series). A similarity measure between the trained models is used to detect anomalies.

\section{Prediction-based Methods}

Prediction-based methods aim to detect anomalies by predicting the expected normal behaviors based on a training set of time series or sub-sequences (containing anomalies or not). For instance, some methods will be trained to predict the next value or sub-sequence based on the previous one. Then, the prediction error is used as an anomaly score. The underlying assumption of prediction-based methods is that normal data are easier to predict, while anomalies are unexpected, leading to higher prediction error. Such assumptions are valid when the training set contains no or few time series with anomalies. Therefore, prediction-based methods are usually more optimal under semi-supervised settings. 
Within the prediction-based methods, there come two second-level categories: forecasting-based and reconstruction-based. We enumerate all the mentioned methods in Table~\ref{prediction-based_table}.

\begin{table}[ht]
\footnotesize

\caption{Summary of the prediction-based anomaly detection methods.}
\vspace{-0.3cm}
\begin{tabular}[t]{lccccc}
\toprule
&Second Level &Prototype  &Dim &Method & Stream\\

\toprule
ES \cite{snyder_exponential_1983} & Forecasting-based & - & I & Se & \xmark \\
DES \cite{snyder_exponential_1983} & Forecasting-based & - & I & Se & \xmark \\
TES \cite{snyder_exponential_1983} & Forecasting-based & - & I & U & \xmark \\
ARIMA \cite{rousseeuw_robust_1987} & Forecasting-based & ARIMA & I & U & \cmark \\
NoveltySVR \cite{MaPerkins2003Online} & Forecasting-based & SVM & I & U & \cmark\\
PCI \cite{YuEtAl2014Time} & Forecasting-based & ARIMA & I & U & \cmark \\
OceanWNN \cite{WangEtAl2019Study} & Forecasting-based & - & I & Se & \xmark \\
MTAD-GAT \cite{ZhaoEtAl2020Multivariate} & Forecasting-based & GRU & M & Se & \cmark \\
AD-LTI \cite{WuEtAl2020Developing} & Forecasting-based & GRU & M & Se & \cmark \\

CoalESN \cite{ObstEtAl2008Using} & Forecasting-based & ESN & M & Se & \cmark \\
MoteESN \cite{ChangEtAl2009MoteBased} & Forecasting-based & ESN & I & Se & \cmark \\
HealthESN \cite{ChenEtAl2020Imbalanced} & Forecasting-based & ESN & I & Se & \xmark \\
Torsk \cite{HeimAvery2019Adaptive}& Forecasting-based & ESN & M & U & \cmark \\

LSTM-AD \cite{MalhotraEtAl2015Long} & Forecasting-based & LSTM & M & Se & \xmark \\
DeepLSTM \cite{ChauhanVig2015Anomaly} & Forecasting-based & LSTM & I & Se & \xmark \\
DeepAnT \cite{MunirEtAl2019DeepAnT} & Forecasting-based & LSTM & M & Se & \xmark \\
Telemanom$\star$ \cite{HundmanEtAl2018Detecting} & Forecasting-based & LSTM & M & Se & \xmark \\
RePAD \cite{LeeEtAl2020RePAD} & Forecasting-based & LSTM & M & U & \xmark \\

NumentaHTM \cite{AhmadEtAl2017Unsupervised} & Forecasting-based & HTM & I & U & \cmark \\
MultiHTM \cite{WuEtAl2018Hierarchical} & Forecasting-based & HTM & M & U & \cmark \\
RADM \cite{DingEtAl2018MultivariateTimeSeriesDriven} & Forecasting-based & HTM & M & Se & \cmark \\

\hline
MAD-GAN \cite{LiEtAl2019MADGAN} & Reconstruction-based & GAN & M & Se & \cmark \\
VAE-GAN \cite{NiuEtAl2020LSTMBased} & Reconstruction-based & GAN & M & Se & \xmark \\
TAnoGAN \cite{BasharNayak2020TAnoGAN} & Reconstruction-based & GAN & M & Se & \xmark \\
USAD \cite{10.1145/3394486.3403392} & Reconstruction-based & GAN & M & Se & \xmark \\
EncDec-AD \cite{MalhotraEtAl2016LSTMbased} & Reconstruction-based & AE & M & Se & \xmark\\
LSTM-VAE \cite{ParkEtAl2018Multimodal} & Reconstruction-based & AE & M & Se & \cmark\\
DONUT \cite{XuEtAl2018Unsupervised} & Reconstruction-based & AE & I & Se & \xmark\\
BAGEL \cite{LiEtAl2018Robust} & Reconstruction-based & AE & I & Se & \xmark\\
OmniAnomaly \cite{SuEtAl2019Robust} & Reconstruction-based & AE & M & Se & \xmark\\
MSCRED \cite{ZhangEtAl2019Deep} & Reconstruction-based & AE & I & U & \xmark \\
VELC \cite{ZhangEtAl2020VELC} & Reconstruction-based & AE & I & Se & \xmark\\
CAE \cite{cae2000,GarciaEtAl2020Time} & Reconstruction-based & AE & I & Se & \xmark\\
DeepNAP \cite{KimEtAl2018DeepNAP} & Reconstruction-based & AE & M & Se & \cmark \\
STORN \cite{SoelchEtAl2016Variational} & Reconstruction-based & AE & M & Se & \cmark \\

\toprule
\end{tabular}
    \begin{tablenotes}
      \scriptsize
      \centering
      \item  I: Univariate; M: Multivariate // S: Supervised; Se: Semi-Supervised U: Unsupervised
     \end{tablenotes}
\label{prediction-based_table}
\end{table}%

\subsection{Forecasting-based Methods}

Forecasting-based methods use a model trained to forecast several time steps based on previous points or sequences. The forecasting results are thus directly connected to previous observations in the time series. The forecasted points or sequences are then compared to the original ones to determine how anomalous or unusual these original points are. 

\subsubsection{Exponential Smoothing}
One of the first forecasting-based approaches proposed in the literature is the Exponential Smoothing~\cite{snyder_exponential_1983}. The latter is a non-linear smoothing technique to predict the time series points by taking the previous time series data and assigning exponential weights to these previous individual observations. The anomalies are then detected by comparing the predicted and actual results. Formally, the prediction of the current value $\hat{T_i}$ is defined as follows:

\begin{equation}
   \hat{T_i} = (1-\alpha)^{N-1} T_{i-N} + \sum_{j=1}^{N-1} \alpha(1-\alpha)^{j-1} T_{i-j} \: \: \: \alpha \in [0,1] 
\end{equation}

Thus, the estimated sub-sequence is a linear combination of the previous data points, with the weights varying exponentially. The parameter $\alpha$ stands for the rate of exponential decrease. The smaller the $\alpha$ is, the more the weight is assigned to the distant data points.

In addition, several approaches based on exponential smoothing have been proposed. For example, Double Exponential Smoothing (DES) and Triple Exponential Smoothing (TES)~\cite{snyder_exponential_1983} are variants of the exponential smoothing techniques for non-stationary time series. In DES, a further parameter $\beta$ is utilized to smooth the trend that a time series can have. For the special case of time series containing seasonality, TES enables another parameter $\gamma$ to control it.

\subsubsection{ARIMA}
Another early category of forecasting-based approaches proposed in the literature is ARIMA models~\cite{rousseeuw_robust_1987}. The latter assumes a linear correlation among the time series data. The algorithm fits the ARIMA model on the time series and draws anomalies by comparing the prediction of the ARIMA model and real data. Formally, An $\mathbf{ARIMA}(p^{\prime}, q)$ model is built upon the following iterative equations: 

\begin{equation}
    T_i = \sum_{k=1}^{p^\prime} \alpha_k T_{t-k} + \epsilon_i + \sum_{j=1}^{q} \theta_j^p \epsilon_{i-j}
\end{equation}

Overall, using ARIMA models, we assume that every next time series values correspond to a linear combination of the previous values and residuals. Note that the residuals must be estimated in an iterated manner. 
Moreover, Prediction Confidence Interval (PCI)~\cite{YuEtAl2014Time} is an extension of the ARIMA model, which further combines the nearest neighbor method. The prediction confidence interval allows dynamic thresholding. The threshold can be estimated on the historical nearest neighbors.

\subsubsection{Long Short-Term Memory}

Long Short-Term Memory (LSTM)~\cite{Hochreiter:1997:LSM:1246443.1246450} network has been demonstrated to be particularly efficient in learning inner features for sub-sequences classification or time series forecasting. Such a model can also be used for anomaly detection purposes~\cite {LSTManomaly,LSTM_ON_GHL}. 
The two latter papers' principle is as follows: A stacked LSTM model is trained on {\it normal} parts of the data that we call $N$. The objective is to predict the point $N_i \in N$ or the sub-sequence $N_{i,l_1}$ using the sub-sequence $N_{i-l,l}$. Consequently, the model will be trained to forecast a healthy state of the time series, and, therefore, will fail to forecast when it will encounter an anomaly.  

\begin{figure}[t]
	\centering
	\includegraphics[scale=0.45]{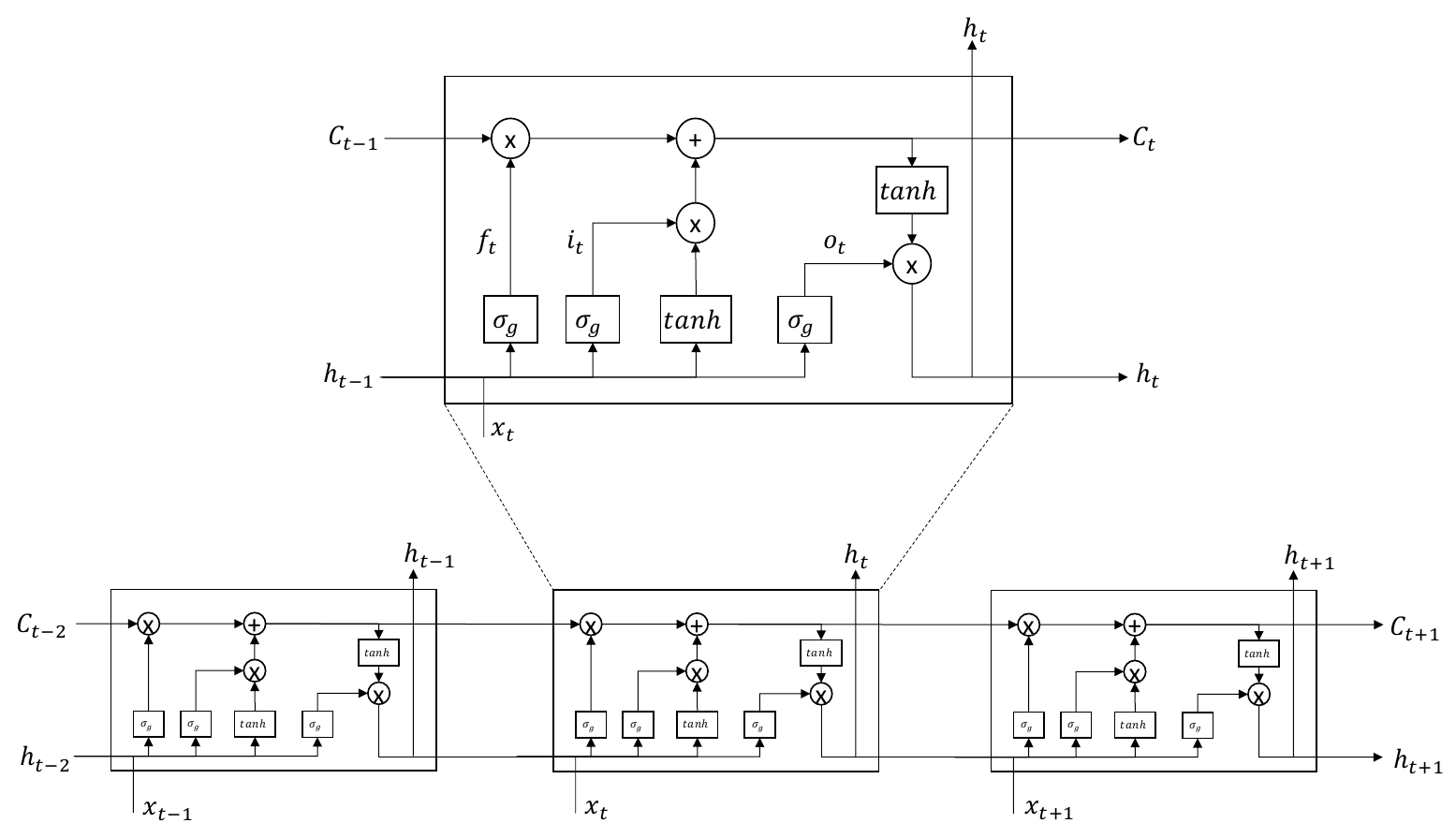}
	\caption{LSTM cell architecture. 
 }
	\label{fig:LSTMcell}
\end{figure}

LSTM network is a special type of Recurrent Neural Network (RNN), based on LSTM units as memory cells to encode hidden information. Figure~\ref{fig:LSTMcell} depicts the components and interactions within an LSTM cell. The various components are given by:

\begin{align*}
f_t &= \sigma_g(W_fx_t + U_fh_{t-1} + b_f) \\
i_t &= \sigma_g(W_ix_t + U_ih_{t-1} + b_i) \\
o_t &= \sigma_g(W_0x_t + U_0h_{t-1} + b_0) \\
c_t &= f_t \circ c_{t-1} + i_t \circ \sigma_c(W_c x_t + U_c h_{t-1} + b_c) \\
h_t &= o_t \circ \sigma_h(c_t) 
\end{align*}

By combining a large number of cell (outlined in Figure \ref{fig:LSTMcell}) and stacking them~\cite{LSTManomaly}, one can fit the weights to forecast the time series in two different ways described as follow: (i) The first is to train the network using a fixed time window length $T_{t-\ell-1,\ell} = [T_{t-\ell}, ..., T_{t-1}]$ to predict $T_t$, (ii) or using the same input to predict the incoming sequence $T_{t,\ell'} = [T_t, ..., T_{t+\ell'}]$. For the specific purpose of anomaly detection, we will assume that such a model can be trained to achieve both of the previously enumerated tasks. Then, what has to be done is to train this model on the normal section of the time series (apriori annotated by the knowledge expert) and use the forecasting error as an anomaly score. Therefore, one can expect to obtain a bigger forecasting error for a sub-sequence that the model has never seen (like an anomaly), rather than a usual sub-sequence.

There exists a large variety of methods based on LSTM neural networks proposed in the literature. First, DeepLSTM~\cite{ChauhanVig2015Anomaly} is a standard implementation of LSTM networks. The generative model stacks the LSTM network trained from normal sections of the time series. Then, LSTM-AD~\cite{MalhotraEtAl2015Long} adopts a similar approach to DeepLSTM. In addition to training the LSTM model to predict time series, LSTM-AD also estimates the training dataset's errors with multivariate normal distribution and calculates the anomaly score with the Mahalanobis distance.

Moreover, Telemanom~\cite{HundmanEtAl2018Detecting} is an LSTM-based approach that focuses on channeled data (i.e., multivariate time series). An LSTM network is trained for each channel. The prediction error is further smoothed over time, and low errors are pruned retroactively. Then, RePad~\cite{LeeEtAl2020RePAD} is another LSTM-based algorithm that considers short-term historical data points to predict future anomalies in streaming data. The detection is based on the Average Absolute Relative Error (AARE) of LSTM, and RePad also implements a dynamic threshold adjustment to vary the threshold value at different timestamps.

\subsubsection{Gated Recurrent Unit}

In addition to LSTM cells frequently implemented in time series settings, other neural network architectures have also been in use. One example is the Gated Recurrent Unit (GRU) which is also an RNN but operates in a different gated unit than LSTM to forecast time series values. We will summarize some of the approaches used in these different architectures.

MTAD-GAT~\cite{ZhaoEtAl2020Multivariate} is the first example of anomaly detection methods based on GRU units. The latter uses both the prediction error and reconstruction error for the detection of anomalies (This method could fit in both forecasting and reconstruction-based categories).
The model utilizes two parallel graph attention layers to preprocess the time series dataset and then implements a GRU network to reconstruct and predict the next values.
AD-ITL~\cite{WuEtAl2020Developing} is another GRU-based algorithm with seasonal and raw features as input. The input time series is first used to extract seasonal features and further fed to the GRU network. The GRU then predicts each value of the window, and Local Trend Inconsistency (LTI) is used as a measure of the error to assess the abnormality between predicted and actual values.

\subsubsection{Echo State Networks}

Researchers have also proposed multiple Echo State Networks for detecting anomalies in time series. An Echo State Network (ESN) is a variant of RNN, which has a sparsely connected random hidden layer. The model randomizes the weights in hidden and input layers and also connects neurons randomly. Only the values in the output layers are learned, rendering the method a linear model that is easily trained. The random hidden layers act as a dynamic reservoir that transforms the input into sequences of non-linear, complicated processes. The trainable output layer organizes the encoding of the inputs in the dynamic reservoir, enabling complex representation of the data despite its linearity. The initial values of input and hidden layers are also chosen carefully, usually tuned with multiple parameters.

First, CoalESN~\cite{ObstEtAl2008Using} is a simple implementation of Echo State by predicting time series values and comparing the estimated results with real ones to determine abnormality. MoteESN~\cite{ChangEtAl2009MoteBased} adopts a similar approach to CoalESN but uses the absolute difference to measure the anomaly score. The model is optimized for a sensor device, where the network is trained offline before deployment on the sensor. Torsk~\cite{HeimAvery2019Adaptive} is another adaptation of ESN. Like its precursors, Torsk uses the previous window as training data and then predicts the following ones. The model further implements automatic thresholding. Finally, HealthESN~\cite{ChenEtAl2020Imbalanced} is an Echo State Network applied to the medical and health domain. The algorithm utilizes the default architecture with training and testing steps; after a sequence of data preprocessing, intelligent threshold computation is used to estimate the adaptive threshold and declare anomalies by the ESN predictions. 

\subsubsection{Hierarchical Temporal Memory}

Another recurrent neural network type of approach is Hierarchical Temporal Memory (HTM). The latter is the core component of multiple anomaly detection methods proposed in the literature. 
The HTM method is based on the theory and ideas proposed in the Thousand Brains Theory of Intelligence~\cite{Athousandbrains}. 
The latter proposes that many models are learned for each object or concept, rather than only one single model per object, as most of the methods described in the previous sections usually handle.

In particular, HTM focuses on three main tasks: sequence learning, continual learning, and sparse distributed representations.
Even though HTM-based methods can be seen as RNN-based methods (such as LSTM, GRU, and ESN-based approach), the main difference is between the neuron definition and the learning process. For HTM, the unsupervised Hebbian-learning rule~\cite{hebb-organization-of-behavior-1949} is applied to train the model rather than the classical back-propagation is not applied. 

The first time-series anomaly detection method using HTM proposed in the literature is the NumentaHTM~\cite{AhmadEtAl2017Unsupervised} and MultiHTM~\cite{WuEtAl2018Hierarchical} approaches. Moreover, RADM~\cite{DingEtAl2018MultivariateTimeSeriesDriven} combines HTM with Naive Bayesian Networks to detect anomalies in multivariate time series.

\subsubsection{Other Forecasting-based Methods}

Finally, it is important to note that forecasting-based approaches are a generic concept that requires to have a model that can predict the incoming value from historical data. Therefore, any regression approach can be used as a forecasting-based approach. In the previous sections, we described on a high-level the most popular methods used to perform anomaly detection using forecasting-based techniques. We can complement the list with methods using more specific architecture on specific applications such as OceanWNN~\cite{WangEtAl2019Study} using Wavelet-Neural Networks, or more classical regression techniques used as forecasting-based core units such as NoveltySVR~\cite{MaPerkins2003Online} using Support-Vector-Machine (SVM).

\subsection{Reconstruction-based Methods}

Reconstruction-based methods represent normal behavior by encoding sub-sequences of a normal training time series into a low-dimensional space. The sub-sequences are then reconstructed from the low-dimensional space, and the reconstructed sub-sequences are then compared to the original sub-sequences. The difference between the reconstruction and the original sequence is used to detect anomalies. In general, the inputs to the reconstruction process are training sub-sequences.

\subsubsection{Autoencoder}

\begin{figure}[t]
 \centering
 \includegraphics[scale=1]{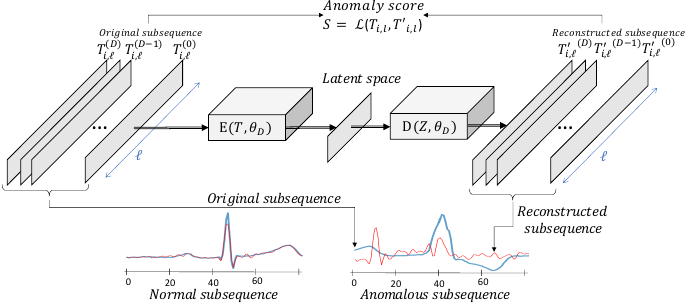}
 \caption{Overview of autoencoders methods for time-series anomaly detection.}
 \label{fig:ae}
\end{figure}

Autoencoder is a type of artificial neural network used to learn to reconstruct the dataset given as input using a smaller encoding size to avoid identity reconstruction. As a general idea, the autoencoder will try to learn the best latent representation (also called encoding) using a reconstruction loss. Therefore, it will learn to compress the dataset into a shorter code and then uncompress it into a dataset that closely matches the original. Figure~\ref{fig:ae} depicts an overview of autoencoders for time series. Formally, given two transition functions $E$ and $D$, respectively called encoder and decoder, the task of an autoencoder is the following one:

\begin{align}
\phi: &\mathbb{T}_{\ell} \rightarrow \mathcal{Z} \\
\psi: &\mathcal{Z} \rightarrow \mathbb{T}_{\ell} \\
\phi,\psi = &arg \min_{\phi,\psi} \mathcal{L} (T_{i,\ell}, \psi(\phi(T_{i,\ell})) )
\end{align}

$\mathcal{L}$ is a loss function that is usually set to the mean square error of the input and its reconstruction, formally written $|| X - \psi(\phi(T_{i,\ell})) ||^2$. This loss fits the task well for the specific case of sub-sequences in a time series since it coincides with the Euclidian distance. 

The reconstruction error can be used as an anomalous score for the specific anomaly detection task. As the model is trained on the non-anomalous sub-sequence of the time series, it is optimized to reconstruct the normal sub-sequences. Therefore, all the sub-sequences far from the training set will have a bigger reconstruction error. 

As autoencoder has been a popular method in the recent decade, many anomaly detection algorithms are based on autoencoder algorithms' implementation. EncDec-AD~\cite{MalhotraEtAl2016LSTMbased} is the first model that implements an encoder-decoder by using reconstruction error to score anomalies. LSTM-VAE~\cite{ParkEtAl2018Multimodal} and MSCRED~\cite{ZhangEtAl2019Deep} use LSTM and Convolutional LSTM cells in the autoencoder architecture. Similarly, Omnianomaly~\cite{SuEtAl2019Robust} is another autoencoder method where the autoencoder architecture uses GRU and planar normalizing flow. 

Then, STORN~\cite{SoelchEtAl2016Variational} and DONUT~\cite{XuEtAl2018Unsupervised} proposed a Varational Autoencoder (VAE) method to detect abnormal sub-sequences. For DONUT, it further preprocesses the time series using the MCMC-based missing data imputation technique~\cite{10.5555/3044805.3045035}. Improving from DONUT, BAGEL~\cite{LiEtAl2018Robust} implements conditional VAE instead of VAE. VELC~\cite{ZhangEtAl2020VELC} sets up additional constraints to the VAE. The Decoder phase is regularized due to anomalies in training data, which helps fit normal data and prevent generalization to model abnormalities. 

Moreover, CAE~\cite{cae2000,GarciaEtAl2020Time} uses a convolutional autoencoder to convert time series sub-sequences to image encoding. The algorithm also speeds up nested-loop-search using sub-sequences approximation with SAX word embedding.

Finally, DeepNAP~\cite{KimEtAl2018DeepNAP} is a sequence-to-sequence AE-based model. However, unlike other AE-based models, DeepNAP detects anomalies before they occur.

\subsubsection{Generative Adversarial Networks}

\begin{figure}[t]
 \centering
 \includegraphics[scale=1]{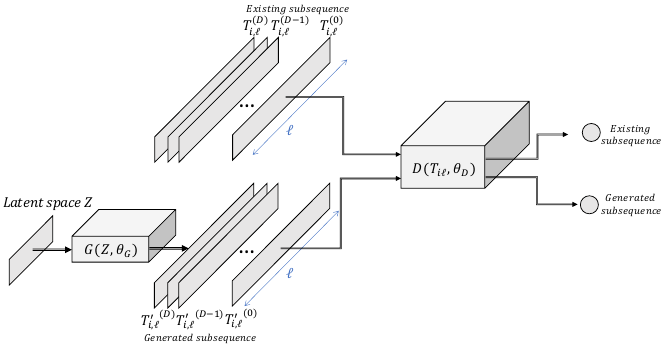}
 \caption{Overview of GAN methods for time-series anomaly detection.}
 \label{fig:gan}
\end{figure}

Generative Adversarial Network (GAN) is initially proposed for image generation purposes \cite{NIPS2014_5423} but can also be used to generate time series. GAN has two components: (i) one to generate new time series and (ii) one to discriminate the existing time series. Both of the components are useful for the detection of anomalies. Figure~\ref{fig:gan} depicts the overview of GAN methods for anomaly time series.

More precisely, a GAN is composed of two networks. The first is called the generator $G(z,\theta_g)$ with $\theta_g$ its parameters. The second one is called the discriminant $G(x,\theta_d)$ with $\theta_d$ its parameters. The output of $G$ is the same shape as the input, and the output of $D$ is a scalar $D(x)$ that represents the probability that $x$ came from the original dataset. Therefore $1-D(x)$ is the probability that $G$ has generated $x$. Formally, $G$ and $D$ have to be optimized, such as the two-player optimization problem where the accuracy of the discriminator has to be maximized but also minimized regarding the generator. 
The value to be minimized, denoted as $V(G,D)$, is defined in the following manner.

\begin{align}
\min_G \max_D V(G,D) = \mathbb{E}_{x \sim p_{data}(x)} [log D(x)] + \mathbb{E}_{z \sim p_{z}(z)} [log(1- D(G(z)))] 
\end{align}

For $\mathbb{T}_{\ell}$ the set of sub-sequences to train on, and $\mathbb{Z}$ the corresponding set of sub-sequences from the latent space (noise sample), we have the following stochastic gradient descend:

\begin{align}
Discriminant:& \nabla_{\theta_d} \frac{1}{|\mathbb{T}|} \sum_{(T,Z) \in (\mathbb{T},\mathbb{Z})} [-log D(T) - log (1 - D(G(Z)))] \\
Generator:& \nabla_{\theta_g} \frac{1}{|\mathbb{Z}|} \sum_{Z \in \mathbb{Z}} [1 - log (1 - D(G(Z)))]
\end{align}

This architecture has been tried for the specific case of time-series anomaly detection~\cite{DBLP:journals/corr/abs-1809-04758}. 
For the purpose of anomaly detection, the generator is trained to produce sub-sequences labeled as normal, and the discriminator is trained to discriminate the anomalies. Thus training such a model requires having a training dataset with normal sub-sequences.
One can use the discriminator and the generator simultaneously to detect the anomaly. First, given that the discriminator has been trained to separate real (i.e., normal) from fake (i.e., anomaly) sub-sequences, it can be used as a direct tool for anomaly detection. Nevertheless, the generator can also be helpful. Given that the generator has been trained to produce a realistic sub-sequence, it will most probably fail to produce a realistic anomaly. Therefore, the Euclidian distance between the sub-sequence to evaluate and what would have generated the generator with the same latent input can have some significance in discriminating anomaly.

Several anomaly detection methods based on GAN have been proposed in the literature, such as MAD-GAN~\cite{LiEtAl2019MADGAN}, USAD~\cite{10.1145/3394486.3403392} and TAnoGAN~\cite{BasharNayak2020TAnoGAN}. These approaches train GAN on the normal sections of the time series. The anomaly score is based on the combination of discriminator and reconstruction loss. VAE-GAN~\cite{NiuEtAl2020LSTMBased} is another GAN-based model that combines GAN and Variational Autoencoder. More specifically, the generator is a VAE, which further competes with the discriminator. The anomaly score is computed the same as the previous two.

\section{Evolution of Methods over Time: A Meta-Analysis}

\begin{figure}[t]
	\centering
	\includegraphics[scale=0.6]{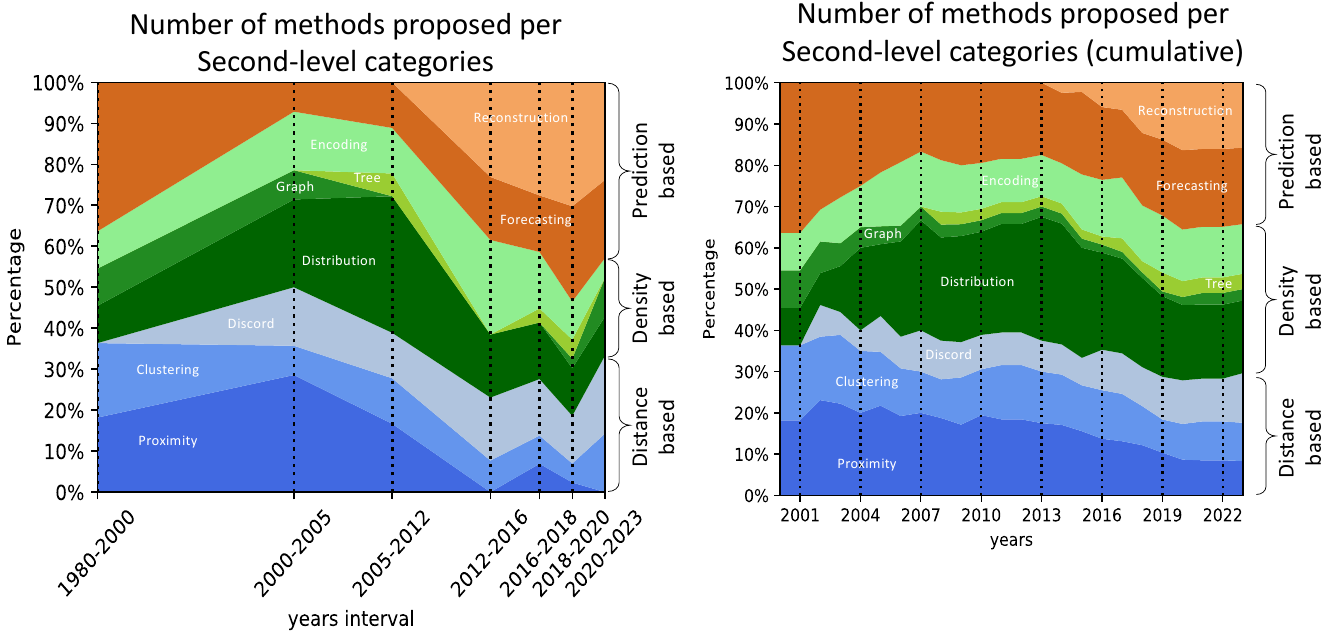}
	\caption{Relative number of methods proposed over time per category, at different times-intervals (left), and cumulative (right).} 
	\label{fig:evol_time}
\end{figure}

At this point, we described the main methods proposed in the literature to detect anomalies in time series. We grouped them into three first-level categories and 9 second-level categories. However, these first or second-level categories do not share the same distribution in time. Figure~\ref{fig:evol_time} shows the number of methods proposed per interval of years (left) and the cumulative number over the years (right).

We first observe that the number of methods proposed yearly was constant between 1990 and 2016. The number of methods proposed in the literature significantly increased after 2016. This first confirms the growing academic interest in the topic of time-series anomaly detection.

We can then dive into the second-level categories, and we observe that the significant increase in methods proposed is caused mainly by the prediction-based approach and, more specifically, by LSTM and autoencoder-based approaches. Between 2020 and 2023, such methods represent almost 50\% of the newly introduced anomaly detection methods. The great success of deep learning for computer vision causes such growth. Moreover, thanks to the open-source deep learning library such as TensorFlow and PyTorch, generic deep learning methods are easy to adapt to time series.

\begin{figure}[t]
	\centering
	\includegraphics[scale=0.62]{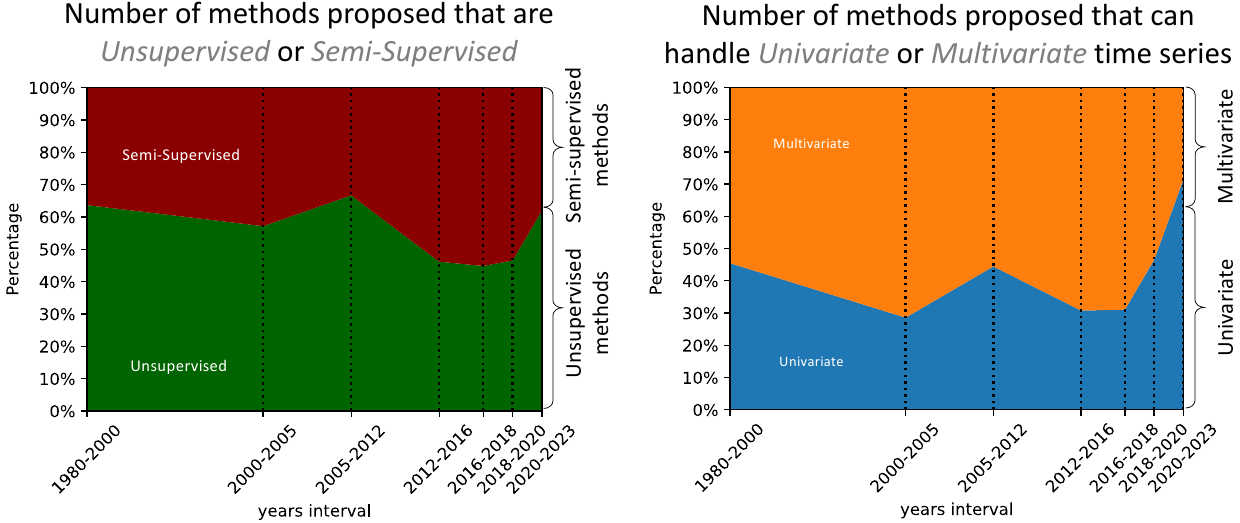}
	\caption{Number of methods proposed over time that are Unsupervised/Semi-supervised (left), and that can handle univariate/multivariate time series (right).}
	\label{fig:evol_time_type}
\end{figure}

We can then inspect the evolution of the number of methods proposed in the literature that can handle univariate or multivariate time series.
Figure~\ref{fig:evol_time_type}(right) shows the number of methods for multivariate and univariate time series per interval of years listed on the x-axis. 

Surprisingly, we observe that most of the methods proposed between 1990 and 2016 were proposed for multivariate time series, whereas, in the last three years, most of the proposed methods are for univariate time series. However, after a deep inspection, most of the methods proposed before 2016 were designed for point anomaly detection (i.e., well-defined problems for multivariate time series). The recent interest in sub-sequence anomaly detection, joined by the fact that the subsequence anomaly detection problem for multivariate time series is harder to define, leads to a significant increase in methods for univariate time series.

Finally, we can measure the evolution of the number of unsupervised and semi-supervised methods over the years. The latter is illustrated in Figure~\ref{fig:evol_time_type}(left). We observe that 65\% of the anomaly detection methods proposed in the literature were unsupervised between 1980 and 2000, whereas 50\% of the methods proposed between 2012 and 2018 were unsupervised.

\section{Evaluating Anomaly Detection}
With the continuous advancement of time-series anomaly detection methods, it becomes evident that different methods possess distinct properties and may be more suitable for specific domains. Moreover, the metrics used to evaluate these methods vary significantly in terms of their characteristics. Consequently, evaluating and selecting the most appropriate method for a given scenario has emerged as a major challenge in this field.
In this section, we will begin by presenting the benchmarks that have been proposed in the literature for evaluating time-series anomaly detection methods. Then, we will discuss different evaluation measures commonly used in the field and examine their limitations when applied to anomaly detection.

\subsection{Existing Benchmarks}

\begin{table}[tb]
\footnotesize
\caption{Summary of existing benchmarks for time-series anomaly detection.}
\begin{tabular}[t]{lcccccc}
\toprule
Benchmark & \multicolumn{1}{c}{\begin{tabular}[c]{@{}c@{}}\# Time\\ Series\end{tabular}} & \multicolumn{1}{c}{\begin{tabular}[c]{@{}c@{}}Average\\ Length\end{tabular}} & \multicolumn{1}{c}{\begin{tabular}[c]{@{}c@{}}Average \#\\ Anomalies\end{tabular}} & \multicolumn{1}{c}{\begin{tabular}[c]{@{}c@{}}Average \\ Anomaly Length\end{tabular}} & \multicolumn{1}{c}{Dim} & \multicolumn{1}{c}{\begin{tabular}[c]{@{}c@{}}Anomaly\\ Type\end{tabular}} \\ 
\toprule
NAB~\cite{lavin2015evaluating} & 58 & 6301.7 & 2 & 287.8 & I & P\&S \\
Yahoo~\cite{yahoo} & 367 & 1561.2 & 5.9 & 1.8 & I & P\&S \\
Exathlon~\cite{jacob2021exathlon} & 93 & 25115.9 & 1 & 9537.6 & M & S\\
KDD21~\cite{kdd21} & 250 & 77415.1 & 1 & 196.5 & I & P\&S \\
TODS~\cite{lai2021revisiting} & 54 & 13469.9 & 266.7 & 2.3 & I\&M & P\&S \\ 
TimeEval~\cite{10.14778/3538598.3538602} & 976 & 30991 & 5.5 & 106.7 & I\&M & P\&S \\ 
TSB-UAD~\cite{paparrizos2022tsb} & 14046 & 34043.6 & 86.3 & 24.9 & I & P\&S \\ 
TSB-AD~\cite{liu2024elephant} & 1070 (Curated) & 105485.2 & 104.2 & 409.5 & I\&M & P\&S \\ 

\toprule
\end{tabular}
    \begin{tablenotes}
      \scriptsize
      \centering
      \item  I: Univariate; M: Multivariate // P: Point; S: Subsequence
      \item The statistics are based on the datasets downloaded during the writing of this article.
     \end{tablenotes}
\label{benchmark_table}
\end{table}

In previous sections, we noted that a substantial number of time-series anomaly detection methods have been developed over the past several decades. 
Multiple surveys and experimental studies have evaluated the performance of various anomaly detectors across different datasets~\cite{paparrizos2022tsb, 10.14778/3538598.3538602,msad-sylligardos23}. These investigations have consistently highlighted the absence of a one-size-fits-all anomaly detector. 
The emerging consensus acknowledges that a model performing well on one dataset is not sufficient to declare an anomaly detection algorithm useful. The effectiveness of an anomaly detector should be demonstrated across a wide range of datasets rather than several cherry-picking datasets.
Consequently, there have been efforts made to establish benchmarks incorporating multiple datasets from various domains to ensure thorough and comprehensive evaluation.

In the following, we will overview recent benchmarks for time-series anomaly detection. These benchmarks are presented chronologically, as illustrated in Table \ref{benchmark_table}, with brief descriptions to demonstrate advancements in this field.
\newline \textbf{NAB}~\cite{lavin2015evaluating} provides 58 labeled real-world and artificial time series, primarily focusing on real-time anomaly detection for streaming data. It comprises diverse domains such as AWS server metrics, online advertisement clicking rates, real-time traffic data, and a collection of Twitter mentions of large publicly-traded companies.
\newline \textbf{Yahoo}~\cite{yahoo} comprises a collection of real and synthetic time series datasets, which are derived from the real production traffic to some of the Yahoo production systems.
\newline \textbf{Exathlon}~\cite{jacob2021exathlon} is proposed for explainable anomaly detection over high-dimensional time series data. It is constructed based on real data traces from repeated executions of large-scale stream processing jobs on an Apache Spark cluster.
\newline \textbf{KDD21} (or UCR Anomaly Archive)~\cite{kdd21} is a composite dataset that covers various domains, such as medicine, sports, and space science. It 
was designed to address the pitfalls of previous benchmarks~\cite{wu2020currentTkde}.
\newline \textbf{TODS}~\cite{lai2021revisiting} refines synthetic criterion and includes five anomaly scenarios categorized by behavior-driven taxonomy as point-global, pattern-contextual, pattern-shapelet, pattern-seasonal, and pattern-trend.
\newline \textbf{TimeEval}~\cite{10.14778/3538598.3538602} comprises a collection of datasets (both real and synthetic) from very different domains. This benchmark contains both univariate and multivariate time series mixing both point and sequence anomalies. In addition, this benchmark has been filtered such that there is no time series that have a normal/abnormal ratio above $0.1$, and that at least one method performs more than 0.8 AUC-ROC for each time series.
\newline \textbf{TSB-UAD}~\cite{paparrizos2022tsb} is a comprehensive and unified benchmark designed for evaluating univariate time-series anomaly detection methods. It includes public datasets that contain real-world anomalies, as well as synthetic datasets that provide eleven transformation methods to emulate different anomaly types. Additionally, the benchmark incorporates artificial datasets that are transformed from time-series classification datasets with varying levels of similarity between normal and abnormal subsequences. This comprehensive coverage of different anomaly scenarios makes TSB-UAD a uniform platform to compare different methods across different realistic scenarios.

We note that there are ongoing discussions regarding the limitations of certain datasets used in existing benchmarks. Wu~\textit{et al.}~\cite{wu2020currentTkde} identify four common flaws: (i) triviality, (ii) unrealistic anomaly density, (iii) mislabeled ground truth, and (iv) run-to-failure bias. Such issues underscore the substantial challenges in designing truly representative benchmarks. In response, Wu ~\textit{et al.} develop a manually curated dataset consisting primarily of univariate time series featuring a single, often artificially introduced anomaly. However, this dataset may not accurately reflect real-world scenarios (given that most previously published real-world datasets contain multiple anomalies) and excludes other potentially anomalous regions, resulting in a new set of labeling ambiguities.
To address these concerns, TSB-AD~\cite{liu2024elephant} introduces the first large-scale, heterogeneous, and meticulously curated dataset, combining human perception with model-driven interpretation to offer improved reliability.

\noindent \textbf{TSB-AD}~\cite{liu2024elephant} is the largest benchmark to date, comprising 1,000 rigorously curated, high-quality time series datasets. This benchmark include both univariate and multivariate cases, ensuring coverage of a wide range of real-world scenarios for anomaly detection. It establishes a reliable framework for evaluating methods and includes comprehensive benchmarking of 40 anomaly detection approaches (continuously updating\footnote{https://thedatumorg.github.io/TSB-AD}). Each method undergoes a thorough hyperparameter tuning to ensure optimal performance. The benchmark also incorporates the latest advances in foundation model-based methods, highlighting their potential for time series anomaly detection.

\subsection{Evaluation Measures}
In this section, we present an overview of evaluation metrics used to assess the performance of anomaly detectors.
There are various ways to categorize the evaluation metrics. It can be classified based on whether a threshold needs to be set or if the evaluation is conducted on independent time points or sequences. In the following, we will categorize the evaluation based on the requirement of threshold setting.

\begin{figure}[t]
 \centering
 \includegraphics[width=\linewidth]{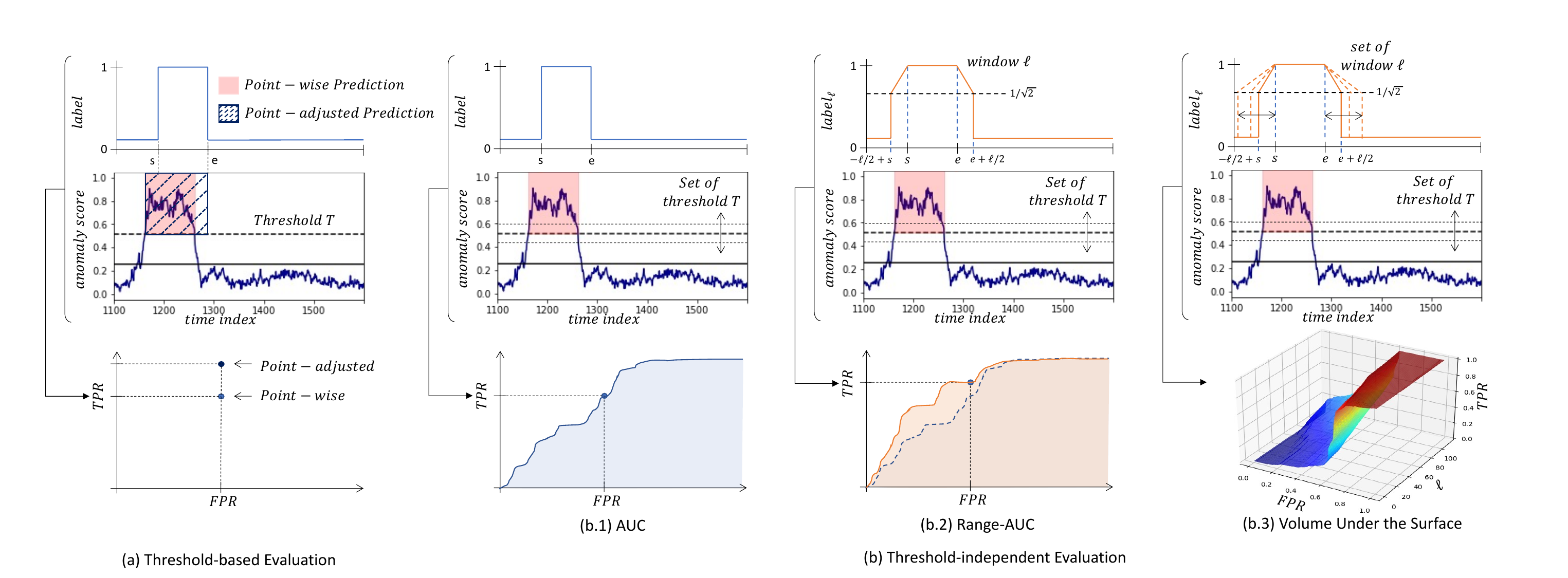}
 \caption{Illustration of evaluation measures for time-series anomaly detection.}
 \label{fig:evaluation}
\end{figure}

\subsubsection{Threshold-based Evaluation}
Threshold-based evaluation requires setting a threshold to classify each point (time step) as an anomaly or not based on the anomaly score $S_T$. Generally, a higher anomaly score value indicates a more abnormal point.
The most straightforward approach is to set the threshold to $\mu(S_T) + \alpha*\sigma(S_T)$, with $\alpha$ set to 3~\cite{statisticaloutliers}, where $\mu(S_T)$ is the mean and $\sigma(S_T)$ is the standard deviation of $S_T$. However, this approach is sensitive to extreme values in the anomaly score and can result in unfair comparisons between different methods due to variations in their statistical properties.

To address this issue, researchers in the field have developed alternative methods for threshold selection that operate automatically, eliminating the need for statistical assumptions regarding errors.
For instance, \cite{ali2013automated} introduced an adaptive thresholding technique that exploits the consistent time correlation structure observed in anomaly scores during benign activity periods. This technique dynamically adjusts the threshold based on predicted future anomaly scores.
Non-parametric dynamic thresholding, proposed in~\cite{hundman2018detecting}, aims to find a threshold such that removing all values above it results in the greatest percentage decrease in the mean and standard deviation of the anomaly scores. Another approach, known as Peaks-Over-Threshold (POT)~\cite{siffer2017anomaly,su2019robust}, involves an initial threshold selection, identification of extreme values in the tails of a probability distribution, fitting the tail portion with a generalized Pareto distribution with parameters, computing anomaly scores based on the estimated distribution, and applying a secondary threshold to identify anomalies.


After setting the threshold, we can classify the points as either normal or abnormal based on whether they exceed the threshold. In the subsequent section, we will review common evaluation measures. We begin by presenting the necessary definitions and formulations for introducing these measures, followed by a brief explanation of their distinctions. Formally, the binary predictions $pred \in \{0,1\}^n$ are obtained by comparing $S_T$ with threshold $Th$ as:

\vspace{-0.5cm}
\begin{equation}
\begin{split}
&\forall i \in [1,|S_T|], pred_i = \left.
\begin{cases}
0,& \text{if: } {S_T}_i < Th \\
1,& \text{if: } {S_T}_i \geq Th
\end{cases}
\right.
\end{split}
\vspace{-0.5cm}
\end{equation}

By comparing $pred$ to the true-labeled anomalies $label \in \{0,1\}^n$, the points can fall into one of the following four categories:
\begin{itemize}
	\item {\bf True Positive (TP)}: Number of points that have been correctly identified as anomalies.
	\item {\bf True Negative (TN)}: Number of points that have been correctly identified as normal.
	\item {\bf False Positive (FP)}: Number of points that have been wrongly identified as anomalies.
	\item {\bf False Negative (FN)}: Number of points that have been wrongly identified as normal.
\end{itemize}
Given these four categories, several point-wise evaluation measures have been proposed to assess the accuracy of anomaly detection methods.
\newline \textbf{Precision} (or positive predictive value) is the number of correctly identified anomalies over the total number of points detected as anomalies by the method: $\frac{TP}{TP+FP}$.
\newline \textbf{Recall} (or True Positive Rate, TPR) is the number of correctly identified anomalies over all anomalies: $\frac{TP}{TP+FN}$.
\newline \textbf{False Positive Rate} (or FPR) is the number of points wrongly identified as anomalies over the total number of normal points: $\frac{FP}{FP+TN}$. Contrary to Recall, the optimal score is obtained by predicting all the points as normal.
\newline \textbf{F-Score} combines precision and recall into a single metric by taking their harmonic mean, with a non-negative real value of $\beta$: $\frac{(1+\beta^2)*Precision*Recall}{\beta^2*Precision+Recall}$. Usually, $\beta$ is set to 1, balancing the importance between Precision and Recall.
\newline \textbf{Precision@k} is the precision of a subset of anomalies corresponding to $k$ highest value in the anomaly score $S_T$.

While most current methods~\cite{xu2018unsupervised,su2019robust,shen2020timeseries} calculate these metrics by treating time points as independent samples, they often employ point adjustment techniques to account for consecutive anomalies. This means that detecting any point within an anomalous segment is considered as if all points within that segment were detected, as is shown in Figure \ref{fig:evaluation}(a).
However, the work of \cite{kim2022towards} criticizes the use of point-adjusted metrics, demonstrating that they have a high likelihood of overestimating detection performance and that even a random anomaly score can yield seemingly good results.
In light of this critique, \cite{kim2022towards} propose \textbf{Point-adjusted metrics at $K$\%}, wherein a predetermined proportion $K$\% of anomalies must be detected before the application of point adjustment.
Other refined point-adjusted metrics include \textbf{Delay thresholded point-adjusted F-score} in~\cite{ren2019time,chen2021joint}. This metric considers an anomaly to be detected only if it is predicted within the first $k$ time steps of the truth-labeled anomaly.

Furthermore, the above-mentioned metrics ignore the sequential nature of time series. A range-based quality measure~\cite{tatbul2018precision} was recently proposed to address the shortcomings of point-based measures. This definition considers several factors: (i) whether a subsequence is detected or not (ExistenceReward); (ii) how many points in the subsequence are detected (OverlapReward); (iii) which part of the subsequence is detected (position-dependent weight function); and (iv) how many fragmented regions correspond to one real subsequence outlier (CardinalityFactor). In this way, point-based Precision and Recall can be extended to calculating \textbf{Range-based F-score}.

Other metrics include \textbf{NAB score}~\cite{lavin2015evaluating}, which penalizes false positive points by assigning a negative value and provides positive value rewards for accurately detecting anomalous segments, with the reward being higher for early prediction of the first anomalous point. It is noteworthy that the utilization of the NAB score itself is not widespread; however, the benchmark introduced in this paper is widely adopted in the research community for evaluation using other metrics.

\subsubsection{Threshold-independent Evaluation}
The requirement to apply a threshold to the anomaly score significantly affects the accuracy measures. They can vary significantly when the threshold changes. According to a recent work~\cite{paparrizos2022volume}, threshold-based measures are particularly sensitive to the noisy anomaly score, which stem from noise in the original time series. As the score fluctuates around the threshold, they become less robust to noise. Moreover, the normal/abnormal ratio, which exhibits considerable variability across different tasks, further influences the threshold. Notably, variations in this ratio can lead to variations in the threshold, consequently impacting the values of threshold-based accuracy measures. Additionally, detectors may introduce a lag into the anomaly score, and there may be inherent lag resulting from the approximation made during the labeling phase. Even small lags can have a significant impact on these evaluation measures.
Therefore, many works consider threshold selection as a problem orthogonal to model evaluation and use metrics that summarize the model performance across all possible thresholds. 
We will introduce several threshold-independent evaluation measures as follows.
\newline \textbf{Best F-Score}: Maximum F-Score over all possible thresholds.
\newline \textbf{AUC}: 
The Area Under the Receiver Operating Characteristics curve (AUC-ROC)~\cite{FAWCETT2006861} is a widely used evaluation metric in anomaly detection, as well as in binary classification in general.
It quantifies the performance of a model by measuring the area under the curve that represents the true positive rate (TPR) on the y-axis against the false positive rate (FPR) on the x-axis, as depicted in Figure \ref{fig:evaluation}(b.1). AUC-ROC represents the probability that a randomly chosen positive example will be ranked higher than a randomly chosen negative example.
It is computed using the trapezoidal rule. For that purpose, we define an ordered set of thresholds, denoted as $Th$, ranging from 0 to 1, where $Th=[Th_0,Th_1,...Th_N]$ with $0=Th_0<Th_1<...<Th_N=1$. Therefore, $AUC\text{-}ROC$ is defined as follows:
\vspace{-0.2cm}
\begin{equation}
\begin{split}
&AUC\text{-}ROC = \frac{1}{2}\sum_{k=1}^{N} \Delta^{k}_{TPR}*\Delta^{k}_{FPR}\\
&\text{with: } \left.
\begin{cases}
\Delta^{k}_{FPR} &= FPR(Th_{k})-FPR(Th_{k-1})\\
\Delta^{k}_{TPR} &= TPR(Th_{k-1})+TPR(Th_{k})
\end{cases}
\right. 
\end{split}
\vspace{-0.3cm}
\label{equAUCROC}
\end{equation}
The Area Under the Precision-Recall curve (AUC-PR)~\cite{10.1145/1143844.1143874} is similar, but with the Recall (TPR) on the x-axis and Precision on the y-axis.
The Precision and FPR exhibit distinct responses to changes in false positives in the context of anomaly detection. In this domain, the number of true negatives tends to be substantially larger than the number of false positives, resulting in low FPR values for various threshold choices that are relevant. Consequently, only a small portion of the ROC curve holds relevance under such circumstances. One potential approach to address this issue is to focus solely on specific segments of the curve~\cite{baker2001proposed}. Alternatively, the use of the AUC-PR has been advocated as a more informative alternative to ROC for imbalanced datasets~\cite{lobo2008auc}.
\newline \textbf{Range-AUC}:
AUC-ROC and AUC-PR are primarily designed for point-based anomalies, treating each point independently and assigning equal weight to the detection of each point in calculating the overall AUC. However, these metrics may not be ideal for assessing subsequence anomalies. There are several reasons for this limitation, including the importance of detecting even small segments within subsequence outliers, the absence of consistent labeling conventions across datasets, especially for subsequences, and the sensitivity of the anomaly scores to time lags introduced by detectors.
To address these limitations, an extension of the ROC and PR curves called Range-AUC~\cite{paparrizos2022volume} has been introduced specifically for subsequences. By adding a buffer region at the outliers' boundaries as shown in Figure \ref{fig:evaluation}(b.2), it accounts for the false tolerance of labeling in the ground truth and assigns higher anomaly scores near the outlier boundaries. It replaces binary labels with continuous values in the range [0, 1]. This refinement enables the adaptation of point-based TPR, FPR, and Precision to better suit subsequence anomaly cases.
\newline \textbf{Volume Under the Surface (VUS)}~\cite{paparrizos2022volume}: The buffer length in Range-AUC, denoted as $l$, needs to be predefined. If not properly set, it can strongly influence range-AUC measures. To eliminate this influence, VUS computes Range-AUC for different buffer lengths from 0 to the $l$, which leads to the creation of a three-dimensional surface in the ROC-PR space as shown in Figure \ref{fig:evaluation}(b.3). The VUS family of measures, including VUS-ROC and VUS-PR, are parameter-free and threshold-independent, applicable for evaluating both point-based and range-based anomalies.

Different evaluation methods have different properties, including robustness to lag and noise, the separability to differentiate between accurate and inaccurate methods, and the need for parameters, etc. Therefore, the selection of evaluation metrics should be approached with caution, considering the specific requirements of the task. For detailed case studies highlighting the properties of different metrics, we recommend referring to the following papers~\cite{sorbo2023navigating,paparrizos2022volume}.
In terms of key takeaways, we recommend utilizing threshold-independent evaluation measures to mitigate potential biases introduced by threshold selection. AUC-based measures have been widely adopted in this regard. However, their limitations lie in the lack of consideration for the consistency of time series. To address this, Range-AUC has refined the AUC-based measures. Among the range-based measures, VUS-ROC stands out for its robustness, separability, and consistency in ranking the accuracy of detectors across diverse datasets, making it a recommended choice as the evaluation measure of preference.

\section{Conclusions}
\label{sec:conclusions}

In this survey, we examined into the anomaly detection problem in time series. We started by defining a taxonomy of time series types, anomaly types, and anomaly detection methods. 
We grouped the methods into multiple process-centric categories. 
We then described the most popular anomaly detection methods for each category in detail, and provided an extensive list of other existing methods. 
We finally discussed the problem of benchmarking and evaluation of anomaly detection methods in time series. 
We initiated this discussion by listing the time series dataset and benchmark proposed in the literature. 
We then listed the traditional evaluation measures used to assess the detection quality, discussed their limitations, and introduced a recent effort to adapt these measures to the context of time series.

Despite the decades-long worth of research in this area, time-series anomaly detection remains a challenging problem. 
Several communities have tackled the problem separately, introducing methods that follow their corresponding fundamental concepts. 
Since these methods were not compared on the same basis (i.e., using the same evaluation measures and datasets), the progress of anomaly detection methods has been challenging to assess. 
However, the recent efforts in proposing benchmarks~\cite{kdd21,paparrizos2022tsb} has helped to evaluate the progress and identify appropriate methods for specific problems~\cite{10.14778/3538598.3538602,paparrizos2022tsb}.

Nevertheless, multiple research directions remain open. 
First, there is no agreement yet on a single benchmark that the entire community should use. 
Even though numerous benchmarks have been proposed, they have their own limitations concerning the diversity of time series, anomaly types, or uncertain labels. 
There is a need to agree as a community on a common basis for comparing the anomaly detection methods. 

Second, encouraged by the current momentum, many novel methods are proposed every year. 
However, recent evaluations suggested~\cite{10.14778/3538598.3538602,paparrizos2022tsb} that no single best method exists (i.e., achieving the best performance on every dataset). 
This observation opens a new direction of research towards ensembling, model selection, and AutoML. 
A recent experimental evaluation~\cite{msad-sylligardos23} concluded that simple time series classification baselines can be used for model selection for time-series anomaly detection, leading to accuracy improvements by a factor of 2 compared to the best performing individual anomaly detection method. 
This study suggests that we can be optimistic about further improvements in accuracy, by continuing the research in this direction.

Finally, even though a large number of unsupervised methods have been proposed for univariate time-series anomaly detection, not much attention has been paid to multivariate time series, streaming time series, series with missing values, series with non-continuous timestamps, heterogeneous time series, or a combination of the above. 
Such times series are often encountered in practice, thus we need robust and accurate methods for these cases, as well.


\bibliographystyle{ACM-Reference-Format}
\bibliography{anomalies}


\begin{thebibliography}{272}


\ifx \showCODEN    \undefined \def \showCODEN     #1{\unskip}     \fi
\ifx \showDOI      \undefined \def \showDOI       #1{#1}\fi
\ifx \showISBNx    \undefined \def \showISBNx     #1{\unskip}     \fi
\ifx \showISBNxiii \undefined \def \showISBNxiii  #1{\unskip}     \fi
\ifx \showISSN     \undefined \def \showISSN      #1{\unskip}     \fi
\ifx \showLCCN     \undefined \def \showLCCN      #1{\unskip}     \fi
\ifx \shownote     \undefined \def \shownote      #1{#1}          \fi
\ifx \showarticletitle \undefined \def \showarticletitle #1{#1}   \fi
\ifx \showURL      \undefined \def \showURL       {\relax}        \fi
\providecommand\bibfield[2]{#2}
\providecommand\bibinfo[2]{#2}
\providecommand\natexlab[1]{#1}
\providecommand\showeprint[2][]{arXiv:#2}

\bibitem[Abdul-Aziz et~al\mbox{.}(2010)]%
        {Abdul-Aziz2010Propulsion}
\bibfield{author}{\bibinfo{person}{Ali Abdul-Aziz}, \bibinfo{person}{Mark Woike}, \bibinfo{person}{Nikunj Oza}, \bibinfo{person}{Bryan Matthews}, {and} \bibinfo{person}{George Baakilini}.} \bibinfo{year}{2010}\natexlab{}.
\newblock \showarticletitle{{Propulsion health monitoring of a turbine engine disk using spin test data}}. In \bibinfo{booktitle}{\emph{Health Monitoring of Structural and Biological Systems 2010}}, \bibfield{editor}{\bibinfo{person}{Tribikram Kundu}} (Ed.), Vol.~\bibinfo{volume}{7650}. International Society for Optics and Photonics, \bibinfo{publisher}{SPIE}, \bibinfo{pages}{431 -- 440}.
\newblock
\urldef\tempurl%
\url{https://doi.org/10.1117/12.847574}
\showDOI{\tempurl}


\bibitem[Aggarwal(2017)]%
        {aggarwal2017introduction}
\bibfield{author}{\bibinfo{person}{Charu~C Aggarwal}.} \bibinfo{year}{2017}\natexlab{}.
\newblock \showarticletitle{An introduction to outlier analysis}.
\newblock In \bibinfo{booktitle}{\emph{Outlier analysis}}. \bibinfo{publisher}{Springer}, \bibinfo{pages}{1--34}.
\newblock


\bibitem[Ahmad et~al\mbox{.}(2017)]%
        {AhmadEtAl2017Unsupervised}
\bibfield{author}{\bibinfo{person}{Subutai Ahmad}, \bibinfo{person}{Alexander Lavin}, \bibinfo{person}{Scott Purdy}, {and} \bibinfo{person}{Zuha Agha}.} \bibinfo{year}{2017}\natexlab{}.
\newblock \showarticletitle{Unsupervised Real-Time Anomaly Detection for Streaming Data}.
\newblock   \bibinfo{volume}{262} (\bibinfo{year}{2017}), \bibinfo{pages}{134--147}.
\newblock
\showISSN{09252312}
\urldef\tempurl%
\url{https://doi.org/10.1016/j.neucom.2017.04.070}
\showDOI{\tempurl}


\bibitem[Alam et~al\mbox{.}(2015)]%
        {alam2015eleventh}
\bibfield{author}{\bibinfo{person}{Shadab Alam}, \bibinfo{person}{Franco~D Albareti}, \bibinfo{person}{Carlos~Allende Prieto}, \bibinfo{person}{Friedrich Anders}, \bibinfo{person}{Scott~F Anderson}, \bibinfo{person}{Timothy Anderton}, \bibinfo{person}{Brett~H Andrews}, \bibinfo{person}{Eric Armengaud}, \bibinfo{person}{{\'E}ric Aubourg}, \bibinfo{person}{Stephen Bailey}, {et~al\mbox{.}}} \bibinfo{year}{2015}\natexlab{}.
\newblock \showarticletitle{The eleventh and twelfth data releases of the Sloan Digital Sky Survey: final data from SDSS-III}.
\newblock \bibinfo{journal}{\emph{The Astrophysical Journal Supplement Series}} \bibinfo{volume}{219}, \bibinfo{number}{1} (\bibinfo{year}{2015}), \bibinfo{pages}{12}.
\newblock


\bibitem[Ali et~al\mbox{.}(2013)]%
        {ali2013automated}
\bibfield{author}{\bibinfo{person}{Muhammad~Qasim Ali}, \bibinfo{person}{Ehab Al-Shaer}, \bibinfo{person}{Hassan Khan}, {and} \bibinfo{person}{Syed~Ali Khayam}.} \bibinfo{year}{2013}\natexlab{}.
\newblock \showarticletitle{Automated anomaly detector adaptation using adaptive threshold tuning}.
\newblock \bibinfo{journal}{\emph{ACM Transactions on Information and System Security (TISSEC)}} \bibinfo{volume}{15}, \bibinfo{number}{4} (\bibinfo{year}{2013}), \bibinfo{pages}{1--30}.
\newblock


\bibitem[Alvarez et~al\mbox{.}(2010)]%
        {alvarez2010energy}
\bibfield{author}{\bibinfo{person}{Francisco~Martinez Alvarez}, \bibinfo{person}{Alicia Troncoso}, \bibinfo{person}{Jose~C Riquelme}, {and} \bibinfo{person}{Jesus S~Aguilar Ruiz}.} \bibinfo{year}{2010}\natexlab{}.
\newblock \showarticletitle{Energy time series forecasting based on pattern sequence similarity}.
\newblock \bibinfo{journal}{\emph{IEEE Transactions on Knowledge and Data Engineering}} \bibinfo{volume}{23}, \bibinfo{number}{8} (\bibinfo{year}{2010}), \bibinfo{pages}{1230--1243}.
\newblock


\bibitem[Antoni and Borghesani(2019)]%
        {AntoniBorghesani2019statistical}
\bibfield{author}{\bibinfo{person}{Jérôme Antoni} {and} \bibinfo{person}{Pietro Borghesani}.} \bibinfo{year}{2019}\natexlab{}.
\newblock \showarticletitle{A Statistical Methodology for the Design of Condition Indicators}.
\newblock   \bibinfo{volume}{114} (\bibinfo{year}{2019}), \bibinfo{pages}{290--327}.
\newblock
\showISSN{08883270}
\urldef\tempurl%
\url{https://doi.org/10.1016/j.ymssp.2018.05.012}
\showDOI{\tempurl}


\bibitem[Audibert et~al\mbox{.}(2020)]%
        {10.1145/3394486.3403392}
\bibfield{author}{\bibinfo{person}{Julien Audibert}, \bibinfo{person}{Pietro Michiardi}, \bibinfo{person}{Fr{\'e}d{\'e}ric Guyard}, \bibinfo{person}{S{\'e}bastien Marti}, {and} \bibinfo{person}{Maria~A Zuluaga}.} \bibinfo{year}{2020}\natexlab{}.
\newblock \showarticletitle{Usad: Unsupervised anomaly detection on multivariate time series}. In \bibinfo{booktitle}{\emph{SIGKDD}}. \bibinfo{pages}{3395--3404}.
\newblock


\bibitem[Bach-Andersen et~al\mbox{.}(2017)]%
        {bach2017flexible}
\bibfield{author}{\bibinfo{person}{Martin Bach-Andersen}, \bibinfo{person}{Bo R{\o}mer-Odgaard}, {and} \bibinfo{person}{Ole Winther}.} \bibinfo{year}{2017}\natexlab{}.
\newblock \showarticletitle{Flexible non-linear predictive models for large-scale wind turbine diagnostics}.
\newblock \bibinfo{journal}{\emph{Wind Energy}} \bibinfo{volume}{20}, \bibinfo{number}{5} (\bibinfo{year}{2017}), \bibinfo{pages}{753--764}.
\newblock


\bibitem[Baker and Pinsky(2001)]%
        {baker2001proposed}
\bibfield{author}{\bibinfo{person}{Stuart~G Baker} {and} \bibinfo{person}{Paul~F Pinsky}.} \bibinfo{year}{2001}\natexlab{}.
\newblock \showarticletitle{A proposed design and analysis for comparing digital and analog mammography: special receiver operating characteristic methods for cancer screening}.
\newblock \bibinfo{journal}{\emph{J. Amer. Statist. Assoc.}} \bibinfo{volume}{96}, \bibinfo{number}{454} (\bibinfo{year}{2001}), \bibinfo{pages}{421--428}.
\newblock


\bibitem[Bar-Joseph(2004)]%
        {bar2004analyzing}
\bibfield{author}{\bibinfo{person}{Ziv Bar-Joseph}.} \bibinfo{year}{2004}\natexlab{}.
\newblock \showarticletitle{Analyzing time series gene expression data}.
\newblock \bibinfo{journal}{\emph{Bioinformatics}} \bibinfo{volume}{20}, \bibinfo{number}{16} (\bibinfo{year}{2004}), \bibinfo{pages}{2493--2503}.
\newblock


\bibitem[Bar-Joseph et~al\mbox{.}(2003)]%
        {bar2003continuous}
\bibfield{author}{\bibinfo{person}{Ziv Bar-Joseph}, \bibinfo{person}{Georg~K Gerber}, \bibinfo{person}{David~K Gifford}, \bibinfo{person}{Tommi~S Jaakkola}, {and} \bibinfo{person}{Itamar Simon}.} \bibinfo{year}{2003}\natexlab{}.
\newblock \showarticletitle{Continuous representations of time-series gene expression data}.
\newblock \bibinfo{journal}{\emph{Journal of Computational Biology}} \bibinfo{volume}{10}, \bibinfo{number}{3-4} (\bibinfo{year}{2003}), \bibinfo{pages}{341--356}.
\newblock


\bibitem[Bar-Joseph et~al\mbox{.}(2012)]%
        {bar2012studying}
\bibfield{author}{\bibinfo{person}{Ziv Bar-Joseph}, \bibinfo{person}{Anthony Gitter}, {and} \bibinfo{person}{Itamar Simon}.} \bibinfo{year}{2012}\natexlab{}.
\newblock \showarticletitle{Studying and modelling dynamic biological processes using time-series gene expression data}.
\newblock \bibinfo{journal}{\emph{Nature Reviews Genetics}} \bibinfo{volume}{13}, \bibinfo{number}{8} (\bibinfo{year}{2012}), \bibinfo{pages}{552}.
\newblock


\bibitem[Bariya et~al\mbox{.}(2021)]%
        {bariya2021k}
\bibfield{author}{\bibinfo{person}{Mohini Bariya}, \bibinfo{person}{Alexandra von Meier}, \bibinfo{person}{John Paparrizos}, {and} \bibinfo{person}{Michael~J Franklin}.} \bibinfo{year}{2021}\natexlab{}.
\newblock \showarticletitle{k-ShapeStream: Probabilistic Streaming Clustering for Electric Grid Events}. In \bibinfo{booktitle}{\emph{2021 IEEE Madrid PowerTech}}. IEEE, \bibinfo{pages}{1--6}.
\newblock


\bibitem[Barnet and Lewis(1994)]%
        {statisticaloutliers}
\bibfield{author}{\bibinfo{person}{V. Barnet} {and} \bibinfo{person}{T. Lewis}.} \bibinfo{year}{1994}\natexlab{}.
\newblock \bibinfo{booktitle}{\emph{{Outliers in Statistical Data}}}.
\newblock \bibinfo{publisher}{{John Wiley and Sons, Inc.}}
\newblock


\bibitem[Barnett and Lewis(1984)]%
        {barnett1984outliers}
\bibfield{author}{\bibinfo{person}{Vic Barnett} {and} \bibinfo{person}{Toby Lewis}.} \bibinfo{year}{1984}\natexlab{}.
\newblock \showarticletitle{Outliers in statistical data}.
\newblock \bibinfo{journal}{\emph{Wiley Series in Probability and Mathematical Statistics. Applied Probability and Statistics}} (\bibinfo{year}{1984}).
\newblock


\bibitem[Bashar and Nayak(2020)]%
        {BasharNayak2020TAnoGAN}
\bibfield{author}{\bibinfo{person}{Md~Abul Bashar} {and} \bibinfo{person}{Richi Nayak}.} \bibinfo{year}{2020}\natexlab{}.
\newblock \showarticletitle{TAnoGAN: Time series anomaly detection with generative adversarial networks}. In \bibinfo{booktitle}{\emph{SSCI}}. IEEE, \bibinfo{pages}{1778--1785}.
\newblock


\bibitem[Basu and Meckesheimer(2007)]%
        {BasuMeckesheimer2007Automatic}
\bibfield{author}{\bibinfo{person}{Sabyasachi Basu} {and} \bibinfo{person}{Martin Meckesheimer}.} \bibinfo{year}{2007}\natexlab{}.
\newblock \showarticletitle{Automatic Outlier Detection for Time Series: An Application to Sensor Data}.
\newblock  \bibinfo{volume}{11}, \bibinfo{number}{2} (\bibinfo{year}{2007}), \bibinfo{pages}{137--154}.
\newblock
\showISSN{0219-1377, 0219-3116}
\urldef\tempurl%
\url{https://doi.org/10.1007/s10115-006-0026-6}
\showDOI{\tempurl}


\bibitem[Bernoulli and Allen(1961)]%
        {bernoulli1961most}
\bibfield{author}{\bibinfo{person}{Daniel Bernoulli} {and} \bibinfo{person}{CG Allen}.} \bibinfo{year}{1961}\natexlab{}.
\newblock \showarticletitle{The most probable choice between several discrepant observations and the formation therefrom of the most likely induction}.
\newblock \bibinfo{journal}{\emph{Biometrika}} \bibinfo{volume}{48}, \bibinfo{number}{1-2} (\bibinfo{year}{1961}), \bibinfo{pages}{3--18}.
\newblock


\bibitem[Bhargava and Raghuvanshi(2013)]%
        {BhargavaRaghuvanshi2013Anomaly}
\bibfield{author}{\bibinfo{person}{Arpita Bhargava} {and} \bibinfo{person}{AS Raghuvanshi}.} \bibinfo{year}{2013}\natexlab{}.
\newblock \showarticletitle{Anomaly detection in wireless sensor networks using S-Transform in combination with SVM}. In \bibinfo{booktitle}{\emph{2013 5th International Conference and Computational Intelligence and Communication Networks}}. IEEE, \bibinfo{pages}{111--116}.
\newblock


\bibitem[Biswal et~al\mbox{.}(2010)]%
        {biswal2010toward}
\bibfield{author}{\bibinfo{person}{Bharat~B Biswal}, \bibinfo{person}{Maarten Mennes}, \bibinfo{person}{Xi-Nian Zuo}, \bibinfo{person}{Suril Gohel}, \bibinfo{person}{Clare Kelly}, \bibinfo{person}{Steve~M Smith}, \bibinfo{person}{Christian~F Beckmann}, \bibinfo{person}{Jonathan~S Adelstein}, \bibinfo{person}{Randy~L Buckner}, \bibinfo{person}{Stan Colcombe}, {et~al\mbox{.}}} \bibinfo{year}{2010}\natexlab{}.
\newblock \showarticletitle{Toward discovery science of human brain function}.
\newblock \bibinfo{journal}{\emph{Proceedings of the National Academy of Sciences}} \bibinfo{volume}{107}, \bibinfo{number}{10} (\bibinfo{year}{2010}), \bibinfo{pages}{4734--4739}.
\newblock


\bibitem[Bl{\'a}zquez-Garc{\'\i}a et~al\mbox{.}(2021)]%
        {Blazquez-GarciaEtAl2020review}
\bibfield{author}{\bibinfo{person}{Ane Bl{\'a}zquez-Garc{\'\i}a}, \bibinfo{person}{Angel Conde}, \bibinfo{person}{Usue Mori}, {and} \bibinfo{person}{Jose~A Lozano}.} \bibinfo{year}{2021}\natexlab{}.
\newblock \showarticletitle{A review on outlier/anomaly detection in time series data}.
\newblock \bibinfo{journal}{\emph{ACM Computing Surveys (CSUR)}} \bibinfo{volume}{54}, \bibinfo{number}{3} (\bibinfo{year}{2021}), \bibinfo{pages}{1--33}.
\newblock


\bibitem[Boniol et~al\mbox{.}(2020a)]%
        {BoniolEtAl2020Automated}
\bibfield{author}{\bibinfo{person}{Paul Boniol}, \bibinfo{person}{Michele Linardi}, \bibinfo{person}{Federico Roncallo}, {and} \bibinfo{person}{Themis Palpanas}.} \bibinfo{year}{2020}\natexlab{a}.
\newblock \showarticletitle{Automated {{Anomaly Detection}} in {{Large Sequences}}}. In \bibinfo{booktitle}{\emph{Proceedings of the {{International Conference}} on {{Data Engineering}} ({{ICDE}})}}. \bibinfo{pages}{1834--1837}.
\newblock
\showISBNx{978-1-72812-903-7}
\urldef\tempurl%
\url{https://doi.org/10.1109/ICDE48307.2020.00182}
\showDOI{\tempurl}


\bibitem[Boniol et~al\mbox{.}(2020b)]%
        {BoniolEtAl2020SAD}
\bibfield{author}{\bibinfo{person}{Paul Boniol}, \bibinfo{person}{Michele Linardi}, \bibinfo{person}{Federico Roncallo}, {and} \bibinfo{person}{Themis Palpanas}.} \bibinfo{year}{2020}\natexlab{b}.
\newblock \showarticletitle{{{SAD}}: {{An Unsupervised System}} for {{Subsequence Anomaly Detection}}}. In \bibinfo{booktitle}{\emph{Proceedings of the {{International Conference}} on {{Data Engineering}} ({{ICDE}})}}. \bibinfo{pages}{1778--1781}.
\newblock
\showISSN{2375-026X}
\urldef\tempurl%
\url{https://doi.org/10.1109/ICDE48307.2020.00168}
\showDOI{\tempurl}


\bibitem[Boniol et~al\mbox{.}(2021a)]%
        {boniol_unsupervised_2021}
\bibfield{author}{\bibinfo{person}{Paul Boniol}, \bibinfo{person}{Michele Linardi}, \bibinfo{person}{Federico Roncallo}, \bibinfo{person}{Themis Palpanas}, \bibinfo{person}{Mohammed Meftah}, {and} \bibinfo{person}{Emmanuel Remy}.} \bibinfo{year}{2021}\natexlab{a}.
\newblock \showarticletitle{Unsupervised and scalable subsequence anomaly detection in large data series}.
\newblock \bibinfo{journal}{\emph{The VLDB Journal}} (\bibinfo{date}{March} \bibinfo{year}{2021}).
\newblock
\showISSN{0949-877X}
\urldef\tempurl%
\url{https://doi.org/10.1007/s00778-021-00655-8}
\showDOI{\tempurl}


\bibitem[Boniol and Palpanas(2020)]%
        {BoniolPalpanas2020Series2Graph}
\bibfield{author}{\bibinfo{person}{Paul Boniol} {and} \bibinfo{person}{Themis Palpanas}.} \bibinfo{year}{2020}\natexlab{}.
\newblock \showarticletitle{{{Series2Graph}}: {{Graph}}-Based {{Subsequence Anomaly Detection}} for {{Time Series}}}.
\newblock  \bibinfo{volume}{13}, \bibinfo{number}{11} (\bibinfo{year}{2020}), \bibinfo{pages}{14}.
\newblock
\urldef\tempurl%
\url{https://doi.org/10.14778/3407790.3407792}
\showDOI{\tempurl}


\bibitem[Boniol et~al\mbox{.}(2022)]%
        {boniol2022theseus}
\bibfield{author}{\bibinfo{person}{Paul Boniol}, \bibinfo{person}{John Paparrizos}, \bibinfo{person}{Yuhao Kang}, \bibinfo{person}{Themis Palpanas}, \bibinfo{person}{Ruey~S Tsay}, \bibinfo{person}{Aaron~J Elmore}, {and} \bibinfo{person}{Michael~J Franklin}.} \bibinfo{year}{2022}\natexlab{}.
\newblock \showarticletitle{Theseus: navigating the labyrinth of time-series anomaly detection}.
\newblock \bibinfo{journal}{\emph{Proceedings of the VLDB Endowment}} \bibinfo{volume}{15}, \bibinfo{number}{12} (\bibinfo{year}{2022}), \bibinfo{pages}{3702--3705}.
\newblock


\bibitem[Boniol et~al\mbox{.}(2023)]%
        {boniol2023new}
\bibfield{author}{\bibinfo{person}{Paul Boniol}, \bibinfo{person}{John Paparrizos}, {and} \bibinfo{person}{Themis Palpanas}.} \bibinfo{year}{2023}\natexlab{}.
\newblock \showarticletitle{New Trends in Time Series Anomaly Detection.}. In \bibinfo{booktitle}{\emph{EDBT}}. \bibinfo{pages}{847--850}.
\newblock


\bibitem[Boniol et~al\mbox{.}(2024a)]%
        {boniol2024interactive}
\bibfield{author}{\bibinfo{person}{Paul Boniol}, \bibinfo{person}{John Paparrizos}, {and} \bibinfo{person}{Themis Palpanas}.} \bibinfo{year}{2024}\natexlab{a}.
\newblock \showarticletitle{An Interactive Dive into Time-Series Anomaly Detection}. In \bibinfo{booktitle}{\emph{2024 IEEE 40th International Conference on Data Engineering (ICDE)}}.
\newblock


\bibitem[Boniol et~al\mbox{.}(2021b)]%
        {boniol2021sandinaction}
\bibfield{author}{\bibinfo{person}{Paul Boniol}, \bibinfo{person}{John Paparrizos}, \bibinfo{person}{Themis Palpanas}, {and} \bibinfo{person}{Michael~J Franklin}.} \bibinfo{year}{2021}\natexlab{b}.
\newblock \showarticletitle{Sand in action: subsequence anomaly detection for streams}.
\newblock \bibinfo{journal}{\emph{Proceedings of the VLDB Endowment}} \bibinfo{volume}{14}, \bibinfo{number}{12} (\bibinfo{year}{2021}), \bibinfo{pages}{2867--2870}.
\newblock


\bibitem[Boniol et~al\mbox{.}(2021c)]%
        {boniol2021sand}
\bibfield{author}{\bibinfo{person}{Paul Boniol}, \bibinfo{person}{John Paparrizos}, \bibinfo{person}{Themis Palpanas}, {and} \bibinfo{person}{Michael~J Franklin}.} \bibinfo{year}{2021}\natexlab{c}.
\newblock \showarticletitle{SAND: streaming subsequence anomaly detection}.
\newblock \bibinfo{journal}{\emph{PVLDB}} \bibinfo{volume}{14}, \bibinfo{number}{10} (\bibinfo{year}{2021}), \bibinfo{pages}{1717--1729}.
\newblock


\bibitem[Boniol et~al\mbox{.}(2024b)]%
        {boniol2024adecimo}
\bibfield{author}{\bibinfo{person}{Paul Boniol}, \bibinfo{person}{Emmanouil Sylligardos}, \bibinfo{person}{John Paparrizos}, \bibinfo{person}{Panos Trahanias}, {and} \bibinfo{person}{Themis Palpanas}.} \bibinfo{year}{2024}\natexlab{b}.
\newblock \showarticletitle{ADecimo: Model Selection for Time Series Anomaly Detection}. In \bibinfo{booktitle}{\emph{2024 IEEE 40th International Conference on Data Engineering (ICDE)}}.
\newblock


\bibitem[Braei and Wagner(2020)]%
        {BraeiWagner2020Anomaly}
\bibfield{author}{\bibinfo{person}{Mohammad Braei} {and} \bibinfo{person}{Sebastian Wagner}.} \bibinfo{year}{2020}\natexlab{}.
\newblock \showarticletitle{Anomaly detection in univariate time-series: A survey on the state-of-the-art}.
\newblock \bibinfo{journal}{\emph{arXiv preprint arXiv:2004.00433}} (\bibinfo{year}{2020}).
\newblock


\bibitem[Breunig et~al\mbox{.}(2000a)]%
        {BreunigEtAl2000LOF}
\bibfield{author}{\bibinfo{person}{Markus~M Breunig}, \bibinfo{person}{Hans-Peter Kriegel}, \bibinfo{person}{Raymond~T Ng}, {and} \bibinfo{person}{J{\"o}rg Sander}.} \bibinfo{year}{2000}\natexlab{a}.
\newblock \showarticletitle{LOF: identifying density-based local outliers}. In \bibinfo{booktitle}{\emph{Proceedings of the 2000 ACM SIGMOD international conference on Management of data}}. \bibinfo{pages}{93--104}.
\newblock


\bibitem[Breunig et~al\mbox{.}(2000b)]%
        {Breunig:2000:LID:342009.335388}
\bibfield{author}{\bibinfo{person}{Markus~M. Breunig}, \bibinfo{person}{Hans-Peter Kriegel}, \bibinfo{person}{Raymond~T. Ng}, {and} \bibinfo{person}{J\"{o}rg Sander}.} \bibinfo{year}{2000}\natexlab{b}.
\newblock \showarticletitle{LOF: Identifying Density-based Local Outliers}. In \bibinfo{booktitle}{\emph{SIGMOD}}.
\newblock


\bibitem[Brockwell and Davis(2016)]%
        {brockwell2016introduction}
\bibfield{author}{\bibinfo{person}{Peter~J Brockwell} {and} \bibinfo{person}{Richard~A Davis}.} \bibinfo{year}{2016}\natexlab{}.
\newblock \bibinfo{booktitle}{\emph{Introduction to time series and forecasting}}.
\newblock \bibinfo{publisher}{springer}.
\newblock


\bibitem[Bu et~al\mbox{.}(2007)]%
        {DBLP:conf/sdm/BuLFKPM07}
\bibfield{author}{\bibinfo{person}{Yingyi Bu}, \bibinfo{person}{Oscar~Tat{-}Wing Leung}, \bibinfo{person}{Ada~Wai{-}Chee Fu}, \bibinfo{person}{Eamonn~J. Keogh}, \bibinfo{person}{Jian Pei}, {and} \bibinfo{person}{Sam Meshkin}.} \bibinfo{year}{2007}\natexlab{}.
\newblock \showarticletitle{{WAT:} Finding Top-K Discords in Time Series Database}. In \bibinfo{booktitle}{\emph{SIAM}}.
\newblock


\bibitem[Budalakoti et~al\mbox{.}(2008)]%
        {BudalakotiEtAl2009Anomaly}
\bibfield{author}{\bibinfo{person}{Suratna Budalakoti}, \bibinfo{person}{Ashok~N Srivastava}, {and} \bibinfo{person}{Matthew~Eric Otey}.} \bibinfo{year}{2008}\natexlab{}.
\newblock \showarticletitle{Anomaly detection and diagnosis algorithms for discrete symbol sequences with applications to airline safety}.
\newblock \bibinfo{journal}{\emph{IEEE Transactions on Systems, Man, and Cybernetics, Part C (Applications and Reviews)}} \bibinfo{volume}{39}, \bibinfo{number}{1} (\bibinfo{year}{2008}), \bibinfo{pages}{101--113}.
\newblock


\bibitem[Bui et~al\mbox{.}(2018)]%
        {BuiEtAl2018Time}
\bibfield{author}{\bibinfo{person}{C. Bui}, \bibinfo{person}{N. Pham}, \bibinfo{person}{A. Vo}, \bibinfo{person}{A. Tran}, \bibinfo{person}{A. Nguyen}, {and} \bibinfo{person}{T. Le}.} \bibinfo{year}{2018}\natexlab{}.
\newblock \showarticletitle{Time Series Forecasting for Healthcare Diagnosis and Prognostics with the Focus on Cardiovascular Diseases}. In \bibinfo{booktitle}{\emph{International Conference on the Development of Biomedical Engineering in Vietnam (BME6)}}. \bibinfo{publisher}{Springer Singapore}, \bibinfo{address}{Singapore}, \bibinfo{pages}{809--818}.
\newblock
\showISBNx{978-981-10-4361-1}


\bibitem[Carre{\~n}o et~al\mbox{.}(2020)]%
        {carreno2020analyzing}
\bibfield{author}{\bibinfo{person}{Ander Carre{\~n}o}, \bibinfo{person}{I{\~n}aki Inza}, {and} \bibinfo{person}{Jose~A Lozano}.} \bibinfo{year}{2020}\natexlab{}.
\newblock \showarticletitle{Analyzing rare event, anomaly, novelty and outlier detection terms under the supervised classification framework}.
\newblock \bibinfo{journal}{\emph{Artificial Intelligence Review}} \bibinfo{volume}{53}, \bibinfo{number}{5} (\bibinfo{year}{2020}), \bibinfo{pages}{3575--3594}.
\newblock


\bibitem[Celik et~al\mbox{.}(2011)]%
        {Celik2011AnomalyDI}
\bibfield{author}{\bibinfo{person}{Mete Celik}, \bibinfo{person}{Filiz Dadaser‐Celik}, {and} \bibinfo{person}{Ahmet~Sakir Dokuz}.} \bibinfo{year}{2011}\natexlab{}.
\newblock \showarticletitle{Anomaly detection in temperature data using DBSCAN algorithm}.
\newblock \bibinfo{journal}{\emph{2011 International Symposium on Innovations in Intelligent Systems and Applications}} (\bibinfo{year}{2011}), \bibinfo{pages}{91--95}.
\newblock


\bibitem[Chakrabarti et~al\mbox{.}(1998)]%
        {ChakrabartiEtAl1998Mining}
\bibfield{author}{\bibinfo{person}{Soumen Chakrabarti}, \bibinfo{person}{Sunita Sarawagi}, {and} \bibinfo{person}{Byron Dom}.} \bibinfo{year}{1998}\natexlab{}.
\newblock \showarticletitle{Mining {{Surprising Patterns Using Temporal Description Length}}}. In \bibinfo{booktitle}{\emph{Proceedings of the {{International Conference}} on {{Very Large Databases}} ({{VLDB}})}} \emph{(\bibinfo{series}{{{VLDB}} '98}, Vol.~\bibinfo{volume}{24})}. \bibinfo{publisher}{{Morgan Kaufmann Publishers Inc.}}, \bibinfo{pages}{606--617}.
\newblock
\showISBNx{978-1-55860-566-4}
\urldef\tempurl%
\url{https://dl.acm.org/doi/10.5555/645924.671328}
\showURL{%
\tempurl}


\bibitem[Chalapathy and Chawla(2019)]%
        {ChalapathyChawla2019Deep}
\bibfield{author}{\bibinfo{person}{Raghavendra Chalapathy} {and} \bibinfo{person}{Sanjay Chawla}.} \bibinfo{year}{2019}\natexlab{}.
\newblock \bibinfo{booktitle}{\emph{Deep {{Learning}} for {{Anomaly Detection}}: {{A Survey}}}}.
\newblock
\showeprint[arxiv]{1901.03407}~[cs, stat]
\urldef\tempurl%
\url{http://arxiv.org/abs/1901.03407}
\showURL{%
\tempurl}


\bibitem[Chalapathy et~al\mbox{.}(2017)]%
        {deepPCA}
\bibfield{author}{\bibinfo{person}{Raghavendra Chalapathy}, \bibinfo{person}{Aditya~Krishna Menon}, {and} \bibinfo{person}{Sanjay Chawla}.} \bibinfo{year}{2017}\natexlab{}.
\newblock \showarticletitle{Robust, Deep and Inductive Anomaly Detection}. In \bibinfo{booktitle}{\emph{Machine Learning and Knowledge Discovery in Databases}}, \bibfield{editor}{\bibinfo{person}{Michelangelo Ceci}, \bibinfo{person}{Jaakko Hollm{\'e}n}, \bibinfo{person}{Ljup{\v{c}}o Todorovski}, \bibinfo{person}{Celine Vens}, {and} \bibinfo{person}{Sa{\v{s}}o D{\v{z}}eroski}} (Eds.). \bibinfo{publisher}{Springer International Publishing}, \bibinfo{address}{Cham}, \bibinfo{pages}{36--51}.
\newblock
\showISBNx{978-3-319-71249-9}


\bibitem[Chandola et~al\mbox{.}(2009a)]%
        {chandola2009anomaly}
\bibfield{author}{\bibinfo{person}{Varun Chandola}, \bibinfo{person}{Arindam Banerjee}, {and} \bibinfo{person}{Vipin Kumar}.} \bibinfo{year}{2009}\natexlab{a}.
\newblock \showarticletitle{Anomaly detection: A survey}.
\newblock \bibinfo{journal}{\emph{ACM computing surveys (CSUR)}} \bibinfo{volume}{41}, \bibinfo{number}{3} (\bibinfo{year}{2009}), \bibinfo{pages}{1--58}.
\newblock


\bibitem[Chandola et~al\mbox{.}(2009b)]%
        {ChandolaEtAl2009Anomaly}
\bibfield{author}{\bibinfo{person}{Varun Chandola}, \bibinfo{person}{Arindam Banerjee}, {and} \bibinfo{person}{Vipin Kumar}.} \bibinfo{year}{2009}\natexlab{b}.
\newblock \showarticletitle{Anomaly Detection: {{A}} Survey}.
\newblock  \bibinfo{volume}{41}, \bibinfo{number}{3} (\bibinfo{year}{2009}), \bibinfo{pages}{1--58}.
\newblock
\showISSN{0360-0300, 1557-7341}
\urldef\tempurl%
\url{https://doi.org/10.1145/1541880.1541882}
\showDOI{\tempurl}


\bibitem[Chandola et~al\mbox{.}(2012)]%
        {ChandolaEtAl2012Anomaly}
\bibfield{author}{\bibinfo{person}{V. Chandola}, \bibinfo{person}{A. Banerjee}, {and} \bibinfo{person}{V. Kumar}.} \bibinfo{year}{2012}\natexlab{}.
\newblock \showarticletitle{Anomaly {{Detection}} for {{Discrete Sequences}}: {{A Survey}}}.
\newblock  \bibinfo{volume}{24}, \bibinfo{number}{5} (\bibinfo{year}{2012}), \bibinfo{pages}{823--839}.
\newblock
\showISSN{1041-4347}
\urldef\tempurl%
\url{https://doi.org/10.1109/TKDE.2010.235}
\showDOI{\tempurl}


\bibitem[Chang et~al\mbox{.}(1988)]%
        {chang1988estimation}
\bibfield{author}{\bibinfo{person}{Ih Chang}, \bibinfo{person}{George~C Tiao}, {and} \bibinfo{person}{Chung Chen}.} \bibinfo{year}{1988}\natexlab{}.
\newblock \showarticletitle{Estimation of time series parameters in the presence of outliers}.
\newblock \bibinfo{journal}{\emph{Technometrics}} \bibinfo{volume}{30}, \bibinfo{number}{2} (\bibinfo{year}{1988}), \bibinfo{pages}{193--204}.
\newblock


\bibitem[Chang et~al\mbox{.}(2009)]%
        {ChangEtAl2009MoteBased}
\bibfield{author}{\bibinfo{person}{Marcus Chang}, \bibinfo{person}{Andreas Terzis}, {and} \bibinfo{person}{Philippe Bonnet}.} \bibinfo{year}{2009}\natexlab{}.
\newblock \showarticletitle{Mote-{{Based Online Anomaly Detection Using Echo State Networks}}}. In \bibinfo{booktitle}{\emph{Proceedings of the {{International Conference}} on {{Distributed Computing}} in {{Sensor Systems}} ({{DCOOS}})}} \emph{(\bibinfo{series}{Lecture {{Notes}} in {{Computer Science}}}, Vol.~\bibinfo{volume}{5516})}, \bibfield{editor}{\bibinfo{person}{Bhaskar Krishnamachari}, \bibinfo{person}{Subhash Suri}, \bibinfo{person}{Wendi Heinzelman}, {and} \bibinfo{person}{Urbashi Mitra}} (Eds.). \bibinfo{publisher}{{Springer Berlin Heidelberg}}, \bibinfo{pages}{72--86}.
\newblock
\showISBNx{978-3-642-02084-1 978-3-642-02085-8}
\urldef\tempurl%
\url{https://doi.org/10.1007/978-3-642-02085-8_6}
\showDOI{\tempurl}


\bibitem[Chauhan and Vig(2015)]%
        {ChauhanVig2015Anomaly}
\bibfield{author}{\bibinfo{person}{S. Chauhan} {and} \bibinfo{person}{L. Vig}.} \bibinfo{year}{2015}\natexlab{}.
\newblock \showarticletitle{Anomaly Detection in {{ECG}} Time Signals via Deep Long Short-Term Memory Networks}. In \bibinfo{booktitle}{\emph{Proceedings of the {{International Conference}} on {{Data Science}} and {{Advanced Analytics}} ({{DSAA}})}}. \bibinfo{pages}{1--7}.
\newblock
\urldef\tempurl%
\url{https://doi.org/10.1109/DSAA.2015.7344872}
\showDOI{\tempurl}


\bibitem[Chen et~al\mbox{.}(2020)]%
        {ChenEtAl2020Imbalanced}
\bibfield{author}{\bibinfo{person}{Qing Chen}, \bibinfo{person}{Anguo Zhang}, \bibinfo{person}{Tingwen Huang}, \bibinfo{person}{Qianping He}, {and} \bibinfo{person}{Yongduan Song}.} \bibinfo{year}{2020}\natexlab{}.
\newblock \showarticletitle{Imbalanced Dataset-Based Echo State Networks for Anomaly Detection}.
\newblock  \bibinfo{volume}{32}, \bibinfo{number}{8} (\bibinfo{year}{2020}), \bibinfo{pages}{3685--3694}.
\newblock
\showISSN{0941-0643, 1433-3058}
\urldef\tempurl%
\url{https://doi.org/10.1007/s00521-018-3747-z}
\showDOI{\tempurl}


\bibitem[Chen et~al\mbox{.}(2021)]%
        {chen2021joint}
\bibfield{author}{\bibinfo{person}{Run-Qing Chen}, \bibinfo{person}{Guang-Hui Shi}, \bibinfo{person}{Wan-Lei Zhao}, {and} \bibinfo{person}{Chang-Hui Liang}.} \bibinfo{year}{2021}\natexlab{}.
\newblock \showarticletitle{A joint model for IT operation series prediction and anomaly detection}.
\newblock \bibinfo{journal}{\emph{Neurocomputing}}  \bibinfo{volume}{448} (\bibinfo{year}{2021}), \bibinfo{pages}{130--139}.
\newblock


\bibitem[Cheng et~al\mbox{.}(2019)]%
        {ChengEtAl2019Outlier}
\bibfield{author}{\bibinfo{person}{Zhangyu Cheng}, \bibinfo{person}{Chengming Zou}, {and} \bibinfo{person}{Jianwei Dong}.} \bibinfo{year}{2019}\natexlab{}.
\newblock \showarticletitle{Outlier Detection Using Isolation Forest and Local Outlier Factor}. In \bibinfo{booktitle}{\emph{Proceedings of the {{Conference}} on {{Research}} in {{Adaptive}} and {{Convergent Systems}} ({{RACS}})}}. \bibinfo{publisher}{{ACM}}, \bibinfo{pages}{161--168}.
\newblock
\showISBNx{978-1-4503-6843-8}
\urldef\tempurl%
\url{https://doi.org/10.1145/3338840.3355641}
\showDOI{\tempurl}


\bibitem[Choudhary et~al\mbox{.}(2017)]%
        {ChoudharyEtAl2017RuntimeEfficacy}
\bibfield{author}{\bibinfo{person}{Dhruv Choudhary}, \bibinfo{person}{Arun Kejariwal}, {and} \bibinfo{person}{Francois Orsini}.} \bibinfo{year}{2017}\natexlab{}.
\newblock \bibinfo{booktitle}{\emph{On the {{Runtime}}-{{Efficacy Trade}}-off of {{Anomaly Detection Techniques}} for {{Real}}-{{Time Streaming Data}}}}.
\newblock
\showeprint[arxiv]{1710.04735}~[cs, eess, stat]
\urldef\tempurl%
\url{http://arxiv.org/abs/1710.04735}
\showURL{%
\tempurl}


\bibitem[Cohen(1995)]%
        {COHEN1995115}
\bibfield{author}{\bibinfo{person}{William~W. Cohen}.} \bibinfo{year}{1995}\natexlab{}.
\newblock \showarticletitle{Fast Effective Rule Induction}.
\newblock In \bibinfo{booktitle}{\emph{Machine Learning Proceedings 1995}}, \bibfield{editor}{\bibinfo{person}{Armand Prieditis} {and} \bibinfo{person}{Stuart Russell}} (Eds.). \bibinfo{publisher}{Morgan Kaufmann}, \bibinfo{address}{San Francisco (CA)}, \bibinfo{pages}{115--123}.
\newblock
\showISBNx{978-1-55860-377-6}
\urldef\tempurl%
\url{https://doi.org/10.1016/B978-1-55860-377-6.50023-2}
\showDOI{\tempurl}


\bibitem[Cook et~al\mbox{.}(2020)]%
        {CookEtAl2020Anomaly}
\bibfield{author}{\bibinfo{person}{Andrew~A. Cook}, \bibinfo{person}{Goksel Misirli}, {and} \bibinfo{person}{Zhong Fan}.} \bibinfo{year}{2020}\natexlab{}.
\newblock \showarticletitle{Anomaly {{Detection}} for {{IoT Time}}-{{Series Data}}: {{A Survey}}}.
\newblock  \bibinfo{volume}{7}, \bibinfo{number}{7} (\bibinfo{year}{2020}), \bibinfo{pages}{6481--6494}.
\newblock
\showISSN{2327-4662, 2372-2541}
\urldef\tempurl%
\url{https://doi.org/10.1109/JIOT.2019.2958185}
\showDOI{\tempurl}


\bibitem[Costa et~al\mbox{.}(2002)]%
        {costa2002multiscale}
\bibfield{author}{\bibinfo{person}{Madalena Costa}, \bibinfo{person}{Ary~L Goldberger}, {and} \bibinfo{person}{C-K Peng}.} \bibinfo{year}{2002}\natexlab{}.
\newblock \showarticletitle{Multiscale entropy analysis of complex physiologic time series}.
\newblock \bibinfo{journal}{\emph{Physical review letters}} \bibinfo{volume}{89}, \bibinfo{number}{6} (\bibinfo{year}{2002}), \bibinfo{pages}{068102}.
\newblock


\bibitem[Davis and Goadrich(2006)]%
        {10.1145/1143844.1143874}
\bibfield{author}{\bibinfo{person}{Jesse Davis} {and} \bibinfo{person}{Mark Goadrich}.} \bibinfo{year}{2006}\natexlab{}.
\newblock \showarticletitle{The relationship between Precision-Recall and ROC curves}. In \bibinfo{booktitle}{\emph{ICML}}. \bibinfo{pages}{233--240}.
\newblock


\bibitem[Ding et~al\mbox{.}(2018)]%
        {DingEtAl2018MultivariateTimeSeriesDriven}
\bibfield{author}{\bibinfo{person}{Nan Ding}, \bibinfo{person}{Huanbo Gao}, \bibinfo{person}{Hongyu Bu}, \bibinfo{person}{Haoxuan Ma}, {and} \bibinfo{person}{Huaiwei Si}.} \bibinfo{year}{2018}\natexlab{}.
\newblock \showarticletitle{Multivariate-{{Time}}-{{Series}}-{{Driven Real}}-Time {{Anomaly Detection Based}} on {{Bayesian Network}}}.
\newblock  \bibinfo{volume}{18}, \bibinfo{number}{10} (\bibinfo{year}{2018}), \bibinfo{pages}{3367}.
\newblock
\showISSN{1424-8220}
\urldef\tempurl%
\url{https://doi.org/10.3390/s18103367}
\showDOI{\tempurl}


\bibitem[Dixon(1950)]%
        {dixon1950analysis}
\bibfield{author}{\bibinfo{person}{Wilfred~J Dixon}.} \bibinfo{year}{1950}\natexlab{}.
\newblock \showarticletitle{Analysis of extreme values}.
\newblock \bibinfo{journal}{\emph{The Annals of Mathematical Statistics}} \bibinfo{volume}{21}, \bibinfo{number}{4} (\bibinfo{year}{1950}), \bibinfo{pages}{488--506}.
\newblock


\bibitem[Dziedzic et~al\mbox{.}(2019)]%
        {dziedzic2019band}
\bibfield{author}{\bibinfo{person}{Adam Dziedzic}, \bibinfo{person}{John Paparrizos}, \bibinfo{person}{Sanjay Krishnan}, \bibinfo{person}{Aaron Elmore}, {and} \bibinfo{person}{Michael Franklin}.} \bibinfo{year}{2019}\natexlab{}.
\newblock \showarticletitle{Band-limited training and inference for convolutional neural networks}. In \bibinfo{booktitle}{\emph{ICML}}. PMLR, \bibinfo{pages}{1745--1754}.
\newblock


\bibitem[d’Hondt et~al\mbox{.}(2024)]%
        {d2024beyond}
\bibfield{author}{\bibinfo{person}{Jens~E d’Hondt}, \bibinfo{person}{Odysseas Papapetrou}, {and} \bibinfo{person}{John Paparrizos}.} \bibinfo{year}{2024}\natexlab{}.
\newblock \showarticletitle{Beyond the Dimensions: A Structured Evaluation of Multivariate Time Series Distance Measures}. In \bibinfo{booktitle}{\emph{2024 IEEE 40th International Conference on Data Engineering Workshops (ICDEW)}}. IEEE, \bibinfo{pages}{107--112}.
\newblock


\bibitem[Edgeworth(1887)]%
        {edgeworth1887xli}
\bibfield{author}{\bibinfo{person}{Francis~Ysidro Edgeworth}.} \bibinfo{year}{1887}\natexlab{}.
\newblock \showarticletitle{Xli. on discordant observations}.
\newblock \bibinfo{journal}{\emph{The london, edinburgh, and dublin philosophical magazine and journal of science}} \bibinfo{volume}{23}, \bibinfo{number}{143} (\bibinfo{year}{1887}), \bibinfo{pages}{364--375}.
\newblock


\bibitem[Ernst and Bar-Joseph(2006)]%
        {ernst2006stem}
\bibfield{author}{\bibinfo{person}{Jason Ernst} {and} \bibinfo{person}{Ziv Bar-Joseph}.} \bibinfo{year}{2006}\natexlab{}.
\newblock \showarticletitle{STEM: a tool for the analysis of short time series gene expression data}.
\newblock \bibinfo{journal}{\emph{BMC bioinformatics}} \bibinfo{volume}{7}, \bibinfo{number}{1} (\bibinfo{year}{2006}), \bibinfo{pages}{191}.
\newblock


\bibitem[Esling and Agon(2012)]%
        {esling2012time}
\bibfield{author}{\bibinfo{person}{Philippe Esling} {and} \bibinfo{person}{Carlos Agon}.} \bibinfo{year}{2012}\natexlab{}.
\newblock \showarticletitle{Time-series data mining}.
\newblock \bibinfo{journal}{\emph{ACM Computing Surveys (CSUR)}} \bibinfo{volume}{45}, \bibinfo{number}{1} (\bibinfo{year}{2012}), \bibinfo{pages}{12}.
\newblock


\bibitem[Fawcett(2006)]%
        {FAWCETT2006861}
\bibfield{author}{\bibinfo{person}{Tom Fawcett}.} \bibinfo{year}{2006}\natexlab{}.
\newblock \showarticletitle{An introduction to ROC analysis}.
\newblock \bibinfo{journal}{\emph{Pattern Recognition Letters}} \bibinfo{volume}{27}, \bibinfo{number}{8} (\bibinfo{year}{2006}), \bibinfo{pages}{861--874}.
\newblock
\showISSN{0167-8655}
\urldef\tempurl%
\url{https://doi.org/10.1016/j.patrec.2005.10.010}
\showDOI{\tempurl}
\newblock
\shownote{ROC Analysis in Pattern Recognition}.


\bibitem[Filonov et~al\mbox{.}(2016)]%
        {LSTM_ON_GHL}
\bibfield{author}{\bibinfo{person}{Pavel Filonov}, \bibinfo{person}{Andrey Lavrentyev}, {and} \bibinfo{person}{Artem Vorontsov}.} \bibinfo{year}{2016}\natexlab{}.
\newblock \showarticletitle{Multivariate industrial time series with cyber-attack simulation: Fault detection using an lstm-based predictive data model}.
\newblock \bibinfo{journal}{\emph{arXiv preprint arXiv:1612.06676}} (\bibinfo{year}{2016}).
\newblock


\bibitem[Foorthuis(2020)]%
        {foorthuis2020nature}
\bibfield{author}{\bibinfo{person}{Ralph Foorthuis}.} \bibinfo{year}{2020}\natexlab{}.
\newblock \showarticletitle{On the Nature and Types of Anomalies: A Review}.
\newblock \bibinfo{journal}{\emph{arXiv preprint arXiv:2007.15634}} (\bibinfo{year}{2020}).
\newblock


\bibitem[Fox(1972)]%
        {fox1972outliers}
\bibfield{author}{\bibinfo{person}{Anthony~J Fox}.} \bibinfo{year}{1972}\natexlab{}.
\newblock \showarticletitle{Outliers in time series}.
\newblock \bibinfo{journal}{\emph{Journal of the Royal Statistical Society: Series B (Methodological)}} \bibinfo{volume}{34}, \bibinfo{number}{3} (\bibinfo{year}{1972}), \bibinfo{pages}{350--363}.
\newblock


\bibitem[Fu et~al\mbox{.}(2006)]%
        {DBLP:conf/adma/FuLKL06}
\bibfield{author}{\bibinfo{person}{Ada~Wai{-}Chee Fu}, \bibinfo{person}{Oscar~Tat{-}Wing Leung}, \bibinfo{person}{Eamonn~J. Keogh}, {and} \bibinfo{person}{Jessica Lin}.} \bibinfo{year}{2006}\natexlab{}.
\newblock \showarticletitle{Finding Time Series Discords Based on Haar Transform}. In \bibinfo{booktitle}{\emph{ADMA}}.
\newblock


\bibitem[Gao et~al\mbox{.}(2020)]%
        {GaoEtAl2020Ensemble}
\bibfield{author}{\bibinfo{person}{Yifeng Gao}, \bibinfo{person}{Jessica Lin}, {and} \bibinfo{person}{Constantin Brif}.} \bibinfo{year}{2020}\natexlab{}.
\newblock \showarticletitle{Ensemble {{Grammar Induction For Detecting Anomalies}} in {{Time Series}}}. In \bibinfo{booktitle}{\emph{Proceedings of the {{International Conference}} on {{Extending Database Technology}} ({{EDBT}})}}.
\newblock
\urldef\tempurl%
\url{https://doi.org/10.5441/002/edbt.2020.09}
\showDOI{\tempurl}


\bibitem[Garcia et~al\mbox{.}(2020b)]%
        {cae2000}
\bibfield{author}{\bibinfo{person}{Gabriel Garcia}, \bibinfo{person}{Gabriel Michau}, \bibinfo{person}{Mélanie Ducoffe}, \bibinfo{person}{Jayant Sen~Gupta}, {and} \bibinfo{person}{Olga Fink}.} \bibinfo{year}{2020}\natexlab{b}.
\newblock \showarticletitle{Time Series to Images: Monitoring the Condition of Industrial Assets with Deep Learning Image Processing Algorithms}.
\newblock  (\bibinfo{date}{05} \bibinfo{year}{2020}).
\newblock


\bibitem[Garcia et~al\mbox{.}(2020a)]%
        {GarciaEtAl2020Time}
\bibfield{author}{\bibinfo{person}{Gabriel~Rodriguez Garcia}, \bibinfo{person}{Gabriel Michau}, \bibinfo{person}{Mélanie Ducoffe}, \bibinfo{person}{Jayant~Sen Gupta}, {and} \bibinfo{person}{Olga Fink}.} \bibinfo{year}{2020}\natexlab{a}.
\newblock \bibinfo{booktitle}{\emph{Time {{Series}} to {{Images}}: {{Monitoring}} the {{Condition}} of {{Industrial Assets}} with {{Deep Learning Image Processing Algorithms}}}}.
\newblock
\showeprint[arxiv]{2005.07031}~[cs, eess, stat]
\urldef\tempurl%
\url{http://arxiv.org/abs/2005.07031}
\showURL{%
\tempurl}


\bibitem[Gavrilov et~al\mbox{.}(2000)]%
        {gavrilov2000mining}
\bibfield{author}{\bibinfo{person}{Martin Gavrilov}, \bibinfo{person}{Dragomir Anguelov}, \bibinfo{person}{Piotr Indyk}, {and} \bibinfo{person}{Rajeev Motwani}.} \bibinfo{year}{2000}\natexlab{}.
\newblock \showarticletitle{Mining the stock market: Which measure is best}. In \bibinfo{booktitle}{\emph{Proc. of the 6th ACM SIGKDD}}. \bibinfo{pages}{487--496}.
\newblock


\bibitem[George(2019)]%
        {iotstats}
\bibfield{author}{\bibinfo{person}{Sam George}.} \bibinfo{year}{2019}\natexlab{}.
\newblock \bibinfo{booktitle}{\emph{{IoT Signals report: IoT's promise will be unlocked by addressing skills shortage, complexity and security.}}}
\newblock
\newblock
\shownote{\url{https://blogs.microsoft.com/blog/2019/07/30/}}.


\bibitem[Glaisher(1873)]%
        {glaisher1873rejection}
\bibfield{author}{\bibinfo{person}{JWL Glaisher}.} \bibinfo{year}{1873}\natexlab{}.
\newblock \showarticletitle{On the rejection of discordant observations}.
\newblock \bibinfo{journal}{\emph{Monthly Notices of the Royal Astronomical Society}}  \bibinfo{volume}{33} (\bibinfo{year}{1873}), \bibinfo{pages}{391--402}.
\newblock


\bibitem[Goddard et~al\mbox{.}(2003)]%
        {goddard2003geospatial}
\bibfield{author}{\bibinfo{person}{Steve Goddard}, \bibinfo{person}{Sherri~K Harms}, \bibinfo{person}{Stephen~E Reichenbach}, \bibinfo{person}{Tsegaye Tadesse}, {and} \bibinfo{person}{William~J Waltman}.} \bibinfo{year}{2003}\natexlab{}.
\newblock \showarticletitle{Geospatial decision support for drought risk management}.
\newblock \bibinfo{journal}{\emph{Commun. ACM}} \bibinfo{volume}{46}, \bibinfo{number}{1} (\bibinfo{year}{2003}), \bibinfo{pages}{35--37}.
\newblock


\bibitem[Goel et~al\mbox{.}(2016)]%
        {goel2016social}
\bibfield{author}{\bibinfo{person}{Rahul Goel}, \bibinfo{person}{Sandeep Soni}, \bibinfo{person}{Naman Goyal}, \bibinfo{person}{John Paparrizos}, \bibinfo{person}{Hanna Wallach}, \bibinfo{person}{Fernando Diaz}, {and} \bibinfo{person}{Jacob Eisenstein}.} \bibinfo{year}{2016}\natexlab{}.
\newblock \showarticletitle{The social dynamics of language change in online networks}. In \bibinfo{booktitle}{\emph{Social Informatics: 8th International Conference, SocInfo 2016, Bellevue, WA, USA, November 11-14, 2016, Proceedings, Part I 8}}. Springer, \bibinfo{pages}{41--57}.
\newblock


\bibitem[Goldstein and Dengel(2013)]%
        {Goldstein_histogram-basedoutlier}
\bibfield{author}{\bibinfo{person}{Markus Goldstein} {and} \bibinfo{person}{Andreas Dengel}.} \bibinfo{year}{2013}\natexlab{}.
\newblock \bibinfo{title}{Histogram-based Outlier Score (HBOS): A fast Unsupervised Anomaly Detection Algorithm}.
\newblock
\newblock


\bibitem[Goldstein and Uchida(2016)]%
        {goldstein2016comparative}
\bibfield{author}{\bibinfo{person}{Markus Goldstein} {and} \bibinfo{person}{Seiichi Uchida}.} \bibinfo{year}{2016}\natexlab{}.
\newblock \showarticletitle{A comparative evaluation of unsupervised anomaly detection algorithms for multivariate data}.
\newblock \bibinfo{journal}{\emph{PloS one}} \bibinfo{volume}{11}, \bibinfo{number}{4} (\bibinfo{year}{2016}), \bibinfo{pages}{e0152173}.
\newblock


\bibitem[G{\'o}mez-Verdejo et~al\mbox{.}(2011)]%
        {Gomez-VerdejoEtAl2011Adaptive}
\bibfield{author}{\bibinfo{person}{Vanessa G{\'o}mez-Verdejo}, \bibinfo{person}{Jer{\'o}nimo Arenas-Garc{\'\i}a}, \bibinfo{person}{Miguel Lazaro-Gredilla}, {and} \bibinfo{person}{{\'A}ngel Navia-Vazquez}.} \bibinfo{year}{2011}\natexlab{}.
\newblock \showarticletitle{Adaptive one-class support vector machine}.
\newblock \bibinfo{journal}{\emph{IEEE Transactions on Signal Processing}} \bibinfo{volume}{59}, \bibinfo{number}{6} (\bibinfo{year}{2011}), \bibinfo{pages}{2975--2981}.
\newblock


\bibitem[Goodfellow et~al\mbox{.}(2014)]%
        {NIPS2014_5423}
\bibfield{author}{\bibinfo{person}{Ian Goodfellow}, \bibinfo{person}{Jean Pouget-Abadie}, \bibinfo{person}{Mehdi Mirza}, \bibinfo{person}{Bing Xu}, \bibinfo{person}{David Warde-Farley}, \bibinfo{person}{Sherjil Ozair}, \bibinfo{person}{Aaron Courville}, {and} \bibinfo{person}{Yoshua Bengio}.} \bibinfo{year}{2014}\natexlab{}.
\newblock \showarticletitle{Generative adversarial nets}.
\newblock \bibinfo{journal}{\emph{NeurIPS}}  \bibinfo{volume}{27} (\bibinfo{year}{2014}).
\newblock


\bibitem[Gould(1855)]%
        {gould1855peirce}
\bibfield{author}{\bibinfo{person}{BA Gould}.} \bibinfo{year}{1855}\natexlab{}.
\newblock \showarticletitle{On Peirce's Criterion for the Rejection of Doubtful Observations, with tables for facilitating its application}.
\newblock \bibinfo{journal}{\emph{The Astronomical Journal}}  \bibinfo{volume}{4} (\bibinfo{year}{1855}), \bibinfo{pages}{81--87}.
\newblock


\bibitem[Grover et~al\mbox{.}(2015)]%
        {grover2015deep}
\bibfield{author}{\bibinfo{person}{Aditya Grover}, \bibinfo{person}{Ashish Kapoor}, {and} \bibinfo{person}{Eric Horvitz}.} \bibinfo{year}{2015}\natexlab{}.
\newblock \showarticletitle{A deep hybrid model for weather forecasting}. In \bibinfo{booktitle}{\emph{Proceedings of the 21th ACM SIGKDD International Conference on Knowledge Discovery and Data Mining}}. ACM, \bibinfo{pages}{379--386}.
\newblock


\bibitem[Grubbs(1950)]%
        {grubbs1950sample}
\bibfield{author}{\bibinfo{person}{Frank~E Grubbs}.} \bibinfo{year}{1950}\natexlab{}.
\newblock \showarticletitle{Sample criteria for testing outlying observations}.
\newblock \bibinfo{journal}{\emph{The Annals of Mathematical Statistics}} (\bibinfo{year}{1950}), \bibinfo{pages}{27--58}.
\newblock


\bibitem[Gupta et~al\mbox{.}(2014)]%
        {GuptaEtAl2014Outlier}
\bibfield{author}{\bibinfo{person}{Manish Gupta}, \bibinfo{person}{Jing Gao}, \bibinfo{person}{Charu~C. Aggarwal}, {and} \bibinfo{person}{Jiawei Han}.} \bibinfo{year}{2014}\natexlab{}.
\newblock \showarticletitle{Outlier {{Detection}} for {{Temporal Data}}: {{A Survey}}}.
\newblock  \bibinfo{volume}{26}, \bibinfo{number}{9} (\bibinfo{year}{2014}), \bibinfo{pages}{2250--2267}.
\newblock
\showISSN{1041-4347}
\urldef\tempurl%
\url{https://doi.org/10.1109/TKDE.2013.184}
\showDOI{\tempurl}


\bibitem[Görnitz et~al\mbox{.}(2015)]%
        {GornitzEtAl2015Hidden}
\bibfield{author}{\bibinfo{person}{Nico Görnitz}, \bibinfo{person}{Mikio Braun}, {and} \bibinfo{person}{Marius Kloft}.} \bibinfo{year}{2015}\natexlab{}.
\newblock \showarticletitle{Hidden {{Markov}} Anomaly Detection}. In \bibinfo{booktitle}{\emph{Proceedings of the {{International Conference}} on {{Machine Learning}} ({{ICML}})}} \emph{(\bibinfo{series}{{{ICML}}'15})}. \bibinfo{publisher}{{JMLR.org}}, \bibinfo{pages}{1833--1842}.
\newblock
\urldef\tempurl%
\url{https://dl.acm.org/doi/10.5555/3045118.3045313}
\showURL{%
\tempurl}


\bibitem[Hahsler and Bolaos(2016)]%
        {10.1109/TKDE.2016.2522412}
\bibfield{author}{\bibinfo{person}{Michael Hahsler} {and} \bibinfo{person}{Matthew Bolaos}.} \bibinfo{year}{2016}\natexlab{}.
\newblock \showarticletitle{Clustering Data Streams Based on Shared Density between Micro-Clusters}.
\newblock \bibinfo{journal}{\emph{IEEE Trans. on Knowl. and Data Eng.}} \bibinfo{volume}{28}, \bibinfo{number}{6} (\bibinfo{date}{jun} \bibinfo{year}{2016}), \bibinfo{pages}{1449–1461}.
\newblock
\showISSN{1041-4347}
\urldef\tempurl%
\url{https://doi.org/10.1109/TKDE.2016.2522412}
\showDOI{\tempurl}


\bibitem[Hardin and Rocke(2004)]%
        {HARDIN2004625}
\bibfield{author}{\bibinfo{person}{Johanna Hardin} {and} \bibinfo{person}{David~M Rocke}.} \bibinfo{year}{2004}\natexlab{}.
\newblock \showarticletitle{Outlier detection in the multiple cluster setting using the minimum covariance determinant estimator}.
\newblock \bibinfo{journal}{\emph{Computational Statistics \& Data Analysis}} \bibinfo{volume}{44}, \bibinfo{number}{4} (\bibinfo{year}{2004}), \bibinfo{pages}{625 -- 638}.
\newblock
\showISSN{0167-9473}
\urldef\tempurl%
\url{https://doi.org/10.1016/S0167-9473(02)00280-3}
\showDOI{\tempurl}


\bibitem[Hariri et~al\mbox{.}(2019)]%
        {HaririEtAl2019Extended}
\bibfield{author}{\bibinfo{person}{Sahand Hariri}, \bibinfo{person}{Matias~Carrasco Kind}, {and} \bibinfo{person}{Robert~J. Brunner}.} \bibinfo{year}{2019}\natexlab{}.
\newblock \showarticletitle{Extended {{Isolation Forest}}}.
\newblock  (\bibinfo{year}{2019}).
\newblock
\showISSN{1041-4347, 1558-2191, 2326-3865}
\urldef\tempurl%
\url{https://doi.org/10.1109/TKDE.2019.2947676}
\showDOI{\tempurl}
\showeprint[arxiv]{1811.02141}


\bibitem[Hawkins(1980)]%
        {hawkins_identification_1980}
\bibfield{author}{\bibinfo{person}{D.~M Hawkins}.} \bibinfo{year}{1980}\natexlab{}.
\newblock \bibinfo{booktitle}{\emph{Identification of {Outliers}.}}
\newblock \bibinfo{publisher}{Springer Netherlands}, \bibinfo{address}{Dordrecht}.
\newblock
\showISBNx{9789401539944}
\newblock
\shownote{OCLC: 945065134}.


\bibitem[Hawkins and Dawkins(2021)]%
        {Athousandbrains}
\bibfield{author}{\bibinfo{person}{Jeff Hawkins} {and} \bibinfo{person}{Richard Dawkins}.} \bibinfo{year}{2021}\natexlab{}.
\newblock \showarticletitle{A thousand brains : a new theory of intelligence}.
\newblock  (\bibinfo{year}{2021}).
\newblock
\showISBNx{9781541675803; 1541675800}


\bibitem[He et~al\mbox{.}(2003)]%
        {HeEtAl2003Discovering}
\bibfield{author}{\bibinfo{person}{Zengyou He}, \bibinfo{person}{Xiaofei Xu}, {and} \bibinfo{person}{Shengchun Deng}.} \bibinfo{year}{2003}\natexlab{}.
\newblock \showarticletitle{Discovering cluster-based local outliers}.
\newblock \bibinfo{journal}{\emph{Pattern recognition letters}} \bibinfo{volume}{24}, \bibinfo{number}{9-10} (\bibinfo{year}{2003}), \bibinfo{pages}{1641--1650}.
\newblock


\bibitem[{Hearst} et~al\mbox{.}(1998)]%
        {708428}
\bibfield{author}{\bibinfo{person}{M.~A. {Hearst}}, \bibinfo{person}{S.~T. {Dumais}}, \bibinfo{person}{E. {Osuna}}, \bibinfo{person}{J. {Platt}}, {and} \bibinfo{person}{B. {Scholkopf}}.} \bibinfo{year}{1998}\natexlab{}.
\newblock \showarticletitle{Support vector machines}.
\newblock \bibinfo{journal}{\emph{IEEE Intelligent Systems and their Applications}} \bibinfo{volume}{13}, \bibinfo{number}{4} (\bibinfo{date}{July} \bibinfo{year}{1998}), \bibinfo{pages}{18--28}.
\newblock
\showISSN{1094-7167}
\urldef\tempurl%
\url{https://doi.org/10.1109/5254.708428}
\showDOI{\tempurl}


\bibitem[Hebb(1949)]%
        {hebb-organization-of-behavior-1949}
\bibfield{author}{\bibinfo{person}{Donald~O. Hebb}.} \bibinfo{year}{1949}\natexlab{}.
\newblock \bibinfo{booktitle}{\emph{The organization of behavior: {A} neuropsychological theory}}.
\newblock \bibinfo{publisher}{Wiley}, \bibinfo{address}{New York}.
\newblock
\showISBNx{0-8058-4300-0}


\bibitem[Heim and Avery(2019)]%
        {HeimAvery2019Adaptive}
\bibfield{author}{\bibinfo{person}{Niklas Heim} {and} \bibinfo{person}{James~E. Avery}.} \bibinfo{year}{2019}\natexlab{}.
\newblock \bibinfo{booktitle}{\emph{Adaptive {{Anomaly Detection}} in {{Chaotic Time Series}} with a {{Spatially Aware Echo State Network}}}}.
\newblock
\showeprint[arxiv]{1909.01709}~[cs, stat]
\urldef\tempurl%
\url{http://arxiv.org/abs/1909.01709}
\showURL{%
\tempurl}


\bibitem[Hochenbaum et~al\mbox{.}(2017)]%
        {HochenbaumEtAl2017Automatic}
\bibfield{author}{\bibinfo{person}{Jordan Hochenbaum}, \bibinfo{person}{Owen~S. Vallis}, {and} \bibinfo{person}{Arun Kejariwal}.} \bibinfo{year}{2017}\natexlab{}.
\newblock \bibinfo{booktitle}{\emph{Automatic {{Anomaly Detection}} in the {{Cloud Via Statistical Learning}}}}.
\newblock
\showeprint[arxiv]{1704.07706}~[cs]
\urldef\tempurl%
\url{http://arxiv.org/abs/1704.07706}
\showURL{%
\tempurl}


\bibitem[Hochreiter and Schmidhuber(1997)]%
        {Hochreiter:1997:LSM:1246443.1246450}
\bibfield{author}{\bibinfo{person}{Sepp Hochreiter} {and} \bibinfo{person}{J\"{u}rgen Schmidhuber}.} \bibinfo{year}{1997}\natexlab{}.
\newblock \showarticletitle{Long Short-Term Memory}.
\newblock \bibinfo{journal}{\emph{Neural Comput.}} \bibinfo{volume}{9}, \bibinfo{number}{8} (\bibinfo{date}{Nov.} \bibinfo{year}{1997}), \bibinfo{pages}{1735--1780}.
\newblock
\showISSN{0899-7667}
\urldef\tempurl%
\url{https://doi.org/10.1162/neco.1997.9.8.1735}
\showDOI{\tempurl}


\bibitem[Hodge and Austin(2004)]%
        {HodgeAustin2004Survey}
\bibfield{author}{\bibinfo{person}{Victoria~J. Hodge} {and} \bibinfo{person}{Jim Austin}.} \bibinfo{year}{2004}\natexlab{}.
\newblock \showarticletitle{A {{Survey}} of {{Outlier Detection Methodologies}}}.
\newblock  \bibinfo{volume}{22}, \bibinfo{number}{2} (\bibinfo{year}{2004}), \bibinfo{pages}{85--126}.
\newblock
\showISSN{1573-7462}
\urldef\tempurl%
\url{https://doi.org/10.1007/s10462-004-4304-y}
\showDOI{\tempurl}


\bibitem[Hoegh-Guldberg et~al\mbox{.}(2007)]%
        {hoegh2007coral}
\bibfield{author}{\bibinfo{person}{Ove Hoegh-Guldberg}, \bibinfo{person}{Peter~J Mumby}, \bibinfo{person}{Anthony~J Hooten}, \bibinfo{person}{Robert~S Steneck}, \bibinfo{person}{Paul Greenfield}, \bibinfo{person}{Edgardo Gomez}, \bibinfo{person}{C~Drew Harvell}, \bibinfo{person}{Peter~F Sale}, \bibinfo{person}{Alasdair~J Edwards}, \bibinfo{person}{Ken Caldeira}, {et~al\mbox{.}}} \bibinfo{year}{2007}\natexlab{}.
\newblock \showarticletitle{Coral reefs under rapid climate change and ocean acidification}.
\newblock \bibinfo{journal}{\emph{science}} \bibinfo{volume}{318}, \bibinfo{number}{5857} (\bibinfo{year}{2007}), \bibinfo{pages}{1737--1742}.
\newblock


\bibitem[Honda et~al\mbox{.}(2002)]%
        {honda2002mining}
\bibfield{author}{\bibinfo{person}{Rie Honda}, \bibinfo{person}{Shuai Wang}, \bibinfo{person}{Tokio Kikuchi}, {and} \bibinfo{person}{Osamu Konishi}.} \bibinfo{year}{2002}\natexlab{}.
\newblock \showarticletitle{Mining of moving objects from time-series images and its application to satellite weather imagery}.
\newblock \bibinfo{journal}{\emph{Journal of Intelligent Information Systems}} \bibinfo{volume}{19}, \bibinfo{number}{1} (\bibinfo{year}{2002}), \bibinfo{pages}{79--93}.
\newblock


\bibitem[Huijse et~al\mbox{.}(2014)]%
        {huijse2014computational}
\bibfield{author}{\bibinfo{person}{Pablo Huijse}, \bibinfo{person}{Pablo~A Estevez}, \bibinfo{person}{Pavlos Protopapas}, \bibinfo{person}{Jose~C Principe}, {and} \bibinfo{person}{Pablo Zegers}.} \bibinfo{year}{2014}\natexlab{}.
\newblock \showarticletitle{Computational intelligence challenges and applications on large-scale astronomical time series databases}.
\newblock \bibinfo{journal}{\emph{IEEE Computational Intelligence Magazine}} \bibinfo{volume}{9}, \bibinfo{number}{3} (\bibinfo{year}{2014}), \bibinfo{pages}{27--39}.
\newblock


\bibitem[Hundman et~al\mbox{.}(2018a)]%
        {HundmanEtAl2018Detecting}
\bibfield{author}{\bibinfo{person}{Kyle Hundman}, \bibinfo{person}{Valentino Constantinou}, \bibinfo{person}{Christopher Laporte}, \bibinfo{person}{Ian Colwell}, {and} \bibinfo{person}{Tom Soderstrom}.} \bibinfo{year}{2018}\natexlab{a}.
\newblock \showarticletitle{Detecting {{Spacecraft Anomalies Using LSTMs}} and {{Nonparametric Dynamic Thresholding}}}. In \bibinfo{booktitle}{\emph{SIGKDD}}. \bibinfo{publisher}{{ACM}}, \bibinfo{pages}{387--395}.
\newblock
\showISBNx{978-1-4503-5552-0}
\urldef\tempurl%
\url{https://doi.org/10.1145/3219819.3219845}
\showDOI{\tempurl}


\bibitem[Hundman et~al\mbox{.}(2018b)]%
        {hundman2018detecting}
\bibfield{author}{\bibinfo{person}{Kyle Hundman}, \bibinfo{person}{Valentino Constantinou}, \bibinfo{person}{Christopher Laporte}, \bibinfo{person}{Ian Colwell}, {and} \bibinfo{person}{Tom Soderstrom}.} \bibinfo{year}{2018}\natexlab{b}.
\newblock \showarticletitle{Detecting spacecraft anomalies using lstms and nonparametric dynamic thresholding}. In \bibinfo{booktitle}{\emph{SIGKDD}}. \bibinfo{pages}{387--395}.
\newblock


\bibitem[Hung(2017)]%
        {hung2017leading}
\bibfield{author}{\bibinfo{person}{Mark Hung}.} \bibinfo{year}{2017}\natexlab{}.
\newblock \showarticletitle{Leading the iot, gartner insights on how to lead in a connected world}.
\newblock \bibinfo{journal}{\emph{Gartner Research}} (\bibinfo{year}{2017}), \bibinfo{pages}{1--29}.
\newblock


\bibitem[Irwin(1925)]%
        {irwin1925criterion}
\bibfield{author}{\bibinfo{person}{JO Irwin}.} \bibinfo{year}{1925}\natexlab{}.
\newblock \showarticletitle{On a criterion for the rejection of outlying observations}.
\newblock \bibinfo{journal}{\emph{Biometrika}} (\bibinfo{year}{1925}), \bibinfo{pages}{238--250}.
\newblock


\bibitem[Izenman(2008)]%
        {izenman2008modern}
\bibfield{author}{\bibinfo{person}{Alan~Julian Izenman}.} \bibinfo{year}{2008}\natexlab{}.
\newblock \showarticletitle{Modern multivariate statistical techniques}.
\newblock \bibinfo{journal}{\emph{Regression, classification and manifold learning}}  \bibinfo{volume}{10} (\bibinfo{year}{2008}), \bibinfo{pages}{978--0}.
\newblock


\bibitem[Jacob et~al\mbox{.}(2021)]%
        {jacob2021exathlon}
\bibfield{author}{\bibinfo{person}{Vincent Jacob}, \bibinfo{person}{Fei Song}, \bibinfo{person}{Arnaud Stiegler}, \bibinfo{person}{Bijan Rad}, \bibinfo{person}{Yanlei Diao}, {and} \bibinfo{person}{Nesime Tatbul}.} \bibinfo{year}{2021}\natexlab{}.
\newblock \showarticletitle{Exathlon: a benchmark for explainable anomaly detection over time series}.
\newblock \bibinfo{journal}{\emph{PVLDB}} \bibinfo{volume}{14}, \bibinfo{number}{11} (\bibinfo{year}{2021}), \bibinfo{pages}{2613--2626}.
\newblock


\bibitem[Jeung et~al\mbox{.}(2010)]%
        {jeung2010effective}
\bibfield{author}{\bibinfo{person}{Hoyoung Jeung}, \bibinfo{person}{Sofiane Sarni}, \bibinfo{person}{Ioannis Paparrizos}, \bibinfo{person}{Saket Sathe}, \bibinfo{person}{Karl Aberer}, \bibinfo{person}{Nicholas Dawes}, \bibinfo{person}{Thanasis~G Papaioannou}, {and} \bibinfo{person}{Michael Lehning}.} \bibinfo{year}{2010}\natexlab{}.
\newblock \showarticletitle{Effective metadata management in federated sensor networks}. In \bibinfo{booktitle}{\emph{2010 IEEE International Conference on Sensor Networks, Ubiquitous, and Trustworthy Computing}}. IEEE, \bibinfo{pages}{107--114}.
\newblock


\bibitem[Jiang et~al\mbox{.}(2020)]%
        {jiang2020pids}
\bibfield{author}{\bibinfo{person}{Hao Jiang}, \bibinfo{person}{Chunwei Liu}, \bibinfo{person}{Qi Jin}, \bibinfo{person}{John Paparrizos}, {and} \bibinfo{person}{Aaron~J Elmore}.} \bibinfo{year}{2020}\natexlab{}.
\newblock \showarticletitle{Pids: attribute decomposition for improved compression and query performance in columnar storage}.
\newblock \bibinfo{journal}{\emph{Proc. VLDB Endow}} \bibinfo{volume}{13}, \bibinfo{number}{6} (\bibinfo{year}{2020}), \bibinfo{pages}{925--938}.
\newblock


\bibitem[Jiang et~al\mbox{.}(2021)]%
        {jiang2021good}
\bibfield{author}{\bibinfo{person}{Hao Jiang}, \bibinfo{person}{Chunwei Liu}, \bibinfo{person}{John Paparrizos}, \bibinfo{person}{Andrew~A Chien}, \bibinfo{person}{Jihong Ma}, {and} \bibinfo{person}{Aaron~J Elmore}.} \bibinfo{year}{2021}\natexlab{}.
\newblock \showarticletitle{Good to the Last Bit: Data-Driven Encoding with CodecDB}. In \bibinfo{booktitle}{\emph{SIGMOD}}. \bibinfo{pages}{843--856}.
\newblock


\bibitem[Kashino et~al\mbox{.}(1999)]%
        {kashino1999time}
\bibfield{author}{\bibinfo{person}{Kunio Kashino}, \bibinfo{person}{Gavin Smith}, {and} \bibinfo{person}{Hiroshi Murase}.} \bibinfo{year}{1999}\natexlab{}.
\newblock \showarticletitle{Time-series active search for quick retrieval of audio and video}. In \bibinfo{booktitle}{\emph{1999 IEEE International Conference on Acoustics, Speech, and Signal Processing. Proceedings. ICASSP99 (Cat. No. 99CH36258)}}, Vol.~\bibinfo{volume}{6}. IEEE, \bibinfo{pages}{2993--2996}.
\newblock


\bibitem[Keogh et~al\mbox{.}(2021)]%
        {kdd21}
\bibfield{author}{\bibinfo{person}{E. Keogh}, \bibinfo{person}{T. Dutta~Roy}, \bibinfo{person}{U. Naik}, {and} \bibinfo{person}{A Agrawal}.} \bibinfo{year}{2021}\natexlab{}.
\newblock \bibinfo{title}{{Multi-dataset Time-Series Anomaly Detection Competition 2021}, {\url{https://compete.hexagon-ml.com/practice/competition/39/}}}.
\newblock
\newblock


\bibitem[Keogh et~al\mbox{.}(2005)]%
        {KeoghEtAl2005HOT}
\bibfield{author}{\bibinfo{person}{Eamonn Keogh}, \bibinfo{person}{Jessica Lin}, {and} \bibinfo{person}{Ada Fu}.} \bibinfo{year}{2005}\natexlab{}.
\newblock \showarticletitle{Hot sax: Efficiently finding the most unusual time series subsequence}. In \bibinfo{booktitle}{\emph{Fifth IEEE International Conference on Data Mining (ICDM'05)}}. Ieee, \bibinfo{pages}{8--pp}.
\newblock


\bibitem[Keogh et~al\mbox{.}(2002)]%
        {KeoghEtAl2002Finding}
\bibfield{author}{\bibinfo{person}{Eamonn Keogh}, \bibinfo{person}{Stefano Lonardi}, {and} \bibinfo{person}{Bill'Yuan-chi' Chiu}.} \bibinfo{year}{2002}\natexlab{}.
\newblock \showarticletitle{Finding surprising patterns in a time series database in linear time and space}. In \bibinfo{booktitle}{\emph{Proceedings of the eighth ACM SIGKDD international conference on Knowledge discovery and data mining}}. \bibinfo{pages}{550--556}.
\newblock


\bibitem[Keogh et~al\mbox{.}(2007)]%
        {Keogh2007}
\bibfield{author}{\bibinfo{person}{Eamonn Keogh}, \bibinfo{person}{Stefano Lonardi}, \bibinfo{person}{Chotirat~Ann Ratanamahatana}, \bibinfo{person}{Li Wei}, \bibinfo{person}{Sang-Hee Lee}, {and} \bibinfo{person}{John Handley}.} \bibinfo{year}{2007}\natexlab{}.
\newblock \showarticletitle{Compression-based data mining of sequential data}.
\newblock \bibinfo{journal}{\emph{Data Mining and Knowledge Discovery}} (\bibinfo{year}{2007}).
\newblock


\bibitem[Kim et~al\mbox{.}(2018)]%
        {KimEtAl2018DeepNAP}
\bibfield{author}{\bibinfo{person}{Chunggyeom Kim}, \bibinfo{person}{Jinhyuk Lee}, \bibinfo{person}{Raehyun Kim}, \bibinfo{person}{Youngbin Park}, {and} \bibinfo{person}{Jaewoo Kang}.} \bibinfo{year}{2018}\natexlab{}.
\newblock \showarticletitle{{{DeepNAP}}: {{Deep}} Neural Anomaly Pre-Detection in a Semiconductor Fab}.
\newblock   \bibinfo{volume}{457-458} (\bibinfo{year}{2018}), \bibinfo{pages}{1--11}.
\newblock
\showISSN{00200255}
\urldef\tempurl%
\url{https://doi.org/10.1016/j.ins.2018.05.020}
\showDOI{\tempurl}


\bibitem[Kim et~al\mbox{.}(2022)]%
        {kim2022towards}
\bibfield{author}{\bibinfo{person}{Siwon Kim}, \bibinfo{person}{Kukjin Choi}, \bibinfo{person}{Hyun-Soo Choi}, \bibinfo{person}{Byunghan Lee}, {and} \bibinfo{person}{Sungroh Yoon}.} \bibinfo{year}{2022}\natexlab{}.
\newblock \showarticletitle{Towards a rigorous evaluation of time-series anomaly detection}. In \bibinfo{booktitle}{\emph{Proceedings of the AAAI Conference on Artificial Intelligence}}, Vol.~\bibinfo{volume}{36}. \bibinfo{pages}{7194--7201}.
\newblock


\bibitem[Knieling et~al\mbox{.}(2017)]%
        {knieling2017online}
\bibfield{author}{\bibinfo{person}{S Knieling}, \bibinfo{person}{J Niediek}, \bibinfo{person}{E Kutter}, \bibinfo{person}{J Bostroem}, \bibinfo{person}{CE Elger}, {and} \bibinfo{person}{F Mormann}.} \bibinfo{year}{2017}\natexlab{}.
\newblock \showarticletitle{An online adaptive screening procedure for selective neuronal responses}.
\newblock \bibinfo{journal}{\emph{Journal of neuroscience methods}}  \bibinfo{volume}{291} (\bibinfo{year}{2017}), \bibinfo{pages}{36--42}.
\newblock


\bibitem[Kontaki et~al\mbox{.}(2011)]%
        {KontakiEtAl2011Continuous}
\bibfield{author}{\bibinfo{person}{Maria Kontaki}, \bibinfo{person}{Anastasios Gounaris}, \bibinfo{person}{Apostolos~N Papadopoulos}, \bibinfo{person}{Kostas Tsichlas}, {and} \bibinfo{person}{Yannis Manolopoulos}.} \bibinfo{year}{2011}\natexlab{}.
\newblock \showarticletitle{Continuous monitoring of distance-based outliers over data streams}. In \bibinfo{booktitle}{\emph{2011 IEEE 27th International Conference on Data Engineering}}. IEEE, \bibinfo{pages}{135--146}.
\newblock


\bibitem[Krensky and Hare(2018)]%
        {krensky2018hype}
\bibfield{author}{\bibinfo{person}{Peter Krensky} {and} \bibinfo{person}{Jim Hare}.} \bibinfo{year}{2018}\natexlab{}.
\newblock \showarticletitle{Hype Cycle for Data Science and Machine Learning, 2018}.
\newblock \bibinfo{journal}{\emph{Gartner Company}} (\bibinfo{year}{2018}).
\newblock


\bibitem[Krishnan et~al\mbox{.}(2019)]%
        {krishnan2019artificial}
\bibfield{author}{\bibinfo{person}{Sanjay Krishnan}, \bibinfo{person}{Aaron~J Elmore}, \bibinfo{person}{Michael Franklin}, \bibinfo{person}{John Paparrizos}, \bibinfo{person}{Zechao Shang}, \bibinfo{person}{Adam Dziedzic}, {and} \bibinfo{person}{Rui Liu}.} \bibinfo{year}{2019}\natexlab{}.
\newblock \showarticletitle{Artificial intelligence in resource-constrained and shared environments}.
\newblock \bibinfo{journal}{\emph{ACM SIGOPS Operating Systems Review}} \bibinfo{volume}{53}, \bibinfo{number}{1} (\bibinfo{year}{2019}), \bibinfo{pages}{1--6}.
\newblock


\bibitem[Lai et~al\mbox{.}(2021)]%
        {lai2021revisiting}
\bibfield{author}{\bibinfo{person}{Kwei-Herng Lai}, \bibinfo{person}{Daochen Zha}, \bibinfo{person}{Junjie Xu}, \bibinfo{person}{Yue Zhao}, \bibinfo{person}{Guanchu Wang}, {and} \bibinfo{person}{Xia Hu}.} \bibinfo{year}{2021}\natexlab{}.
\newblock \showarticletitle{Revisiting Time Series Outlier Detection: Definitions and Benchmarks}. In \bibinfo{booktitle}{\emph{NeurIPS}}.
\newblock


\bibitem[Lamrini et~al\mbox{.}(2018)]%
        {LamriniEtAl2018Anomaly}
\bibfield{author}{\bibinfo{person}{Bouchra Lamrini}, \bibinfo{person}{Augustin Gjini}, \bibinfo{person}{Simon Daudin}, \bibinfo{person}{Pascal Pratmarty}, \bibinfo{person}{Fran{\c{c}}ois Armando}, {and} \bibinfo{person}{Louise Trav{\'e}-Massuy{\`e}s}.} \bibinfo{year}{2018}\natexlab{}.
\newblock \showarticletitle{Anomaly Detection using Similarity-based One-Class SVM for Network Traffic Characterization.}. In \bibinfo{booktitle}{\emph{DX}}.
\newblock


\bibitem[Laptev et~al\mbox{.}(2015)]%
        {yahoo}
\bibfield{author}{\bibinfo{person}{N. Laptev}, \bibinfo{person}{S. Amizadeh}, {and} \bibinfo{person}{Y. Billawala}.} \bibinfo{year}{2015}\natexlab{}.
\newblock \bibinfo{booktitle}{\emph{{S5 - A Labeled Anomaly Detection Dataset, version 1.0(16M)}}}.
\newblock
\urldef\tempurl%
\url{https://webscope.sandbox.yahoo.com/catalog.php?datatype=s&did=70}
\showURL{%
\tempurl}


\bibitem[Lavin and Ahmad(2015)]%
        {lavin2015evaluating}
\bibfield{author}{\bibinfo{person}{Alexander Lavin} {and} \bibinfo{person}{Subutai Ahmad}.} \bibinfo{year}{2015}\natexlab{}.
\newblock \showarticletitle{Evaluating real-time anomaly detection algorithms--the Numenta anomaly benchmark}. In \bibinfo{booktitle}{\emph{ICMLA}}. IEEE, \bibinfo{pages}{38--44}.
\newblock


\bibitem[Lee et~al\mbox{.}(2020)]%
        {LeeEtAl2020RePAD}
\bibfield{author}{\bibinfo{person}{Ming-Chang Lee}, \bibinfo{person}{Jia-Chun Lin}, {and} \bibinfo{person}{Ernst~Gunnar Gran}.} \bibinfo{year}{2020}\natexlab{}.
\newblock \showarticletitle{{{RePAD}}: {{Real}}-{{Time Proactive Anomaly Detection}} for {{Time Series}}}. In \bibinfo{booktitle}{\emph{AINA}}, \bibfield{editor}{\bibinfo{person}{Leonard Barolli}, \bibinfo{person}{Flora Amato}, \bibinfo{person}{Francesco Moscato}, \bibinfo{person}{Tomoya Enokido}, {and} \bibinfo{person}{Makoto Takizawa}} (Eds.). \bibinfo{publisher}{{Springer International Publishing}}, \bibinfo{pages}{1291--1302}.
\newblock
\showISBNx{978-3-030-44041-1}
\urldef\tempurl%
\url{https://doi.org/10.1007/978-3-030-44041-1_110}
\showDOI{\tempurl}


\bibitem[Li et~al\mbox{.}(2018b)]%
        {DBLP:journals/corr/abs-1809-04758}
\bibfield{author}{\bibinfo{person}{Dan Li}, \bibinfo{person}{Dacheng Chen}, \bibinfo{person}{Jonathan Goh}, {and} \bibinfo{person}{See{-}Kiong Ng}.} \bibinfo{year}{2018}\natexlab{b}.
\newblock \showarticletitle{Anomaly Detection with Generative Adversarial Networks for Multivariate Time Series}.
\newblock \bibinfo{journal}{\emph{CoRR}}  \bibinfo{volume}{abs/1809.04758} (\bibinfo{year}{2018}).
\newblock
\showeprint[arxiv]{1809.04758}
\urldef\tempurl%
\url{http://arxiv.org/abs/1809.04758}
\showURL{%
\tempurl}


\bibitem[Li et~al\mbox{.}(2019)]%
        {LiEtAl2019MADGAN}
\bibfield{author}{\bibinfo{person}{Dan Li}, \bibinfo{person}{Dacheng Chen}, \bibinfo{person}{Baihong Jin}, \bibinfo{person}{Lei Shi}, \bibinfo{person}{Jonathan Goh}, {and} \bibinfo{person}{See-Kiong Ng}.} \bibinfo{year}{2019}\natexlab{}.
\newblock \showarticletitle{MAD-GAN: Multivariate anomaly detection for time series data with generative adversarial networks}. In \bibinfo{booktitle}{\emph{International conference on artificial neural networks}}. Springer, \bibinfo{pages}{703--716}.
\newblock


\bibitem[Li et~al\mbox{.}(2018a)]%
        {LiEtAl2018Robust}
\bibfield{author}{\bibinfo{person}{Zeyan Li}, \bibinfo{person}{Wenxiao Chen}, {and} \bibinfo{person}{Dan Pei}.} \bibinfo{year}{2018}\natexlab{a}.
\newblock \showarticletitle{Robust and {{Unsupervised KPI Anomaly Detection Based}} on {{Conditional Variational Autoencoder}}}. In \bibinfo{booktitle}{\emph{Proceedings of the {{International Performance Computing}} and {{Communications Conference}} ({{IPCCC}})}}. \bibinfo{publisher}{{IEEE}}, \bibinfo{pages}{1--9}.
\newblock
\showISBNx{978-1-5386-6808-5}
\urldef\tempurl%
\url{https://doi.org/10.1109/PCCC.2018.8710885}
\showDOI{\tempurl}


\bibitem[Li et~al\mbox{.}(2017)]%
        {LiEtAl2017LocalityBased}
\bibfield{author}{\bibinfo{person}{Zhihua Li}, \bibinfo{person}{Ziyuan Li}, \bibinfo{person}{Ning Yu}, \bibinfo{person}{Steven Wen}, {et~al\mbox{.}}} \bibinfo{year}{2017}\natexlab{}.
\newblock \showarticletitle{Locality-based visual outlier detection algorithm for time series}.
\newblock \bibinfo{journal}{\emph{Security and Communication Networks}}  \bibinfo{volume}{2017} (\bibinfo{year}{2017}).
\newblock


\bibitem[Li et~al\mbox{.}(2007)]%
        {10.1007/978-3-540-71701-0_17}
\bibfield{author}{\bibinfo{person}{Zhi Li}, \bibinfo{person}{Hong Ma}, {and} \bibinfo{person}{Yongbing Mei}.} \bibinfo{year}{2007}\natexlab{}.
\newblock \showarticletitle{A Unifying Method for Outlier and Change Detection from Data Streams Based on Local Polynomial Fitting}. In \bibinfo{booktitle}{\emph{Advances in Knowledge Discovery and Data Mining}}, \bibfield{editor}{\bibinfo{person}{Zhi-Hua Zhou}, \bibinfo{person}{Hang Li}, {and} \bibinfo{person}{Qiang Yang}} (Eds.). \bibinfo{publisher}{Springer Berlin Heidelberg}, \bibinfo{address}{Berlin, Heidelberg}, \bibinfo{pages}{150--161}.
\newblock
\showISBNx{978-3-540-71701-0}


\bibitem[Li et~al\mbox{.}(2020)]%
        {li2020copod}
\bibfield{author}{\bibinfo{person}{Zheng Li}, \bibinfo{person}{Yue Zhao}, \bibinfo{person}{Nicola Botta}, \bibinfo{person}{Cezar Ionescu}, {and} \bibinfo{person}{Xiyang Hu}.} \bibinfo{year}{2020}\natexlab{}.
\newblock \showarticletitle{{COPOD:} Copula-Based Outlier Detection}. In \bibinfo{booktitle}{\emph{IEEE International Conference on Data Mining (ICDM)}}. IEEE.
\newblock


\bibitem[Linardi et~al\mbox{.}(2018)]%
        {LinardiEtAl2018Matrix}
\bibfield{author}{\bibinfo{person}{Michele Linardi}, \bibinfo{person}{Yan Zhu}, \bibinfo{person}{Themis Palpanas}, {and} \bibinfo{person}{Eamonn Keogh}.} \bibinfo{year}{2018}\natexlab{}.
\newblock \showarticletitle{Matrix {{Profile X}}: {{VALMOD}} - {{Scalable Discovery}} of {{Variable}}-{{Length Motifs}} in {{Data Series}}}. In \bibinfo{booktitle}{\emph{Proceedings of the {{International Conference}} on {{Management}} of {{Data}} ({{SIGMOD}})}} (2018). \bibinfo{publisher}{{ACM Press}}, \bibinfo{pages}{1053--1066}.
\newblock
\showISBNx{978-1-4503-4703-7}
\urldef\tempurl%
\url{https://doi.org/10.1145/3183713.3183744}
\showDOI{\tempurl}


\bibitem[Linardi et~al\mbox{.}(2020a)]%
        {LinardiEtAl2020Matrix}
\bibfield{author}{\bibinfo{person}{Michele Linardi}, \bibinfo{person}{Yan Zhu}, \bibinfo{person}{Themis Palpanas}, {and} \bibinfo{person}{Eamonn Keogh}.} \bibinfo{year}{2020}\natexlab{a}.
\newblock \showarticletitle{Matrix profile goes MAD: variable-length motif and discord discovery in data series}.
\newblock \bibinfo{journal}{\emph{Data Mining and Knowledge Discovery}}  \bibinfo{volume}{34} (\bibinfo{year}{2020}), \bibinfo{pages}{1022--1071}.
\newblock


\bibitem[Linardi et~al\mbox{.}(2020b)]%
        {valmodjournal}
\bibfield{author}{\bibinfo{person}{Michele Linardi}, \bibinfo{person}{Yan Zhu}, \bibinfo{person}{Themis Palpanas}, {and} \bibinfo{person}{Eamonn~J. Keogh}.} \bibinfo{year}{2020}\natexlab{b}.
\newblock \showarticletitle{{Matrix Profile Goes MAD: Variable-Length Motif And Discord Discovery in Data Series}}. In \bibinfo{booktitle}{\emph{{DAMI}}}.
\newblock


\bibitem[Liu et~al\mbox{.}(2021)]%
        {liu2021decomposed}
\bibfield{author}{\bibinfo{person}{Chunwei Liu}, \bibinfo{person}{Hao Jiang}, \bibinfo{person}{John Paparrizos}, {and} \bibinfo{person}{Aaron~J Elmore}.} \bibinfo{year}{2021}\natexlab{}.
\newblock \showarticletitle{Decomposed bounded floats for fast compression and queries}.
\newblock \bibinfo{journal}{\emph{Proc. VLDB Endow}} \bibinfo{volume}{14}, \bibinfo{number}{11} (\bibinfo{year}{2021}), \bibinfo{pages}{2586--2598}.
\newblock


\bibitem[Liu et~al\mbox{.}(2024b)]%
        {liu2024adaedge}
\bibfield{author}{\bibinfo{person}{Chunwei Liu}, \bibinfo{person}{John Paparrizos}, {and} \bibinfo{person}{Aaron~J Elmore}.} \bibinfo{year}{2024}\natexlab{b}.
\newblock \showarticletitle{Adaedge: A dynamic compression selection framework for resource constrained devices}. In \bibinfo{booktitle}{\emph{2024 IEEE 40th International Conference on Data Engineering (ICDE)}}. IEEE, \bibinfo{pages}{1506--1519}.
\newblock


\bibitem[Liu et~al\mbox{.}(2008a)]%
        {LiuEtAl2008Isolation}
\bibfield{author}{\bibinfo{person}{Fei~Tony Liu}, \bibinfo{person}{Kai~Ming Ting}, {and} \bibinfo{person}{Zhi-Hua Zhou}.} \bibinfo{year}{2008}\natexlab{a}.
\newblock \showarticletitle{Isolation {{Forest}}}. In \bibinfo{booktitle}{\emph{Proceedings of the {{International Conference}} on {{Data Mining}} ({{ICDM}})}}. \bibinfo{publisher}{{IEEE}}, \bibinfo{pages}{413--422}.
\newblock
\showISBNx{978-0-7695-3502-9}
\urldef\tempurl%
\url{https://doi.org/10.1109/ICDM.2008.17}
\showDOI{\tempurl}


\bibitem[Liu et~al\mbox{.}(2008b)]%
        {Liu:2008:IF:1510528.1511387}
\bibfield{author}{\bibinfo{person}{Fei~Tony Liu}, \bibinfo{person}{Kai~Ming Ting}, {and} \bibinfo{person}{Zhi-Hua Zhou}.} \bibinfo{year}{2008}\natexlab{b}.
\newblock \showarticletitle{Isolation forest}. In \bibinfo{booktitle}{\emph{ICDM}}. IEEE, \bibinfo{pages}{413--422}.
\newblock


\bibitem[Liu et~al\mbox{.}(2024a)]%
        {liu2024time}
\bibfield{author}{\bibinfo{person}{Qinghua Liu}, \bibinfo{person}{Paul Boniol}, \bibinfo{person}{Themis Palpanas}, {and} \bibinfo{person}{John Paparrizos}.} \bibinfo{year}{2024}\natexlab{a}.
\newblock \showarticletitle{Time-Series Anomaly Detection: Overview and New Trends}.
\newblock \bibinfo{journal}{\emph{PVLDB}} \bibinfo{volume}{17}, \bibinfo{number}{12} (\bibinfo{year}{2024}), \bibinfo{pages}{4229--4232}.
\newblock


\bibitem[Liu and Paparrizos(2024)]%
        {liu2024elephant}
\bibfield{author}{\bibinfo{person}{Qinghua Liu} {and} \bibinfo{person}{John Paparrizos}.} \bibinfo{year}{2024}\natexlab{}.
\newblock \showarticletitle{The Elephant in the Room: Towards A Reliable Time-Series Anomaly Detection Benchmark}. In \bibinfo{booktitle}{\emph{NeurIPS 2024}}.
\newblock


\bibitem[Liu et~al\mbox{.}(2023)]%
        {liu2023amir}
\bibfield{author}{\bibinfo{person}{Shinan Liu}, \bibinfo{person}{Tarun Mangla}, \bibinfo{person}{Ted Shaowang}, \bibinfo{person}{Jinjin Zhao}, \bibinfo{person}{John Paparrizos}, \bibinfo{person}{Sanjay Krishnan}, {and} \bibinfo{person}{Nick Feamster}.} \bibinfo{year}{2023}\natexlab{}.
\newblock \showarticletitle{Amir: Active multimodal interaction recognition from video and network traffic in connected environments}.
\newblock \bibinfo{journal}{\emph{Proceedings of the ACM on Interactive, Mobile, Wearable and Ubiquitous Technologies}} \bibinfo{volume}{7}, \bibinfo{number}{1} (\bibinfo{year}{2023}), \bibinfo{pages}{1--26}.
\newblock


\bibitem[Liu et~al\mbox{.}(2009)]%
        {Liu2009}
\bibfield{author}{\bibinfo{person}{Yubao Liu}, \bibinfo{person}{Xiuwei Chen}, {and} \bibinfo{person}{Fei Wang}.} \bibinfo{year}{2009}\natexlab{}.
\newblock \showarticletitle{{Efficient Detection of Discords for Time Series Stream}}.
\newblock \bibinfo{journal}{\emph{Advances in Data and Web Management}} (\bibinfo{year}{2009}), \bibinfo{pages}{629--634}.
\newblock
\urldef\tempurl%
\url{http://www.springerlink.com/index/n546h380446p95r7.pdf}
\showURL{%
\tempurl}


\bibitem[Lobo et~al\mbox{.}(2008)]%
        {lobo2008auc}
\bibfield{author}{\bibinfo{person}{Jorge~M Lobo}, \bibinfo{person}{Alberto Jim{\'e}nez-Valverde}, {and} \bibinfo{person}{Raimundo Real}.} \bibinfo{year}{2008}\natexlab{}.
\newblock \showarticletitle{AUC: a misleading measure of the performance of predictive distribution models}.
\newblock \bibinfo{journal}{\emph{Global ecology and Biogeography}} \bibinfo{volume}{17}, \bibinfo{number}{2} (\bibinfo{year}{2008}), \bibinfo{pages}{145--151}.
\newblock


\bibitem[Lu et~al\mbox{.}(2022)]%
        {10.1145/3534678.3539271}
\bibfield{author}{\bibinfo{person}{Yue Lu}, \bibinfo{person}{Renjie Wu}, \bibinfo{person}{Abdullah Mueen}, \bibinfo{person}{Maria~A Zuluaga}, {and} \bibinfo{person}{Eamonn Keogh}.} \bibinfo{year}{2022}\natexlab{}.
\newblock \showarticletitle{Matrix profile XXIV: scaling time series anomaly detection to trillions of datapoints and ultra-fast arriving data streams}. In \bibinfo{booktitle}{\emph{SIGKDD}}. \bibinfo{pages}{1173--1182}.
\newblock


\bibitem[Luo and Gallagher(2011)]%
        {Parameter-Free_Discord}
\bibfield{author}{\bibinfo{person}{Wei Luo} {and} \bibinfo{person}{Marcus Gallagher}.} \bibinfo{year}{2011}\natexlab{}.
\newblock \showarticletitle{Faster and Parameter-Free Discord Search in Quasi-Periodic Time Series}. In \bibinfo{booktitle}{\emph{Advances in Knowledge Discovery and Data Mining}}, \bibfield{editor}{\bibinfo{person}{Joshua~Zhexue Huang}, \bibinfo{person}{Longbing Cao}, {and} \bibinfo{person}{Jaideep Srivastava}} (Eds.).
\newblock


\bibitem[L{\"u}tkepohl et~al\mbox{.}(2004)]%
        {lutkepohl2004applied}
\bibfield{author}{\bibinfo{person}{Helmut L{\"u}tkepohl}, \bibinfo{person}{Markus Kr{\"a}tzig}, {and} \bibinfo{person}{Peter~CB Phillips}.} \bibinfo{year}{2004}\natexlab{}.
\newblock \bibinfo{booktitle}{\emph{Applied time series econometrics}}.
\newblock \bibinfo{publisher}{Cambridge university press}.
\newblock


\bibitem[Ma and Perkins(2003a)]%
        {MaPerkins2003Online}
\bibfield{author}{\bibinfo{person}{Junshui Ma} {and} \bibinfo{person}{Simon Perkins}.} \bibinfo{year}{2003}\natexlab{a}.
\newblock \showarticletitle{Online Novelty Detection on Temporal Sequences}. In \bibinfo{booktitle}{\emph{Proceedings of the {{International Conference}} on {{Knowledge Discovery}} and {{Data Mining}} ({{SIGKDD}})}}. \bibinfo{publisher}{{ACM Press}}, \bibinfo{pages}{613}.
\newblock
\showISBNx{978-1-58113-737-8}
\urldef\tempurl%
\url{https://doi.org/10.1145/956750.956828}
\showDOI{\tempurl}


\bibitem[Ma and Perkins(2003b)]%
        {MaPerkins2003Timeseries}
\bibfield{author}{\bibinfo{person}{Junshui Ma} {and} \bibinfo{person}{Simon Perkins}.} \bibinfo{year}{2003}\natexlab{b}.
\newblock \showarticletitle{Time-series novelty detection using one-class support vector machines}. In \bibinfo{booktitle}{\emph{Proceedings of the International Joint Conference on Neural Networks, 2003.}}, Vol.~\bibinfo{volume}{3}. IEEE, \bibinfo{pages}{1741--1745}.
\newblock


\bibitem[Mahdavinejad et~al\mbox{.}(2017)]%
        {mahdavinejad2017machine}
\bibfield{author}{\bibinfo{person}{Mohammad~Saeid Mahdavinejad}, \bibinfo{person}{Mohammadreza Rezvan}, \bibinfo{person}{Mohammadamin Barekatain}, \bibinfo{person}{Peyman Adibi}, \bibinfo{person}{Payam Barnaghi}, {and} \bibinfo{person}{Amit~P Sheth}.} \bibinfo{year}{2017}\natexlab{}.
\newblock \showarticletitle{Machine learning for Internet of Things data analysis: A survey}.
\newblock \bibinfo{journal}{\emph{Digital Communications and Networks}} (\bibinfo{year}{2017}).
\newblock


\bibitem[Malhotra et~al\mbox{.}(2016)]%
        {MalhotraEtAl2016LSTMbased}
\bibfield{author}{\bibinfo{person}{Pankaj Malhotra}, \bibinfo{person}{Anusha Ramakrishnan}, \bibinfo{person}{Gaurangi Anand}, \bibinfo{person}{Lovekesh Vig}, \bibinfo{person}{Puneet Agarwal}, {and} \bibinfo{person}{Gautam Shroff}.} \bibinfo{year}{2016}\natexlab{}.
\newblock \bibinfo{booktitle}{\emph{{{LSTM}}-Based {{Encoder}}-{{Decoder}} for {{Multi}}-Sensor {{Anomaly Detection}}}}.
\newblock
\showeprint[arxiv]{1607.00148}~[cs, stat]
\urldef\tempurl%
\url{http://arxiv.org/abs/1607.00148}
\showURL{%
\tempurl}


\bibitem[Malhotra et~al\mbox{.}(2015b)]%
        {MalhotraEtAl2015Long}
\bibfield{author}{\bibinfo{person}{Pankaj Malhotra}, \bibinfo{person}{Lovekesh Vig}, \bibinfo{person}{Gautam Shroff}, {and} \bibinfo{person}{Puneet Agarwal}.} \bibinfo{year}{2015}\natexlab{b}.
\newblock \showarticletitle{Long {{Short Term Memory Networks}} for {{Anomaly Detection}} in {{Time Series}}}. In \bibinfo{booktitle}{\emph{Proceedings of the {{European Symposium}} on {{Artificial Neural Networks}}, {{Computational Intelligence}} and {{Machine Learning}} ({{ESANN}})}}, Vol.~\bibinfo{volume}{23}.
\newblock
\urldef\tempurl%
\url{http://www.elen.ucl.ac.be/Proceedings/esann/esannpdf/es2015-56.pdf}
\showURL{%
\tempurl}


\bibitem[Malhotra et~al\mbox{.}(2015a)]%
        {LSTManomaly}
\bibfield{author}{\bibinfo{person}{Pankaj Malhotra}, \bibinfo{person}{Lovekesh Vig}, \bibinfo{person}{Gautam Shroff}, \bibinfo{person}{Puneet Agarwal}, {et~al\mbox{.}}} \bibinfo{year}{2015}\natexlab{a}.
\newblock \showarticletitle{Long Short Term Memory Networks for Anomaly Detection in Time Series.}. In \bibinfo{booktitle}{\emph{Esann}}, Vol.~\bibinfo{volume}{2015}. \bibinfo{pages}{89}.
\newblock


\bibitem[Mantegna(1999)]%
        {mantegna1999hierarchical}
\bibfield{author}{\bibinfo{person}{Rosario~N Mantegna}.} \bibinfo{year}{1999}\natexlab{}.
\newblock \showarticletitle{Hierarchical structure in financial markets}.
\newblock \bibinfo{journal}{\emph{The European Physical Journal B-Condensed Matter and Complex Systems}} \bibinfo{volume}{11}, \bibinfo{number}{1} (\bibinfo{year}{1999}), \bibinfo{pages}{193--197}.
\newblock


\bibitem[Marceau(2000)]%
        {Marceau2000Characterizing}
\bibfield{author}{\bibinfo{person}{Carla Marceau}.} \bibinfo{year}{2000}\natexlab{}.
\newblock \showarticletitle{Characterizing the Behavior of a Program Using Multiple-Length {{N}}-Grams}. In \bibinfo{booktitle}{\emph{Proceedings of the {{Workshop}} on {{New Security Paradigms}} ({{NSPW}})}}. \bibinfo{publisher}{{ACM Press}}, \bibinfo{pages}{101--110}.
\newblock
\showISBNx{978-1-58113-260-1}
\urldef\tempurl%
\url{https://doi.org/10.1145/366173.366197}
\showDOI{\tempurl}


\bibitem[Marteau et~al\mbox{.}(2017)]%
        {MarteauEtAl2017Hybrid}
\bibfield{author}{\bibinfo{person}{Pierre-François Marteau}, \bibinfo{person}{Saeid Soheily-Khah}, {and} \bibinfo{person}{Nicolas Béchet}.} \bibinfo{year}{2017}\natexlab{}.
\newblock \bibinfo{booktitle}{\emph{Hybrid {{Isolation Forest}} - {{Application}} to {{Intrusion Detection}}}}.
\newblock
\showeprint[arxiv]{1705.03800}~[cs]
\urldef\tempurl%
\url{http://arxiv.org/abs/1705.03800}
\showURL{%
\tempurl}


\bibitem[Mart{\'\i}nez-{\'A}lvarez et~al\mbox{.}(2015)]%
        {martinez2015survey}
\bibfield{author}{\bibinfo{person}{Francisco Mart{\'\i}nez-{\'A}lvarez}, \bibinfo{person}{Alicia Troncoso}, \bibinfo{person}{Gualberto Asencio-Cort{\'e}s}, {and} \bibinfo{person}{Jos{\'e} Riquelme}.} \bibinfo{year}{2015}\natexlab{}.
\newblock \showarticletitle{A survey on data mining techniques applied to electricity-related time series forecasting}.
\newblock \bibinfo{journal}{\emph{Energies}} \bibinfo{volume}{8}, \bibinfo{number}{11} (\bibinfo{year}{2015}), \bibinfo{pages}{13162--13193}.
\newblock


\bibitem[Mason and Young(2002)]%
        {mason2002multivariate}
\bibfield{author}{\bibinfo{person}{Robert~L Mason} {and} \bibinfo{person}{John~C Young}.} \bibinfo{year}{2002}\natexlab{}.
\newblock \bibinfo{booktitle}{\emph{Multivariate statistical process control with industrial applications}}.
\newblock \bibinfo{publisher}{SIAM}.
\newblock


\bibitem[McCleary et~al\mbox{.}(1980)]%
        {mccleary1980applied}
\bibfield{author}{\bibinfo{person}{Richard McCleary}, \bibinfo{person}{Richard~A Hay}, \bibinfo{person}{Erroll~E Meidinger}, {and} \bibinfo{person}{David McDowall}.} \bibinfo{year}{1980}\natexlab{}.
\newblock \bibinfo{booktitle}{\emph{Applied time series analysis for the social sciences}}.
\newblock \bibinfo{publisher}{Sage Publications Beverly Hills, CA}.
\newblock


\bibitem[McKeown et~al\mbox{.}(2016)]%
        {mckeown2016predicting}
\bibfield{author}{\bibinfo{person}{Kathy McKeown}, \bibinfo{person}{Hal Daume~III}, \bibinfo{person}{Snigdha Chaturvedi}, \bibinfo{person}{John Paparrizos}, \bibinfo{person}{Kapil Thadani}, \bibinfo{person}{Pablo Barrio}, \bibinfo{person}{Or Biran}, \bibinfo{person}{Suvarna Bothe}, \bibinfo{person}{Michael Collins}, \bibinfo{person}{Kenneth~R Fleischmann}, {et~al\mbox{.}}} \bibinfo{year}{2016}\natexlab{}.
\newblock \showarticletitle{Predicting the impact of scientific concepts using full-text features}.
\newblock \bibinfo{journal}{\emph{Journal of the Association for Information Science and Technology}} \bibinfo{volume}{67}, \bibinfo{number}{11} (\bibinfo{year}{2016}), \bibinfo{pages}{2684--2696}.
\newblock


\bibitem[Mirylenka et~al\mbox{.}(2016)]%
        {mirylenka2016characterizing}
\bibfield{author}{\bibinfo{person}{Katsiaryna Mirylenka}, \bibinfo{person}{Vassilis Christophides}, \bibinfo{person}{Themis Palpanas}, \bibinfo{person}{Ioannis Pefkianakis}, {and} \bibinfo{person}{Martin May}.} \bibinfo{year}{2016}\natexlab{}.
\newblock \showarticletitle{Characterizing home device usage from wireless traffic time series}.
\newblock


\bibitem[{Moradi Fard} et~al\mbox{.}(2020)]%
        {MORADIFARD2020185}
\bibfield{author}{\bibinfo{person}{Maziar {Moradi Fard}}, \bibinfo{person}{Thibaut Thonet}, {and} \bibinfo{person}{Eric Gaussier}.} \bibinfo{year}{2020}\natexlab{}.
\newblock \showarticletitle{Deep k-Means: Jointly clustering with k-Means and learning representations}.
\newblock \bibinfo{journal}{\emph{Pattern Recognition Letters}}  \bibinfo{volume}{138} (\bibinfo{year}{2020}), \bibinfo{pages}{185 -- 192}.
\newblock
\showISSN{0167-8655}
\urldef\tempurl%
\url{https://doi.org/10.1016/j.patrec.2020.07.028}
\showDOI{\tempurl}


\bibitem[Morales-Esteban et~al\mbox{.}(2010)]%
        {morales2010pattern}
\bibfield{author}{\bibinfo{person}{A Morales-Esteban}, \bibinfo{person}{Francisco Mart{\'\i}nez-{\'A}lvarez}, \bibinfo{person}{A Troncoso}, \bibinfo{person}{JL Justo}, {and} \bibinfo{person}{Cristina Rubio-Escudero}.} \bibinfo{year}{2010}\natexlab{}.
\newblock \showarticletitle{Pattern recognition to forecast seismic time series}.
\newblock \bibinfo{journal}{\emph{Expert Systems with Applications}} \bibinfo{volume}{37}, \bibinfo{number}{12} (\bibinfo{year}{2010}), \bibinfo{pages}{8333--8342}.
\newblock


\bibitem[Morariu and Borangiu(2018)]%
        {MorariuBorangiu2018Time}
\bibfield{author}{\bibinfo{person}{Cristina Morariu} {and} \bibinfo{person}{Theodor Borangiu}.} \bibinfo{year}{2018}\natexlab{}.
\newblock \showarticletitle{Time series forecasting for dynamic scheduling of manufacturing processes}. In \bibinfo{booktitle}{\emph{2018 IEEE International Conference on Automation, Quality and Testing, Robotics (AQTR)}}. \bibinfo{pages}{1--6}.
\newblock
\urldef\tempurl%
\url{https://doi.org/10.1109/AQTR.2018.8402748}
\showDOI{\tempurl}


\bibitem[Mueen et~al\mbox{.}(2017)]%
        {FastestSimilaritySearch}
\bibfield{author}{\bibinfo{person}{Abdullah Mueen}, \bibinfo{person}{Yan Zhu}, \bibinfo{person}{Michael Yeh}, \bibinfo{person}{Kaveh Kamgar}, \bibinfo{person}{Krishnamurthy Viswanathan}, \bibinfo{person}{Chetan Gupta}, {and} \bibinfo{person}{Eamonn Keogh}.} \bibinfo{year}{2017}\natexlab{}.
\newblock \bibinfo{title}{The Fastest Similarity Search Algorithm for Time Series Subsequences under Euclidean Distance}.
\newblock
\newblock


\bibitem[Munir et~al\mbox{.}(2019)]%
        {MunirEtAl2019DeepAnT}
\bibfield{author}{\bibinfo{person}{Mohsin Munir}, \bibinfo{person}{Shoaib~Ahmed Siddiqui}, \bibinfo{person}{Andreas Dengel}, {and} \bibinfo{person}{Sheraz Ahmed}.} \bibinfo{year}{2019}\natexlab{}.
\newblock \showarticletitle{{{DeepAnT}}: {{A Deep Learning Approach}} for {{Unsupervised Anomaly Detection}} in {{Time Series}}}.
\newblock   \bibinfo{volume}{7} (\bibinfo{year}{2019}), \bibinfo{pages}{1991--2005}.
\newblock
\showISSN{2169-3536}
\urldef\tempurl%
\url{https://doi.org/10.1109/ACCESS.2018.2886457}
\showDOI{\tempurl}


\bibitem[Na et~al\mbox{.}(2018)]%
        {NaEtAl2018DILOF}
\bibfield{author}{\bibinfo{person}{Gyoung~S Na}, \bibinfo{person}{Donghyun Kim}, {and} \bibinfo{person}{Hwanjo Yu}.} \bibinfo{year}{2018}\natexlab{}.
\newblock \showarticletitle{Dilof: Effective and memory efficient local outlier detection in data streams}. In \bibinfo{booktitle}{\emph{SIGKDD}}. \bibinfo{pages}{1993--2002}.
\newblock


\bibitem[Nakamura et~al\mbox{.}(2020)]%
        {DBLP:conf/icdm/NakamuraIMK20}
\bibfield{author}{\bibinfo{person}{Takaaki Nakamura}, \bibinfo{person}{Makoto Imamura}, \bibinfo{person}{Ryan Mercer}, {and} \bibinfo{person}{Eamonn~J. Keogh}.} \bibinfo{year}{2020}\natexlab{}.
\newblock \showarticletitle{{MERLIN:} Parameter-Free Discovery of Arbitrary Length Anomalies in Massive Time Series Archives}. In \bibinfo{booktitle}{\emph{ICDM}}, \bibfield{editor}{\bibinfo{person}{Claudia Plant}, \bibinfo{person}{Haixun Wang}, \bibinfo{person}{Alfredo Cuzzocrea}, \bibinfo{person}{Carlo Zaniolo}, {and} \bibinfo{person}{Xindong Wu}} (Eds.). \bibinfo{publisher}{{IEEE}}, \bibinfo{pages}{1190--1195}.
\newblock
\urldef\tempurl%
\url{https://doi.org/10.1109/ICDM50108.2020.00147}
\showDOI{\tempurl}


\bibitem[Nakamura et~al\mbox{.}(2023)]%
        {MERLINplusplus}
\bibfield{author}{\bibinfo{person}{Takaaki Nakamura}, \bibinfo{person}{Ryan Mercer}, \bibinfo{person}{Makoto Imamura}, {and} \bibinfo{person}{Eamonn Keogh}.} \bibinfo{year}{2023}\natexlab{}.
\newblock \showarticletitle{MERLIN++: parameter-free discovery of time series anomalies}.
\newblock \bibinfo{journal}{\emph{Data Mining and Knowledge Discovery}} \bibinfo{volume}{37}, \bibinfo{number}{2} (\bibinfo{year}{2023}), \bibinfo{pages}{670--709}.
\newblock
\showISBNx{1573-756X}
\urldef\tempurl%
\url{https://doi.org/10.1007/s10618-022-00876-7}
\showDOI{\tempurl}


\bibitem[Niu et~al\mbox{.}(2020)]%
        {NiuEtAl2020LSTMBased}
\bibfield{author}{\bibinfo{person}{Zijian Niu}, \bibinfo{person}{Ke Yu}, {and} \bibinfo{person}{Xiaofei Wu}.} \bibinfo{year}{2020}\natexlab{}.
\newblock \showarticletitle{LSTM-based VAE-GAN for time-series anomaly detection}.
\newblock \bibinfo{journal}{\emph{Sensors}} \bibinfo{volume}{20}, \bibinfo{number}{13} (\bibinfo{year}{2020}), \bibinfo{pages}{3738}.
\newblock


\bibitem[Obst et~al\mbox{.}(2008)]%
        {ObstEtAl2008Using}
\bibfield{author}{\bibinfo{person}{Oliver Obst}, \bibinfo{person}{X.~Rosalind Wang}, {and} \bibinfo{person}{Mikhail Prokopenko}.} \bibinfo{year}{2008}\natexlab{}.
\newblock \showarticletitle{Using {{Echo State Networks}} for {{Anomaly Detection}} in {{Underground Coal Mines}}}. In \bibinfo{booktitle}{\emph{Proceedings of the {{International Conference}} on {{Information Processing}} in {{Sensor Networks}} ({{IPSN}})}}. \bibinfo{publisher}{{IEEE}}, \bibinfo{pages}{219--229}.
\newblock
\showISBNx{978-0-7695-3157-1}
\urldef\tempurl%
\url{https://doi.org/10.1109/IPSN.2008.35}
\showDOI{\tempurl}


\bibitem[Ogbechie et~al\mbox{.}(2017)]%
        {OgbechieEtAl2017Dynamic}
\bibfield{author}{\bibinfo{person}{Alberto Ogbechie}, \bibinfo{person}{Javier Díaz-Rozo}, \bibinfo{person}{Pedro Larrañaga}, {and} \bibinfo{person}{Concha Bielza}.} \bibinfo{year}{2017}\natexlab{}.
\newblock \showarticletitle{Dynamic {{Bayesian Network}}-{{Based Anomaly Detection}} for {{In}}-{{Process Visual Inspection}} of {{Laser Surface Heat Treatment}}}. In \bibinfo{booktitle}{\emph{Proceedings of the {{International Conference}} on {{Machine Learning}} for {{Cyber Physical Systems}} ({{ML4CPS}})}}, \bibfield{editor}{\bibinfo{person}{Jürgen Beyerer}, \bibinfo{person}{Oliver Niggemann}, {and} \bibinfo{person}{Christian Kühnert}} (Eds.). \bibinfo{publisher}{{Springer Berlin Heidelberg}}, \bibinfo{pages}{17--24}.
\newblock
\showISBNx{978-3-662-53805-0 978-3-662-53806-7}
\urldef\tempurl%
\url{https://doi.org/10.1007/978-3-662-53806-7_3}
\showDOI{\tempurl}


\bibitem[Paffenroth et~al\mbox{.}(2018)]%
        {PaffenrothEtAl2018Robust}
\bibfield{author}{\bibinfo{person}{Randy Paffenroth}, \bibinfo{person}{Kathleen Kay}, {and} \bibinfo{person}{Les Servi}.} \bibinfo{year}{2018}\natexlab{}.
\newblock \bibinfo{booktitle}{\emph{Robust {{PCA}} for {{Anomaly Detection}} in {{Cyber Networks}}}}.
\newblock
\showeprint[arxiv]{1801.01571}~[cs]
\urldef\tempurl%
\url{http://arxiv.org/abs/1801.01571}
\showURL{%
\tempurl}


\bibitem[Page(1957)]%
        {page1957problems}
\bibfield{author}{\bibinfo{person}{ES Page}.} \bibinfo{year}{1957}\natexlab{}.
\newblock \showarticletitle{On problems in which a change in a parameter occurs at an unknown point}.
\newblock \bibinfo{journal}{\emph{Biometrika}} \bibinfo{volume}{44}, \bibinfo{number}{1/2} (\bibinfo{year}{1957}), \bibinfo{pages}{248--252}.
\newblock


\bibitem[Palpanas(2015)]%
        {Palpanas2015}
\bibfield{author}{\bibinfo{person}{Themis Palpanas}.} \bibinfo{year}{2015}\natexlab{}.
\newblock \showarticletitle{Data Series Management: The Road to Big Sequence Analytics}.
\newblock \bibinfo{journal}{\emph{SIGMOD Rec.}} \bibinfo{volume}{44}, \bibinfo{number}{2} (\bibinfo{year}{2015}), \bibinfo{pages}{47--52}.
\newblock


\bibitem[Palshikar(2005)]%
        {Palshikar2005DistanceBased}
\bibfield{author}{\bibinfo{person}{Girish~Keshav Palshikar}.} \bibinfo{year}{2005}\natexlab{}.
\newblock \showarticletitle{Distance-based outliers in sequences}. In \bibinfo{booktitle}{\emph{Distributed Computing and Internet Technology: Second International Conference, ICDCIT 2005, Bhubaneswar, India, December 22-24, 2005. Proceedings 2}}. Springer, \bibinfo{pages}{547--552}.
\newblock


\bibitem[Papadimitriou et~al\mbox{.}(2003)]%
        {PapadimitriouEtAl2003LOCI}
\bibfield{author}{\bibinfo{person}{Spiros Papadimitriou}, \bibinfo{person}{Hiroyuki Kitagawa}, \bibinfo{person}{Phillip~B Gibbons}, {and} \bibinfo{person}{Christos Faloutsos}.} \bibinfo{year}{2003}\natexlab{}.
\newblock \showarticletitle{Loci: Fast outlier detection using the local correlation integral}. In \bibinfo{booktitle}{\emph{ICDE}}. IEEE, \bibinfo{pages}{315--326}.
\newblock


\bibitem[Paparrizos(2018)]%
        {paparrizos2018fastthesis}
\bibfield{author}{\bibinfo{person}{Ioannis Paparrizos}.} \bibinfo{year}{2018}\natexlab{}.
\newblock \emph{\bibinfo{title}{Fast, Scalable, and Accurate Algorithms for Time-Series Analysis}}.
\newblock \bibinfo{thesistype}{Ph.\,D. Dissertation}. \bibinfo{school}{Columbia University}.
\newblock


\bibitem[Paparrizos et~al\mbox{.}(2022a)]%
        {paparrizos2022volume}
\bibfield{author}{\bibinfo{person}{John Paparrizos}, \bibinfo{person}{Paul Boniol}, \bibinfo{person}{Themis Palpanas}, \bibinfo{person}{Ruey~S Tsay}, \bibinfo{person}{Aaron Elmore}, {and} \bibinfo{person}{Michael~J Franklin}.} \bibinfo{year}{2022}\natexlab{a}.
\newblock \showarticletitle{Volume under the surface: a new accuracy evaluation measure for time-series anomaly detection}.
\newblock \bibinfo{journal}{\emph{PVLDB}} \bibinfo{volume}{15}, \bibinfo{number}{11} (\bibinfo{year}{2022}), \bibinfo{pages}{2774--2787}.
\newblock


\bibitem[Paparrizos et~al\mbox{.}(2022b)]%
        {paparrizos2022fast}
\bibfield{author}{\bibinfo{person}{John Paparrizos}, \bibinfo{person}{Ikraduya Edian}, \bibinfo{person}{Chunwei Liu}, \bibinfo{person}{Aaron~J Elmore}, {and} \bibinfo{person}{Michael~J Franklin}.} \bibinfo{year}{2022}\natexlab{b}.
\newblock \showarticletitle{Fast adaptive similarity search through variance-aware quantization}. In \bibinfo{booktitle}{\emph{ICDE}}. IEEE, \bibinfo{pages}{2969--2983}.
\newblock


\bibitem[Paparrizos and Franklin(2019)]%
        {paparrizos2019grail}
\bibfield{author}{\bibinfo{person}{John Paparrizos} {and} \bibinfo{person}{Michael~J Franklin}.} \bibinfo{year}{2019}\natexlab{}.
\newblock \showarticletitle{GRAIL: efficient time-series representation learning}.
\newblock \bibinfo{journal}{\emph{Proceedings of the VLDB Endowment}} \bibinfo{volume}{12}, \bibinfo{number}{11} (\bibinfo{year}{2019}), \bibinfo{pages}{1762--1777}.
\newblock


\bibitem[Paparrizos and Gravano(2016)]%
        {paparrizos_k-shape_2016}
\bibfield{author}{\bibinfo{person}{John Paparrizos} {and} \bibinfo{person}{Luis Gravano}.} \bibinfo{year}{2016}\natexlab{}.
\newblock \showarticletitle{k-{Shape}: {Efficient} and {Accurate} {Clustering} of {Time} {Series}}.
\newblock \bibinfo{journal}{\emph{SIGMOD}} \bibinfo{volume}{45}, \bibinfo{number}{1} (\bibinfo{date}{June} \bibinfo{year}{2016}), \bibinfo{pages}{69--76}.
\newblock
\showISSN{0163-5808}
\urldef\tempurl%
\url{https://doi.org/10.1145/2949741.2949758}
\showDOI{\tempurl}


\bibitem[Paparrizos and Gravano(2017)]%
        {paparrizos2017fast}
\bibfield{author}{\bibinfo{person}{John Paparrizos} {and} \bibinfo{person}{Luis Gravano}.} \bibinfo{year}{2017}\natexlab{}.
\newblock \showarticletitle{Fast and Accurate Time-Series Clustering}.
\newblock \bibinfo{journal}{\emph{ACM Transactions on Database Systems (TODS)}} \bibinfo{volume}{42}, \bibinfo{number}{2} (\bibinfo{year}{2017}), \bibinfo{pages}{8}.
\newblock


\bibitem[Paparrizos et~al\mbox{.}(2022c)]%
        {paparrizos2022tsb}
\bibfield{author}{\bibinfo{person}{John Paparrizos}, \bibinfo{person}{Yuhao Kang}, \bibinfo{person}{Paul Boniol}, \bibinfo{person}{Ruey~S Tsay}, \bibinfo{person}{Themis Palpanas}, {and} \bibinfo{person}{Michael~J Franklin}.} \bibinfo{year}{2022}\natexlab{c}.
\newblock \showarticletitle{TSB-UAD: an end-to-end benchmark suite for univariate time-series anomaly detection}.
\newblock \bibinfo{journal}{\emph{PVLDB}} \bibinfo{volume}{15}, \bibinfo{number}{8} (\bibinfo{year}{2022}), \bibinfo{pages}{1697--1711}.
\newblock


\bibitem[Paparrizos et~al\mbox{.}(2021)]%
        {paparrizos2021vergedb}
\bibfield{author}{\bibinfo{person}{John Paparrizos}, \bibinfo{person}{Chunwei Liu}, \bibinfo{person}{Bruno Barbarioli}, \bibinfo{person}{Johnny Hwang}, \bibinfo{person}{Ikraduya Edian}, \bibinfo{person}{Aaron~J Elmore}, \bibinfo{person}{Michael~J Franklin}, {and} \bibinfo{person}{Sanjay Krishnan}.} \bibinfo{year}{2021}\natexlab{}.
\newblock \showarticletitle{VergeDB: A Database for IoT Analytics on Edge Devices.}. In \bibinfo{booktitle}{\emph{CIDR}}.
\newblock


\bibitem[Paparrizos et~al\mbox{.}(2023a)]%
        {paparrizosDEB23}
\bibfield{author}{\bibinfo{person}{John Paparrizos}, \bibinfo{person}{Chunwei Liu}, \bibinfo{person}{Aaron Elmore}, {and} \bibinfo{person}{Michael~J. Franklin}.} \bibinfo{year}{2023}\natexlab{a}.
\newblock \showarticletitle{Querying Time-Series Data: A Comprehensive Comparison of Distance Measures}.
\newblock \bibinfo{journal}{\emph{IEEE Data Engineering Bulletin (DEB 2023)}}  \bibinfo{volume}{47} (\bibinfo{year}{2023}), \bibinfo{pages}{69--88}.
\newblock


\bibitem[Paparrizos et~al\mbox{.}(2020)]%
        {paparrizos2020debunking}
\bibfield{author}{\bibinfo{person}{John Paparrizos}, \bibinfo{person}{Chunwei Liu}, \bibinfo{person}{Aaron~J Elmore}, {and} \bibinfo{person}{Michael~J Franklin}.} \bibinfo{year}{2020}\natexlab{}.
\newblock \showarticletitle{Debunking four long-standing misconceptions of time-series distance measures}. In \bibinfo{booktitle}{\emph{SIGMOD}}. \bibinfo{pages}{1887--1905}.
\newblock


\bibitem[Paparrizos and Reddy(2023)]%
        {paparrizos2023odyssey}
\bibfield{author}{\bibinfo{person}{John Paparrizos} {and} \bibinfo{person}{Sai Prasanna~Teja Reddy}.} \bibinfo{year}{2023}\natexlab{}.
\newblock \showarticletitle{Odyssey: An engine enabling the time-series clustering journey}.
\newblock \bibinfo{journal}{\emph{Proceedings of the VLDB Endowment}} \bibinfo{volume}{16}, \bibinfo{number}{12} (\bibinfo{year}{2023}), \bibinfo{pages}{4066--4069}.
\newblock


\bibitem[Paparrizos et~al\mbox{.}(2016a)]%
        {paparrizos2016detecting}
\bibfield{author}{\bibinfo{person}{John Paparrizos}, \bibinfo{person}{Ryen~W White}, {and} \bibinfo{person}{Eric Horvitz}.} \bibinfo{year}{2016}\natexlab{a}.
\newblock \showarticletitle{Detecting devastating diseases in search logs}. In \bibinfo{booktitle}{\emph{SIGKDD}}. \bibinfo{pages}{559--568}.
\newblock


\bibitem[Paparrizos et~al\mbox{.}(2016b)]%
        {paparrizos2016screening}
\bibfield{author}{\bibinfo{person}{John Paparrizos}, \bibinfo{person}{Ryen~W White}, {and} \bibinfo{person}{Eric Horvitz}.} \bibinfo{year}{2016}\natexlab{b}.
\newblock \showarticletitle{Screening for pancreatic adenocarcinoma using signals from web search logs: Feasibility study and results}.
\newblock \bibinfo{journal}{\emph{Journal of oncology practice}} \bibinfo{volume}{12}, \bibinfo{number}{8} (\bibinfo{year}{2016}), \bibinfo{pages}{737--744}.
\newblock


\bibitem[Paparrizos et~al\mbox{.}(2023b)]%
        {paparrizos2023accelerating}
\bibfield{author}{\bibinfo{person}{John Paparrizos}, \bibinfo{person}{Kaize Wu}, \bibinfo{person}{Aaron Elmore}, \bibinfo{person}{Christos Faloutsos}, {and} \bibinfo{person}{Michael~J Franklin}.} \bibinfo{year}{2023}\natexlab{b}.
\newblock \showarticletitle{Accelerating similarity search for elastic measures: A study and new generalization of lower bounding distances}.
\newblock \bibinfo{journal}{\emph{Proceedings of the VLDB Endowment}} \bibinfo{volume}{16}, \bibinfo{number}{8} (\bibinfo{year}{2023}), \bibinfo{pages}{2019--2032}.
\newblock


\bibitem[Park et~al\mbox{.}(2016)]%
        {ParkEtAl2016Multimodal}
\bibfield{author}{\bibinfo{person}{Daehyung Park}, \bibinfo{person}{Zackory Erickson}, \bibinfo{person}{Tapomayukh Bhattacharjee}, {and} \bibinfo{person}{Charles~C. Kemp}.} \bibinfo{year}{2016}\natexlab{}.
\newblock \showarticletitle{Multimodal Execution Monitoring for Anomaly Detection during Robot Manipulation}. In \bibinfo{booktitle}{\emph{Proceedings of the {{International Conference}} on {{Robotics}} and {{Automation}} ({{ICRA}})}}. \bibinfo{publisher}{{IEEE}}, \bibinfo{pages}{407--414}.
\newblock
\showISBNx{978-1-4673-8026-3}
\urldef\tempurl%
\url{https://doi.org/10.1109/ICRA.2016.7487160}
\showDOI{\tempurl}


\bibitem[Park et~al\mbox{.}(2018)]%
        {ParkEtAl2018Multimodal}
\bibfield{author}{\bibinfo{person}{Daehyung Park}, \bibinfo{person}{Yuuna Hoshi}, {and} \bibinfo{person}{Charles~C. Kemp}.} \bibinfo{year}{2018}\natexlab{}.
\newblock \showarticletitle{A {{Multimodal Anomaly Detector}} for {{Robot}}-{{Assisted Feeding Using}} an {{LSTM}}-{{Based Variational Autoencoder}}}.
\newblock  \bibinfo{volume}{3}, \bibinfo{number}{3} (\bibinfo{year}{2018}), \bibinfo{pages}{1544--1551}.
\newblock
\showISSN{2377-3766, 2377-3774}
\urldef\tempurl%
\url{https://doi.org/10.1109/LRA.2018.2801475}
\showDOI{\tempurl}


\bibitem[Pauwels and Calders(2019a)]%
        {PauwelsCalders2019anomaly}
\bibfield{author}{\bibinfo{person}{Stephen Pauwels} {and} \bibinfo{person}{Toon Calders}.} \bibinfo{year}{2019}\natexlab{a}.
\newblock \showarticletitle{An Anomaly Detection Technique for Business Processes Based on Extended Dynamic Bayesian Networks}. In \bibinfo{booktitle}{\emph{Proceedings of the {{ACM}}/{{SIGAPP Symposium}} on {{Applied Computing}} ({{SAC}})}}. \bibinfo{publisher}{{ACM}}, \bibinfo{pages}{494--501}.
\newblock
\showISBNx{978-1-4503-5933-7}
\urldef\tempurl%
\url{https://doi.org/10.1145/3297280.3297326}
\showDOI{\tempurl}


\bibitem[Pauwels and Calders(2019b)]%
        {PauwelsCalders2019Detecting}
\bibfield{author}{\bibinfo{person}{Stephen Pauwels} {and} \bibinfo{person}{Toon Calders}.} \bibinfo{year}{2019}\natexlab{b}.
\newblock \showarticletitle{Detecting Anomalies in Hybrid Business Process Logs}.
\newblock  \bibinfo{volume}{19}, \bibinfo{number}{2} (\bibinfo{year}{2019}), \bibinfo{pages}{18--30}.
\newblock
\showISSN{1559-6915, 1931-0161}
\urldef\tempurl%
\url{https://doi.org/10.1145/3357385.3357387}
\showDOI{\tempurl}


\bibitem[Pearson and Sekar(1936)]%
        {pearson1936efficiency}
\bibfield{author}{\bibinfo{person}{ERWIN~S Pearson} {and} \bibinfo{person}{C~Chandra Sekar}.} \bibinfo{year}{1936}\natexlab{}.
\newblock \showarticletitle{The efficiency of statistical tools and a criterion for the rejection of outlying observations}.
\newblock \bibinfo{journal}{\emph{Biometrika}} \bibinfo{volume}{28}, \bibinfo{number}{3/4} (\bibinfo{year}{1936}), \bibinfo{pages}{308--320}.
\newblock


\bibitem[Peirce(1852)]%
        {peirce1852criterion}
\bibfield{author}{\bibinfo{person}{Benjamin Peirce}.} \bibinfo{year}{1852}\natexlab{}.
\newblock \showarticletitle{Criterion for the rejection of doubtful observations}.
\newblock \bibinfo{journal}{\emph{The Astronomical Journal}}  \bibinfo{volume}{2} (\bibinfo{year}{1852}), \bibinfo{pages}{161--163}.
\newblock


\bibitem[Peng et~al\mbox{.}(1995)]%
        {peng1995quantification}
\bibfield{author}{\bibinfo{person}{C-K Peng}, \bibinfo{person}{Shlomo Havlin}, \bibinfo{person}{H~Eugene Stanley}, {and} \bibinfo{person}{Ary~L Goldberger}.} \bibinfo{year}{1995}\natexlab{}.
\newblock \showarticletitle{Quantification of scaling exponents and crossover phenomena in nonstationary heartbeat time series}.
\newblock \bibinfo{journal}{\emph{Chaos: An Interdisciplinary Journal of Nonlinear Science}} \bibinfo{volume}{5}, \bibinfo{number}{1} (\bibinfo{year}{1995}), \bibinfo{pages}{82--87}.
\newblock


\bibitem[Pokrajac et~al\mbox{.}(2007)]%
        {PokrajacEtAl2007Incremental}
\bibfield{author}{\bibinfo{person}{Dragoljub Pokrajac}, \bibinfo{person}{Aleksandar Lazarevic}, {and} \bibinfo{person}{Longin~Jan Latecki}.} \bibinfo{year}{2007}\natexlab{}.
\newblock \showarticletitle{Incremental local outlier detection for data streams}. In \bibinfo{booktitle}{\emph{2007 IEEE symposium on computational intelligence and data mining}}. IEEE, \bibinfo{pages}{504--515}.
\newblock


\bibitem[Qiu et~al\mbox{.}(2024)]%
        {qiu2024tfb}
\bibfield{author}{\bibinfo{person}{Xiangfei Qiu}, \bibinfo{person}{Jilin Hu}, \bibinfo{person}{Lekui Zhou}, \bibinfo{person}{Xingjian Wu}, \bibinfo{person}{Junyang Du}, \bibinfo{person}{Buang Zhang}, \bibinfo{person}{Chenjuan Guo}, \bibinfo{person}{Aoying Zhou}, \bibinfo{person}{Christian~S. Jensen}, \bibinfo{person}{Zhenli Sheng}, {and} \bibinfo{person}{Bin Yang}.} \bibinfo{year}{2024}\natexlab{}.
\newblock \showarticletitle{TFB: Towards Comprehensive and Fair Benchmarking of Time Series Forecasting Methods}.
\newblock \bibinfo{journal}{\emph{Proc. {VLDB} Endow.}} \bibinfo{volume}{17}, \bibinfo{number}{9} (\bibinfo{year}{2024}), \bibinfo{pages}{2363--2377}.
\newblock


\bibitem[Ramaswamy et~al\mbox{.}(2000)]%
        {10.1145/335191.335437}
\bibfield{author}{\bibinfo{person}{Sridhar Ramaswamy}, \bibinfo{person}{Rajeev Rastogi}, {and} \bibinfo{person}{Kyuseok Shim}.} \bibinfo{year}{2000}\natexlab{}.
\newblock \showarticletitle{Efficient Algorithms for Mining Outliers from Large Data Sets}.
\newblock \bibinfo{journal}{\emph{SIGMOD Rec.}} \bibinfo{volume}{29}, \bibinfo{number}{2} (\bibinfo{date}{may} \bibinfo{year}{2000}), \bibinfo{pages}{427–438}.
\newblock
\showISSN{0163-5808}
\urldef\tempurl%
\url{https://doi.org/10.1145/335191.335437}
\showDOI{\tempurl}


\bibitem[Raza et~al\mbox{.}(2015)]%
        {raza2015practical}
\bibfield{author}{\bibinfo{person}{Usman Raza}, \bibinfo{person}{Alessandro Camerra}, \bibinfo{person}{Amy~L Murphy}, \bibinfo{person}{Themis Palpanas}, {and} \bibinfo{person}{Gian~Pietro Picco}.} \bibinfo{year}{2015}\natexlab{}.
\newblock \showarticletitle{Practical data prediction for real-world wireless sensor networks}.
\newblock \bibinfo{journal}{\emph{IEEE Transactions on Knowledge and Data Engineering}} \bibinfo{volume}{27}, \bibinfo{number}{8} (\bibinfo{year}{2015}), \bibinfo{pages}{2231--2244}.
\newblock


\bibitem[Ren et~al\mbox{.}(2019)]%
        {ren2019time}
\bibfield{author}{\bibinfo{person}{Hansheng Ren}, \bibinfo{person}{Bixiong Xu}, \bibinfo{person}{Yujing Wang}, \bibinfo{person}{Chao Yi}, \bibinfo{person}{Congrui Huang}, \bibinfo{person}{Xiaoyu Kou}, \bibinfo{person}{Tony Xing}, \bibinfo{person}{Mao Yang}, \bibinfo{person}{Jie Tong}, {and} \bibinfo{person}{Qi Zhang}.} \bibinfo{year}{2019}\natexlab{}.
\newblock \showarticletitle{Time-series anomaly detection service at microsoft}. In \bibinfo{booktitle}{\emph{Proceedings of the 25th ACM SIGKDD international conference on knowledge discovery \& data mining}}. \bibinfo{pages}{3009--3017}.
\newblock


\bibitem[Rezende et~al\mbox{.}(2014)]%
        {10.5555/3044805.3045035}
\bibfield{author}{\bibinfo{person}{Danilo~Jimenez Rezende}, \bibinfo{person}{Shakir Mohamed}, {and} \bibinfo{person}{Daan Wierstra}.} \bibinfo{year}{2014}\natexlab{}.
\newblock \showarticletitle{Stochastic Backpropagation and Approximate Inference in Deep Generative Models}. In \bibinfo{booktitle}{\emph{Proceedings of the 31st International Conference on ICML - Volume 32}} (Beijing, China) \emph{(\bibinfo{series}{ICML'14})}. \bibinfo{publisher}{JMLR.org}, \bibinfo{pages}{II–1278–II–1286}.
\newblock


\bibitem[Richman and Moorman(2000)]%
        {richman2000physiological}
\bibfield{author}{\bibinfo{person}{Joshua~S Richman} {and} \bibinfo{person}{J~Randall Moorman}.} \bibinfo{year}{2000}\natexlab{}.
\newblock \showarticletitle{Physiological time-series analysis using approximate entropy and sample entropy}.
\newblock \bibinfo{journal}{\emph{American Journal of Physiology-Heart and Circulatory Physiology}} \bibinfo{volume}{278}, \bibinfo{number}{6} (\bibinfo{year}{2000}), \bibinfo{pages}{H2039--H2049}.
\newblock


\bibitem[Rong et~al\mbox{.}(2018)]%
        {rong2018locality}
\bibfield{author}{\bibinfo{person}{Kexin Rong}, \bibinfo{person}{Clara~E Yoon}, \bibinfo{person}{Karianne~J Bergen}, \bibinfo{person}{Hashem Elezabi}, \bibinfo{person}{Peter Bailis}, \bibinfo{person}{Philip Levis}, {and} \bibinfo{person}{Gregory~C Beroza}.} \bibinfo{year}{2018}\natexlab{}.
\newblock \showarticletitle{Locality-sensitive hashing for earthquake detection: A case study of scaling data-driven science}.
\newblock \bibinfo{journal}{\emph{Proceedings of the VLDB Endowment}} \bibinfo{volume}{11}, \bibinfo{number}{11} (\bibinfo{year}{2018}), \bibinfo{pages}{1674--1687}.
\newblock


\bibitem[Roth(2006)]%
        {Roth2006Kernel}
\bibfield{author}{\bibinfo{person}{Volker Roth}.} \bibinfo{year}{2006}\natexlab{}.
\newblock \showarticletitle{Kernel {{Fisher Discriminants}} for {{Outlier Detection}}}.
\newblock  \bibinfo{volume}{18}, \bibinfo{number}{4} (\bibinfo{year}{2006}), \bibinfo{pages}{942--960}.
\newblock
\showISSN{0899-7667, 1530-888X}
\urldef\tempurl%
\url{https://doi.org/10.1162/neco.2006.18.4.942}
\showDOI{\tempurl}


\bibitem[Rousseeuw(1984)]%
        {MCDfirst}
\bibfield{author}{\bibinfo{person}{Peter Rousseeuw}.} \bibinfo{year}{1984}\natexlab{}.
\newblock \showarticletitle{Least Median of Squares Regression}.
\newblock \bibinfo{journal}{\emph{Journal of The American Statistical Association - J AMER STATIST ASSN}}  \bibinfo{volume}{79} (\bibinfo{date}{12} \bibinfo{year}{1984}), \bibinfo{pages}{871--880}.
\newblock
\urldef\tempurl%
\url{https://doi.org/10.1080/01621459.1984.10477105}
\showDOI{\tempurl}


\bibitem[Rousseeuw and Driessen(1999)]%
        {Fast_MCD_1999}
\bibfield{author}{\bibinfo{person}{Peter~J. Rousseeuw} {and} \bibinfo{person}{Katrien~Van Driessen}.} \bibinfo{year}{1999}\natexlab{}.
\newblock \showarticletitle{A Fast Algorithm for the Minimum Covariance Determinant Estimator}.
\newblock \bibinfo{journal}{\emph{Technometrics}} \bibinfo{volume}{41}, \bibinfo{number}{3} (\bibinfo{year}{1999}), \bibinfo{pages}{212--223}.
\newblock
\urldef\tempurl%
\url{https://doi.org/10.1080/00401706.1999.10485670}
\showDOI{\tempurl}
\showeprint{https://www.tandfonline.com/doi/pdf/10.1080/00401706.1999.10485670}


\bibitem[Rousseeuw and Leroy(1987)]%
        {rousseeuw_robust_1987}
\bibfield{author}{\bibinfo{person}{Peter~J. Rousseeuw} {and} \bibinfo{person}{Annick~M. Leroy}.} \bibinfo{year}{1987}\natexlab{}.
\newblock \bibinfo{booktitle}{\emph{Robust regression and outlier detection}}.
\newblock \bibinfo{publisher}{Wiley}, \bibinfo{address}{New York}.
\newblock
\showISBNx{9780471852339}


\bibitem[Rousseeuw and Leroy(2005)]%
        {rousseeuw2005robust}
\bibfield{author}{\bibinfo{person}{Peter~J Rousseeuw} {and} \bibinfo{person}{Annick~M Leroy}.} \bibinfo{year}{2005}\natexlab{}.
\newblock \bibinfo{booktitle}{\emph{Robust regression and outlier detection}}. Vol.~\bibinfo{volume}{589}.
\newblock \bibinfo{publisher}{John wiley \& sons}.
\newblock


\bibitem[Ruiz et~al\mbox{.}(2012)]%
        {ruiz2012correlating}
\bibfield{author}{\bibinfo{person}{Eduardo~J Ruiz}, \bibinfo{person}{Vagelis Hristidis}, \bibinfo{person}{Carlos Castillo}, \bibinfo{person}{Aristides Gionis}, {and} \bibinfo{person}{Alejandro Jaimes}.} \bibinfo{year}{2012}\natexlab{}.
\newblock \showarticletitle{Correlating financial time series with micro-blogging activity}. In \bibinfo{booktitle}{\emph{Proceedings of the fifth ACM international conference on Web search and data mining}}. ACM, \bibinfo{pages}{513--522}.
\newblock


\bibitem[Salvador and Chan(2005)]%
        {SalvadorChan2005Learning}
\bibfield{author}{\bibinfo{person}{Stan Salvador} {and} \bibinfo{person}{Philip Chan}.} \bibinfo{year}{2005}\natexlab{}.
\newblock \showarticletitle{Learning {{States}} and {{Rules}} for {{Detecting Anomalies}} in {{Time Series}}}.
\newblock  \bibinfo{volume}{23}, \bibinfo{number}{3} (\bibinfo{year}{2005}), \bibinfo{pages}{241--255}.
\newblock
\showISSN{0924-669X, 1573-7497}
\urldef\tempurl%
\url{https://doi.org/10.1007/s10489-005-4610-3}
\showDOI{\tempurl}


\bibitem[Sander et~al\mbox{.}(1998)]%
        {Sander1998}
\bibfield{author}{\bibinfo{person}{J{\"o}rg Sander}, \bibinfo{person}{Martin Ester}, \bibinfo{person}{Hans-Peter Kriegel}, {and} \bibinfo{person}{Xiaowei Xu}.} \bibinfo{year}{1998}\natexlab{}.
\newblock \showarticletitle{Density-Based Clustering in Spatial Databases: The Algorithm GDBSCAN and Its Applications}.
\newblock \bibinfo{journal}{\emph{Data Mining and Knowledge Discovery}} \bibinfo{volume}{2}, \bibinfo{number}{2} (\bibinfo{date}{01 Jun} \bibinfo{year}{1998}), \bibinfo{pages}{169--194}.
\newblock
\showISSN{1573-756X}
\urldef\tempurl%
\url{https://doi.org/10.1023/A:1009745219419}
\showDOI{\tempurl}


\bibitem[Schmidl et~al\mbox{.}(2022)]%
        {10.14778/3538598.3538602}
\bibfield{author}{\bibinfo{person}{Sebastian Schmidl}, \bibinfo{person}{Phillip Wenig}, {and} \bibinfo{person}{Thorsten Papenbrock}.} \bibinfo{year}{2022}\natexlab{}.
\newblock \showarticletitle{Anomaly Detection in Time Series: A Comprehensive Evaluation}.
\newblock \bibinfo{journal}{\emph{PVLDB}} \bibinfo{volume}{15}, \bibinfo{number}{9} (\bibinfo{date}{may} \bibinfo{year}{2022}), \bibinfo{pages}{1779–1797}.
\newblock
\showISSN{2150-8097}
\urldef\tempurl%
\url{https://doi.org/10.14778/3538598.3538602}
\showDOI{\tempurl}


\bibitem[Schneider et~al\mbox{.}(2021)]%
        {10.1007/s00778-021-00657-6}
\bibfield{author}{\bibinfo{person}{Johannes Schneider}, \bibinfo{person}{Phillip Wenig}, {and} \bibinfo{person}{Thorsten Papenbrock}.} \bibinfo{year}{2021}\natexlab{}.
\newblock \showarticletitle{Distributed Detection of Sequential Anomalies in Univariate Time Series}.
\newblock \bibinfo{journal}{\emph{The VLDB Journal}} \bibinfo{volume}{30}, \bibinfo{number}{4} (\bibinfo{date}{mar} \bibinfo{year}{2021}), \bibinfo{pages}{579–602}.
\newblock
\showISSN{1066-8888}
\urldef\tempurl%
\url{https://doi.org/10.1007/s00778-021-00657-6}
\showDOI{\tempurl}


\bibitem[Sch{\"o}lkopf et~al\mbox{.}(1999)]%
        {NIPS1999_1723}
\bibfield{author}{\bibinfo{person}{Bernhard Sch{\"o}lkopf}, \bibinfo{person}{Robert~C Williamson}, \bibinfo{person}{Alex Smola}, \bibinfo{person}{John Shawe-Taylor}, {and} \bibinfo{person}{John Platt}.} \bibinfo{year}{1999}\natexlab{}.
\newblock \showarticletitle{Support vector method for novelty detection}.
\newblock \bibinfo{journal}{\emph{NeurIPS}}  \bibinfo{volume}{12} (\bibinfo{year}{1999}).
\newblock


\bibitem[Senin et~al\mbox{.}(2015a)]%
        {DBLP:conf/edbt/Senin0WOGBCF15}
\bibfield{author}{\bibinfo{person}{Pavel Senin}, \bibinfo{person}{Jessica Lin}, \bibinfo{person}{Xing Wang}, \bibinfo{person}{Tim Oates}, \bibinfo{person}{Sunil Gandhi}, \bibinfo{person}{Arnold~P. Boedihardjo}, \bibinfo{person}{Crystal Chen}, {and} \bibinfo{person}{Susan Frankenstein}.} \bibinfo{year}{2015}\natexlab{a}.
\newblock \showarticletitle{Time series anomaly discovery with grammar-based compression}. In \bibinfo{booktitle}{\emph{EDBT}}.
\newblock


\bibitem[Senin et~al\mbox{.}(2015b)]%
        {SeninEtAl2015Time}
\bibfield{author}{\bibinfo{person}{Pavel Senin}, \bibinfo{person}{Jessica Lin}, \bibinfo{person}{Xing Wang}, \bibinfo{person}{Tim Oates}, \bibinfo{person}{Sunil Gandhi}, \bibinfo{person}{Arnold~P. Boedihardjo}, \bibinfo{person}{Crystal Chen}, {and} \bibinfo{person}{Susan Frankenstein}.} \bibinfo{year}{2015}\natexlab{b}.
\newblock \bibinfo{title}{Time Series Anomaly Discovery with Grammar-Based Compression}.
\newblock , \bibinfo{numpages}{481--492}~pages.
\newblock
\urldef\tempurl%
\url{https://doi.org/10.5441/002/edbt.2015.42}
\showDOI{\tempurl}


\bibitem[Shasha(1999)]%
        {shasha1999tuning}
\bibfield{author}{\bibinfo{person}{Dennis Shasha}.} \bibinfo{year}{1999}\natexlab{}.
\newblock \showarticletitle{Tuning time series queries in finance: Case studies and recommendations}.
\newblock \bibinfo{journal}{\emph{IEEE Data Eng. Bull.}} \bibinfo{volume}{22}, \bibinfo{number}{2} (\bibinfo{year}{1999}), \bibinfo{pages}{40--46}.
\newblock


\bibitem[Shen et~al\mbox{.}(2020)]%
        {shen2020timeseries}
\bibfield{author}{\bibinfo{person}{Lifeng Shen}, \bibinfo{person}{Zhuocong Li}, {and} \bibinfo{person}{James Kwok}.} \bibinfo{year}{2020}\natexlab{}.
\newblock \showarticletitle{Timeseries anomaly detection using temporal hierarchical one-class network}.
\newblock \bibinfo{journal}{\emph{NeurIPS}}  \bibinfo{volume}{33} (\bibinfo{year}{2020}), \bibinfo{pages}{13016--13026}.
\newblock


\bibitem[Shyu and Chang(2003)]%
        {pcal_2003}
\bibfield{author}{\bibinfo{person}{S-C. Chen K.~Sarinnapakorn Shyu, M-L.} {and} \bibinfo{person}{LW. Chang}.} \bibinfo{year}{2003}\natexlab{}.
\newblock \showarticletitle{A Novel Anomaly Detection Scheme Based on Principal Component Classifier}.
\newblock  (\bibinfo{year}{2003}).
\newblock


\bibitem[Siffer et~al\mbox{.}(2017)]%
        {siffer2017anomaly}
\bibfield{author}{\bibinfo{person}{Alban Siffer}, \bibinfo{person}{Pierre-Alain Fouque}, \bibinfo{person}{Alexandre Termier}, {and} \bibinfo{person}{Christine Largouet}.} \bibinfo{year}{2017}\natexlab{}.
\newblock \showarticletitle{Anomaly detection in streams with extreme value theory}. In \bibinfo{booktitle}{\emph{Proceedings of the 23rd ACM SIGKDD International Conference on Knowledge Discovery and Data Mining}}. \bibinfo{pages}{1067--1075}.
\newblock


\bibitem[Smith(2012)]%
        {smith2012source}
\bibfield{author}{\bibinfo{person}{David~Eugene Smith}.} \bibinfo{year}{2012}\natexlab{}.
\newblock \bibinfo{booktitle}{\emph{A source book in mathematics}}.
\newblock \bibinfo{publisher}{Courier Corporation}.
\newblock


\bibitem[Snyder and Withers(1983)]%
        {snyder_exponential_1983}
\bibfield{author}{\bibinfo{person}{Ralph~D. Snyder} {and} \bibinfo{person}{Stephen~J. Withers}.} \bibinfo{year}{1983}\natexlab{}.
\newblock \bibinfo{booktitle}{\emph{Exponential smoothing with finite sample correction}}.
\newblock Number 1983,1 in \bibinfo{series}{Working paper. {Department} of {Econometrics} and {Operations} {Research}. {Monash} {University}}. \bibinfo{publisher}{Dept., Univ}, \bibinfo{address}{Clayton}.
\newblock
\showISBNx{9780867464894}


\bibitem[Soelch et~al\mbox{.}(2016)]%
        {SoelchEtAl2016Variational}
\bibfield{author}{\bibinfo{person}{Maximilian Soelch}, \bibinfo{person}{Justin Bayer}, \bibinfo{person}{Marvin Ludersdorfer}, {and} \bibinfo{person}{Patrick van~der Smagt}.} \bibinfo{year}{2016}\natexlab{}.
\newblock \bibinfo{booktitle}{\emph{Variational {{Inference}} for {{On}}-Line {{Anomaly Detection}} in {{High}}-{{Dimensional Time Series}}}}.
\newblock
\showeprint[arxiv]{1602.07109}~[cs, stat]
\urldef\tempurl%
\url{http://arxiv.org/abs/1602.07109}
\showURL{%
\tempurl}


\bibitem[Song et~al\mbox{.}(2017)]%
        {Song2017}
\bibfield{author}{\bibinfo{person}{Hongchao Song}, \bibinfo{person}{Zhuqing Jiang}, \bibinfo{person}{Aidong Men}, {and} \bibinfo{person}{Bo Yang}.} \bibinfo{year}{2017}\natexlab{}.
\newblock \showarticletitle{A Hybrid Semi-Supervised Anomaly Detection Model for High-Dimensional Data}.
\newblock \bibinfo{journal}{\emph{Computational Intelligence and Neuroscience}}  \bibinfo{volume}{2017} (\bibinfo{date}{15 Nov} \bibinfo{year}{2017}), \bibinfo{pages}{8501683}.
\newblock
\showISSN{1687-5265}
\urldef\tempurl%
\url{https://doi.org/10.1155/2017/8501683}
\showDOI{\tempurl}


\bibitem[S{\o}rb{\o} and Ruocco(2023)]%
        {sorbo2023navigating}
\bibfield{author}{\bibinfo{person}{Sondre S{\o}rb{\o}} {and} \bibinfo{person}{Massimiliano Ruocco}.} \bibinfo{year}{2023}\natexlab{}.
\newblock \showarticletitle{Navigating the Metric Maze: A Taxonomy of Evaluation Metrics for Anomaly Detection in Time Series}.
\newblock \bibinfo{journal}{\emph{arXiv preprint arXiv:2303.01272}} (\bibinfo{year}{2023}).
\newblock


\bibitem[Stone(1873)]%
        {stone1873rejection}
\bibfield{author}{\bibinfo{person}{EJ Stone}.} \bibinfo{year}{1873}\natexlab{}.
\newblock \showarticletitle{On the rejection of discordant observations}.
\newblock \bibinfo{journal}{\emph{Monthly Notices of the Royal Astronomical Society}}  \bibinfo{volume}{34} (\bibinfo{year}{1873}), \bibinfo{pages}{9}.
\newblock


\bibitem[Su et~al\mbox{.}(2019a)]%
        {SuEtAl2019Robust}
\bibfield{author}{\bibinfo{person}{Ya Su}, \bibinfo{person}{Youjian Zhao}, \bibinfo{person}{Chenhao Niu}, \bibinfo{person}{Rong Liu}, \bibinfo{person}{Wei Sun}, {and} \bibinfo{person}{Dan Pei}.} \bibinfo{year}{2019}\natexlab{a}.
\newblock \showarticletitle{Robust {{Anomaly Detection}} for {{Multivariate Time Series}} through {{Stochastic Recurrent Neural Network}}}. In \bibinfo{booktitle}{\emph{SIGKDD}}. \bibinfo{publisher}{{ACM}}, \bibinfo{pages}{2828--2837}.
\newblock
\showISBNx{978-1-4503-6201-6}
\urldef\tempurl%
\url{https://doi.org/10.1145/3292500.3330672}
\showDOI{\tempurl}


\bibitem[Su et~al\mbox{.}(2019b)]%
        {su2019robust}
\bibfield{author}{\bibinfo{person}{Ya Su}, \bibinfo{person}{Youjian Zhao}, \bibinfo{person}{Chenhao Niu}, \bibinfo{person}{Rong Liu}, \bibinfo{person}{Wei Sun}, {and} \bibinfo{person}{Dan Pei}.} \bibinfo{year}{2019}\natexlab{b}.
\newblock \showarticletitle{Robust anomaly detection for multivariate time series through stochastic recurrent neural network}. In \bibinfo{booktitle}{\emph{SIGKDD}}. \bibinfo{pages}{2828--2837}.
\newblock


\bibitem[Subramaniam et~al\mbox{.}(2006)]%
        {SubramaniamEtAl2006Online}
\bibfield{author}{\bibinfo{person}{S. Subramaniam}, \bibinfo{person}{T. Palpanas}, \bibinfo{person}{D. Papadopoulos}, \bibinfo{person}{V. Kalogeraki}, {and} \bibinfo{person}{D. Gunopulos}.} \bibinfo{year}{2006}\natexlab{}.
\newblock \showarticletitle{Online Outlier Detection in Sensor Data Using Non-Parametric Models}. In \bibinfo{booktitle}{\emph{Proceedings of the {{International Conference}} on {{Very Large Databases}} ({{VLDB}})}} \emph{(\bibinfo{series}{{{VLDB}} '06})}. \bibinfo{publisher}{{VLDB Endowment}}, \bibinfo{pages}{187--198}.
\newblock
\urldef\tempurl%
\url{https://dl.acm.org/doi/10.5555/1182635.1164145}
\showURL{%
\tempurl}


\bibitem[Sun et~al\mbox{.}(2006)]%
        {SunEtAl2006Mining}
\bibfield{author}{\bibinfo{person}{Pei Sun}, \bibinfo{person}{Sanjay Chawla}, {and} \bibinfo{person}{Bavani Arunasalam}.} \bibinfo{year}{2006}\natexlab{}.
\newblock \showarticletitle{Mining for {{Outliers}} in {{Sequential Databases}}}. In \bibinfo{booktitle}{\emph{Proceedings of the {{International Conference}} on {{Data Mining}} ({{ICDM}})}}. \bibinfo{publisher}{{Society for Industrial and Applied Mathematics}}, \bibinfo{pages}{94--105}.
\newblock
\showISBNx{978-0-89871-611-5 978-1-61197-276-4}
\urldef\tempurl%
\url{https://doi.org/10.1137/1.9781611972764.9}
\showDOI{\tempurl}


\bibitem[Sylligardos et~al\mbox{.}(2023)]%
        {msad-sylligardos23}
\bibfield{author}{\bibinfo{person}{Emmanouil Sylligardos}, \bibinfo{person}{Paul Boniol}, \bibinfo{person}{John Paparrizos}, \bibinfo{person}{Panos~E. Trahanias}, {and} \bibinfo{person}{Themis Palpanas}.} \bibinfo{year}{2023}\natexlab{}.
\newblock \showarticletitle{Choose Wisely: An Extensive Evaluation of Model Selection for Anomaly Detection in Time Series}.
\newblock \bibinfo{journal}{\emph{Proc. {VLDB} Endow.}} \bibinfo{volume}{16}, \bibinfo{number}{11} (\bibinfo{year}{2023}), \bibinfo{pages}{3418--3432}.
\newblock
\urldef\tempurl%
\url{https://doi.org/10.14778/3611479.3611536}
\showDOI{\tempurl}


\bibitem[Tang et~al\mbox{.}(2002)]%
        {TangEtAl2002Enhancing}
\bibfield{author}{\bibinfo{person}{Jian Tang}, \bibinfo{person}{Zhixiang Chen}, \bibinfo{person}{Ada Wai-Chee Fu}, {and} \bibinfo{person}{David~W Cheung}.} \bibinfo{year}{2002}\natexlab{}.
\newblock \showarticletitle{Enhancing effectiveness of outlier detections for low density patterns}. In \bibinfo{booktitle}{\emph{PAKDD}}. \bibinfo{pages}{535--548}.
\newblock


\bibitem[Tatbul et~al\mbox{.}(2018)]%
        {tatbul2018precision}
\bibfield{author}{\bibinfo{person}{Nesime Tatbul}, \bibinfo{person}{Tae~Jun Lee}, \bibinfo{person}{Stan Zdonik}, \bibinfo{person}{Mejbah Alam}, {and} \bibinfo{person}{Justin Gottschlich}.} \bibinfo{year}{2018}\natexlab{}.
\newblock \showarticletitle{Precision and recall for time series}. In \bibinfo{booktitle}{\emph{NeurIPS}}. \bibinfo{pages}{1924--1934}.
\newblock


\bibitem[Tax and Duin(2004)]%
        {tax2004support}
\bibfield{author}{\bibinfo{person}{David~MJ Tax} {and} \bibinfo{person}{Robert~PW Duin}.} \bibinfo{year}{2004}\natexlab{}.
\newblock \showarticletitle{Support vector data description}.
\newblock \bibinfo{journal}{\emph{Machine learning}} \bibinfo{volume}{54}, \bibinfo{number}{1} (\bibinfo{year}{2004}), \bibinfo{pages}{45--66}.
\newblock


\bibitem[Tsay(1988)]%
        {tsay1988outliers}
\bibfield{author}{\bibinfo{person}{Ruey~S Tsay}.} \bibinfo{year}{1988}\natexlab{}.
\newblock \showarticletitle{Outliers, level shifts, and variance changes in time series}.
\newblock \bibinfo{journal}{\emph{Journal of forecasting}} \bibinfo{volume}{7}, \bibinfo{number}{1} (\bibinfo{year}{1988}), \bibinfo{pages}{1--20}.
\newblock


\bibitem[Tsay(2014)]%
        {tsay2014financial}
\bibfield{author}{\bibinfo{person}{Ruey~S Tsay}.} \bibinfo{year}{2014}\natexlab{}.
\newblock \showarticletitle{Financial Time Series}.
\newblock \bibinfo{journal}{\emph{Wiley StatsRef: Statistics Reference Online}} (\bibinfo{year}{2014}), \bibinfo{pages}{1--23}.
\newblock


\bibitem[Tsay et~al\mbox{.}(2000)]%
        {tsay2000outliers}
\bibfield{author}{\bibinfo{person}{Ruey~S Tsay}, \bibinfo{person}{Daniel Pena}, {and} \bibinfo{person}{Alan~E Pankratz}.} \bibinfo{year}{2000}\natexlab{}.
\newblock \showarticletitle{Outliers in multivariate time series}.
\newblock \bibinfo{journal}{\emph{Biometrika}} \bibinfo{volume}{87}, \bibinfo{number}{4} (\bibinfo{year}{2000}), \bibinfo{pages}{789--804}.
\newblock


\bibitem[Tukey et~al\mbox{.}(1977)]%
        {tukey1977exploratory}
\bibfield{author}{\bibinfo{person}{John~W Tukey} {et~al\mbox{.}}} \bibinfo{year}{1977}\natexlab{}.
\newblock \bibinfo{booktitle}{\emph{Exploratory data analysis}}. Vol.~\bibinfo{volume}{2}.
\newblock \bibinfo{publisher}{Reading, Mass.}
\newblock


\bibitem[Uehara and Shimada(2002)]%
        {uehara2002extraction}
\bibfield{author}{\bibinfo{person}{Kuniaki Uehara} {and} \bibinfo{person}{Mitsuomi Shimada}.} \bibinfo{year}{2002}\natexlab{}.
\newblock \showarticletitle{Extraction of primitive motion and discovery of association rules from human motion data}.
\newblock In \bibinfo{booktitle}{\emph{Progress in Discovery Science}}. \bibinfo{publisher}{Springer}, \bibinfo{pages}{338--348}.
\newblock


\bibitem[Vieira et~al\mbox{.}(2018)]%
        {VieiraEtAl2018Enhanced}
\bibfield{author}{\bibinfo{person}{Rafael~G. Vieira}, \bibinfo{person}{Marcos A.~Leone Filho}, {and} \bibinfo{person}{Robinson Semolini}.} \bibinfo{year}{2018}\natexlab{}.
\newblock \showarticletitle{An {{Enhanced Seasonal}}-{{Hybrid ESD Technique}} for {{Robust Anomaly Detection}} on {{Time Series}}}. In \bibinfo{booktitle}{\emph{Simpósio {{Brasileiro}} de {{Redes}} de {{Computadores}} ({{SBRC}})}}, Vol.~\bibinfo{volume}{36}.
\newblock


\bibitem[Wachman et~al\mbox{.}(2009)]%
        {wachman2009kernels}
\bibfield{author}{\bibinfo{person}{Gabriel Wachman}, \bibinfo{person}{Roni Khardon}, \bibinfo{person}{Pavlos Protopapas}, {and} \bibinfo{person}{Charles~R Alcock}.} \bibinfo{year}{2009}\natexlab{}.
\newblock \showarticletitle{Kernels for periodic time series arising in astronomy}. In \bibinfo{booktitle}{\emph{Joint European Conference on Machine Learning and Knowledge Discovery in Databases}}. Springer, \bibinfo{pages}{489--505}.
\newblock


\bibitem[Wang et~al\mbox{.}(2019)]%
        {WangEtAl2019Study}
\bibfield{author}{\bibinfo{person}{Yi Wang}, \bibinfo{person}{Linsheng Han}, \bibinfo{person}{Wei Liu}, \bibinfo{person}{Shujia Yang}, {and} \bibinfo{person}{Yanbo Gao}.} \bibinfo{year}{2019}\natexlab{}.
\newblock \showarticletitle{Study on Wavelet Neural Network Based Anomaly Detection in Ocean Observing Data Series}.
\newblock   \bibinfo{volume}{186} (\bibinfo{year}{2019}), \bibinfo{pages}{106129}.
\newblock
\showISSN{00298018}
\urldef\tempurl%
\url{https://doi.org/10.1016/j.oceaneng.2019.106129}
\showDOI{\tempurl}


\bibitem[Webster et~al\mbox{.}(2005)]%
        {webster2005changes}
\bibfield{author}{\bibinfo{person}{Peter~J Webster}, \bibinfo{person}{Greg~J Holland}, \bibinfo{person}{Judith~A Curry}, {and} \bibinfo{person}{H-R Chang}.} \bibinfo{year}{2005}\natexlab{}.
\newblock \showarticletitle{Changes in tropical cyclone number, duration, and intensity in a warming environment}.
\newblock \bibinfo{journal}{\emph{Science}} \bibinfo{volume}{309}, \bibinfo{number}{5742} (\bibinfo{year}{2005}), \bibinfo{pages}{1844--1846}.
\newblock


\bibitem[Williams and Hoel(2003)]%
        {williams2003modeling}
\bibfield{author}{\bibinfo{person}{Billy~M Williams} {and} \bibinfo{person}{Lester~A Hoel}.} \bibinfo{year}{2003}\natexlab{}.
\newblock \showarticletitle{Modeling and forecasting vehicular traffic flow as a seasonal ARIMA process: Theoretical basis and empirical results}.
\newblock \bibinfo{journal}{\emph{Journal of transportation engineering}} \bibinfo{volume}{129}, \bibinfo{number}{6} (\bibinfo{year}{2003}), \bibinfo{pages}{664--672}.
\newblock


\bibitem[Wu et~al\mbox{.}(2018)]%
        {WuEtAl2018Hierarchical}
\bibfield{author}{\bibinfo{person}{Jia Wu}, \bibinfo{person}{Weiru Zeng}, {and} \bibinfo{person}{Fei Yan}.} \bibinfo{year}{2018}\natexlab{}.
\newblock \showarticletitle{Hierarchical {{Temporal Memory}} Method for Time-Series-Based Anomaly Detection}.
\newblock   \bibinfo{volume}{273} (\bibinfo{year}{2018}), \bibinfo{pages}{535--546}.
\newblock
\showISSN{0925-2312}
\urldef\tempurl%
\url{https://doi.org/10.1016/j.neucom.2017.08.026}
\showDOI{\tempurl}


\bibitem[{Wu} et~al\mbox{.}(2020)]%
        {deepocsvm}
\bibfield{author}{\bibinfo{person}{P. {Wu}}, \bibinfo{person}{J. {Liu}}, {and} \bibinfo{person}{F. {Shen}}.} \bibinfo{year}{2020}\natexlab{}.
\newblock \showarticletitle{A Deep One-Class Neural Network for Anomalous Event Detection in Complex Scenes}.
\newblock \bibinfo{journal}{\emph{IEEE Transactions on Neural Networks and Learning Systems}} \bibinfo{volume}{31}, \bibinfo{number}{7} (\bibinfo{year}{2020}), \bibinfo{pages}{2609--2622}.
\newblock
\urldef\tempurl%
\url{https://doi.org/10.1109/TNNLS.2019.2933554}
\showDOI{\tempurl}


\bibitem[Wu and Keogh(2023)]%
        {wu2020currentTkde}
\bibfield{author}{\bibinfo{person}{Renjie Wu} {and} \bibinfo{person}{Eamonn~J. Keogh}.} \bibinfo{year}{2023}\natexlab{}.
\newblock \showarticletitle{Current Time Series Anomaly Detection Benchmarks are Flawed and are Creating the Illusion of Progress}.
\newblock \bibinfo{journal}{\emph{{IEEE} Trans. Knowl. Data Eng.}} \bibinfo{volume}{35}, \bibinfo{number}{3} (\bibinfo{year}{2023}), \bibinfo{pages}{2421--2429}.
\newblock
\urldef\tempurl%
\url{https://doi.org/10.1109/TKDE.2021.3112126}
\showDOI{\tempurl}


\bibitem[Wu et~al\mbox{.}(2020)]%
        {WuEtAl2020Developing}
\bibfield{author}{\bibinfo{person}{Wentai Wu}, \bibinfo{person}{Ligang He}, \bibinfo{person}{Weiwei Lin}, \bibinfo{person}{Yi Su}, \bibinfo{person}{Yuhua Cui}, \bibinfo{person}{Carsten Maple}, {and} \bibinfo{person}{Stephen Jarvis}.} \bibinfo{year}{2020}\natexlab{}.
\newblock \bibinfo{booktitle}{\emph{Developing an {{Unsupervised Real}}-Time {{Anomaly Detection Scheme}} for {{Time Series}} with {{Multi}}-Seasonality}}.
\newblock
\showeprint[arxiv]{1908.01146}~[cs, eess, stat]
\urldef\tempurl%
\url{http://arxiv.org/abs/1908.01146}
\showURL{%
\tempurl}


\bibitem[Xiao et~al\mbox{.}(2009)]%
        {XiaoEtAl2009Multisphere}
\bibfield{author}{\bibinfo{person}{Yanshan Xiao}, \bibinfo{person}{Bo Liu}, \bibinfo{person}{Longbing Cao}, \bibinfo{person}{Xindong Wu}, \bibinfo{person}{Chengqi Zhang}, \bibinfo{person}{Zhifeng Hao}, \bibinfo{person}{Fengzhao Yang}, {and} \bibinfo{person}{Jie Cao}.} \bibinfo{year}{2009}\natexlab{}.
\newblock \showarticletitle{Multi-sphere support vector data description for outliers detection on multi-distribution data}. In \bibinfo{booktitle}{\emph{2009 IEEE international conference on data mining workshops}}. IEEE, \bibinfo{pages}{82--87}.
\newblock


\bibitem[Xu et~al\mbox{.}(2018a)]%
        {XuEtAl2018Unsupervised}
\bibfield{author}{\bibinfo{person}{Haowen Xu}, \bibinfo{person}{Wenxiao Chen}, \bibinfo{person}{Nengwen Zhao}, \bibinfo{person}{Zeyan Li}, \bibinfo{person}{Jiahao Bu}, \bibinfo{person}{Zhihan Li}, \bibinfo{person}{Ying Liu}, \bibinfo{person}{Youjian Zhao}, \bibinfo{person}{Dan Pei}, \bibinfo{person}{Yang Feng}, {et~al\mbox{.}}} \bibinfo{year}{2018}\natexlab{a}.
\newblock \showarticletitle{Unsupervised Anomaly Detection via Variational Auto-Encoder for Seasonal {{KPIs}} in Web Applications}. In \bibinfo{booktitle}{\emph{Proceedings of the {{International Conference}} on {{World Wide Web}} ({{WWW}})}}. {International World Wide Web Conferences Steering Committee}, \bibinfo{publisher}{{International World Wide Web Conferences Steering Committee}}, \bibinfo{pages}{187--196}.
\newblock
\urldef\tempurl%
\url{https://doi.org/10.1145/3178876.3185996}
\showDOI{\tempurl}


\bibitem[Xu et~al\mbox{.}(2018b)]%
        {xu2018unsupervised}
\bibfield{author}{\bibinfo{person}{Haowen Xu}, \bibinfo{person}{Wenxiao Chen}, \bibinfo{person}{Nengwen Zhao}, \bibinfo{person}{Zeyan Li}, \bibinfo{person}{Jiahao Bu}, \bibinfo{person}{Zhihan Li}, \bibinfo{person}{Ying Liu}, \bibinfo{person}{Youjian Zhao}, \bibinfo{person}{Dan Pei}, \bibinfo{person}{Yang Feng}, {et~al\mbox{.}}} \bibinfo{year}{2018}\natexlab{b}.
\newblock \showarticletitle{Unsupervised anomaly detection via variational auto-encoder for seasonal kpis in web applications}. In \bibinfo{booktitle}{\emph{Proceedings of the 2018 world wide web conference}}. \bibinfo{pages}{187--196}.
\newblock


\bibitem[Yamanishi et~al\mbox{.}(2004)]%
        {YamanishiEtAl2004OnLIne}
\bibfield{author}{\bibinfo{person}{Kenji Yamanishi}, \bibinfo{person}{Jun-ichi Takeuchi}, \bibinfo{person}{Graham Williams}, {and} \bibinfo{person}{Peter Milne}.} \bibinfo{year}{2004}\natexlab{}.
\newblock \showarticletitle{On-{{Line Unsupervised Outlier Detection Using Finite Mixtures}} with {{Discounting Learning Algorithms}}}.
\newblock  \bibinfo{volume}{8}, \bibinfo{number}{3} (\bibinfo{year}{2004}), \bibinfo{pages}{275--300}.
\newblock
\showISSN{1384-5810}
\urldef\tempurl%
\url{https://doi.org/10.1023/B:DAMI.0000023676.72185.7c}
\showDOI{\tempurl}


\bibitem[Yang and Liao(2017)]%
        {YangLiao2017Adjacent}
\bibfield{author}{\bibinfo{person}{Chao-Lung Yang} {and} \bibinfo{person}{Wei-Ju Liao}.} \bibinfo{year}{2017}\natexlab{}.
\newblock \showarticletitle{Adjacent Mean Difference (AMD) method for dynamic segmentation in time series anomaly detection}. In \bibinfo{booktitle}{\emph{2017 IEEE/SICE International Symposium on System Integration (SII)}}. IEEE, \bibinfo{pages}{241--246}.
\newblock


\bibitem[Yankov et~al\mbox{.}(2008)]%
        {YankovEtAl2007Disk}
\bibfield{author}{\bibinfo{person}{Dragomir Yankov}, \bibinfo{person}{Eamonn Keogh}, {and} \bibinfo{person}{Umaa Rebbapragada}.} \bibinfo{year}{2008}\natexlab{}.
\newblock \showarticletitle{Disk aware discord discovery: Finding unusual time series in terabyte sized datasets}.
\newblock \bibinfo{journal}{\emph{Knowledge and Information Systems}}  \bibinfo{volume}{17} (\bibinfo{year}{2008}), \bibinfo{pages}{241--262}.
\newblock


\bibitem[Yankov et~al\mbox{.}(2007)]%
        {DBLP:conf/icdm/YankovKR07}
\bibfield{author}{\bibinfo{person}{Dragomir Yankov}, \bibinfo{person}{Eamonn~J. Keogh}, {and} \bibinfo{person}{Umaa Rebbapragada}.} \bibinfo{year}{2007}\natexlab{}.
\newblock \showarticletitle{Disk Aware Discord Discovery: Finding Unusual Time Series in Terabyte Sized Datasets}. In \bibinfo{booktitle}{\emph{ICDM}}.
\newblock


\bibitem[Yao et~al\mbox{.}(2010)]%
        {10.1016/j.peva.2010.08.018}
\bibfield{author}{\bibinfo{person}{Yuan Yao}, \bibinfo{person}{Abhishek Sharma}, \bibinfo{person}{Leana Golubchik}, {and} \bibinfo{person}{Ramesh Govindan}.} \bibinfo{year}{2010}\natexlab{}.
\newblock \showarticletitle{Online Anomaly Detection for Sensor Systems: A Simple and Efficient Approach}.
\newblock \bibinfo{journal}{\emph{Perform. Eval.}} \bibinfo{volume}{67}, \bibinfo{number}{11} (\bibinfo{date}{nov} \bibinfo{year}{2010}), \bibinfo{pages}{1059–1075}.
\newblock
\showISSN{0166-5316}
\urldef\tempurl%
\url{https://doi.org/10.1016/j.peva.2010.08.018}
\showDOI{\tempurl}


\bibitem[Yeh et~al\mbox{.}(2016a)]%
        {DBLP:conf/icdm/YehZUBDDSMK16}
\bibfield{author}{\bibinfo{person}{C.{-}C.M. Yeh}, \bibinfo{person}{Y. Zhu}, \bibinfo{person}{L. Ulanova}, \bibinfo{person}{N. Begum}, \bibinfo{person}{Y. Ding}, \bibinfo{person}{H.A. Dau}, \bibinfo{person}{D.F. Silva}, \bibinfo{person}{A. Mueen}, {and} \bibinfo{person}{E.J. Keogh}.} \bibinfo{year}{2016}\natexlab{a}.
\newblock \showarticletitle{Matrix Profile {I:} All Pairs Similarity Joins for Time Series: {A} Unifying View That Includes Motifs, Discords and Shapelets}. In \bibinfo{booktitle}{\emph{ICDM}}.
\newblock


\bibitem[Yeh et~al\mbox{.}(2016b)]%
        {YehEtAl2016Matrix}
\bibfield{author}{\bibinfo{person}{Chin-Chia~Michael Yeh}, \bibinfo{person}{Yan Zhu}, \bibinfo{person}{Liudmila Ulanova}, \bibinfo{person}{Nurjahan Begum}, \bibinfo{person}{Yifei Ding}, \bibinfo{person}{Hoang~Anh Dau}, \bibinfo{person}{Diego~Furtado Silva}, \bibinfo{person}{Abdullah Mueen}, {and} \bibinfo{person}{Eamonn Keogh}.} \bibinfo{year}{2016}\natexlab{b}.
\newblock \showarticletitle{Matrix {{Profile I}}: {{All Pairs Similarity Joins}} for {{Time Series}}: {{A Unifying View That Includes Motifs}}, {{Discords}} and {{Shapelets}}}. In \bibinfo{booktitle}{\emph{Proceedings of the {{International Conference}} on {{Data Mining}} ({{ICDM}})}}. \bibinfo{pages}{1317--1322}.
\newblock
\showISSN{2374-8486}
\urldef\tempurl%
\url{https://doi.org/10.1109/ICDM.2016.0179}
\showDOI{\tempurl}


\bibitem[Yu et~al\mbox{.}(2014)]%
        {YuEtAl2014Time}
\bibfield{author}{\bibinfo{person}{Yufeng Yu}, \bibinfo{person}{Yuelong Zhu}, \bibinfo{person}{Shijin Li}, {and} \bibinfo{person}{Dingsheng Wan}.} \bibinfo{year}{2014}\natexlab{}.
\newblock \showarticletitle{Time {{Series Outlier Detection Based}} on {{Sliding Window Prediction}}}.
\newblock   \bibinfo{volume}{2014} (\bibinfo{year}{2014}), \bibinfo{pages}{1--14}.
\newblock
\showISSN{1024-123X, 1563-5147}
\urldef\tempurl%
\url{https://doi.org/10.1155/2014/879736}
\showDOI{\tempurl}


\bibitem[Zhang et~al\mbox{.}(2020)]%
        {ZhangEtAl2020VELC}
\bibfield{author}{\bibinfo{person}{Chunkai Zhang}, \bibinfo{person}{Shaocong Li}, \bibinfo{person}{Hongye Zhang}, {and} \bibinfo{person}{Yingyang Chen}.} \bibinfo{year}{2020}\natexlab{}.
\newblock \showarticletitle{{{VELC}}: {{A New Variational AutoEncoder Based Model}} for {{Time Series Anomaly Detection}}}.
\newblock
\showeprint[arxiv]{1907.01702}~[cs, stat]
\urldef\tempurl%
\url{http://arxiv.org/abs/1907.01702}
\showURL{%
\tempurl}


\bibitem[Zhang et~al\mbox{.}(2019)]%
        {ZhangEtAl2019Deep}
\bibfield{author}{\bibinfo{person}{Chuxu Zhang}, \bibinfo{person}{Dongjin Song}, \bibinfo{person}{Yuncong Chen}, \bibinfo{person}{Xinyang Feng}, \bibinfo{person}{Cristian Lumezanu}, \bibinfo{person}{Wei Cheng}, \bibinfo{person}{Jingchao Ni}, \bibinfo{person}{Bo Zong}, \bibinfo{person}{Haifeng Chen}, {and} \bibinfo{person}{Nitesh~V. Chawla}.} \bibinfo{year}{2019}\natexlab{}.
\newblock \showarticletitle{A {{Deep Neural Network}} for {{Unsupervised Anomaly Detection}} and {{Diagnosis}} in {{Multivariate Time Series Data}}}. In \bibinfo{booktitle}{\emph{AAAI}}, Vol.~\bibinfo{volume}{33}. \bibinfo{pages}{1409--1416}.
\newblock
\urldef\tempurl%
\url{https://doi.org/10.1609/aaai.v33i01.33011409}
\showDOI{\tempurl}


\bibitem[Zhang et~al\mbox{.}(2007)]%
        {ZhangEtAl2007One}
\bibfield{author}{\bibinfo{person}{Rui Zhang}, \bibinfo{person}{Shaoyan Zhang}, \bibinfo{person}{Sethuraman Muthuraman}, {and} \bibinfo{person}{Jianmin Jiang}.} \bibinfo{year}{2007}\natexlab{}.
\newblock \showarticletitle{One Class Support Vector Machine for Anomaly Detection in the Communication Network Performance Data}. In \bibinfo{booktitle}{\emph{Proceedings of the {{Conference}} on {{Applied Electromagnetics}}, {{Wireless}} and {{Optical Communications}} ({{ELECTROSCIENCE}})}} \emph{(\bibinfo{series}{{{ELECTROSCIENCE}}'07})}. \bibinfo{publisher}{{World Scientific and Engineering Academy and Society (WSEAS)}}, \bibinfo{pages}{31--37}.
\newblock
\showISBNx{978-960-6766-25-1}


\bibitem[Zhao et~al\mbox{.}(2020)]%
        {ZhaoEtAl2020Multivariate}
\bibfield{author}{\bibinfo{person}{Hang Zhao}, \bibinfo{person}{Yujing Wang}, \bibinfo{person}{Juanyong Duan}, \bibinfo{person}{Congrui Huang}, \bibinfo{person}{Defu Cao}, \bibinfo{person}{Yunhai Tong}, \bibinfo{person}{Bixiong Xu}, \bibinfo{person}{Jing Bai}, \bibinfo{person}{Jie Tong}, {and} \bibinfo{person}{Qi Zhang}.} \bibinfo{year}{2020}\natexlab{}.
\newblock \showarticletitle{Multivariate time-series anomaly detection via graph attention network}. In \bibinfo{booktitle}{\emph{ICDM}}. IEEE, \bibinfo{pages}{841--850}.
\newblock


\bibitem[Zhu et~al\mbox{.}(2018)]%
        {ZhuEtAl2018Matrix}
\bibfield{author}{\bibinfo{person}{Yan Zhu}, \bibinfo{person}{Chin-Chia~Michael Yeh}, \bibinfo{person}{Zachary Zimmerman}, \bibinfo{person}{Kaveh Kamgar}, {and} \bibinfo{person}{Eamonn Keogh}.} \bibinfo{year}{2018}\natexlab{}.
\newblock \showarticletitle{Matrix profile XI: SCRIMP++: time series motif discovery at interactive speeds}. In \bibinfo{booktitle}{\emph{2018 IEEE International Conference on Data Mining (ICDM)}}. IEEE, \bibinfo{pages}{837--846}.
\newblock


\bibitem[Zhu et~al\mbox{.}(2016a)]%
        {ZhuEtAl2016Matrix}
\bibfield{author}{\bibinfo{person}{Yan Zhu}, \bibinfo{person}{Zachary Zimmerman}, \bibinfo{person}{Nader~Shakibay Senobari}, \bibinfo{person}{Chin-Chia~Michael Yeh}, \bibinfo{person}{Gareth Funning}, \bibinfo{person}{Abdullah Mueen}, \bibinfo{person}{Philip Brisk}, {and} \bibinfo{person}{Eamonn Keogh}.} \bibinfo{year}{2016}\natexlab{a}.
\newblock \showarticletitle{Matrix profile ii: Exploiting a novel algorithm and gpus to break the one hundred million barrier for time series motifs and joins}. In \bibinfo{booktitle}{\emph{2016 IEEE 16th international conference on data mining (ICDM)}}. IEEE, \bibinfo{pages}{739--748}.
\newblock


\bibitem[Zhu et~al\mbox{.}(2016b)]%
        {DBLP:conf/icdm/ZhuZSYFMBK16}
\bibfield{author}{\bibinfo{person}{Yan Zhu}, \bibinfo{person}{Zachary Zimmerman}, \bibinfo{person}{Nader~Shakibay Senobari}, \bibinfo{person}{Chin-Chia~Michael Yeh}, \bibinfo{person}{Gareth Funning}, \bibinfo{person}{Abdullah Mueen}, \bibinfo{person}{Philip Brisk}, {and} \bibinfo{person}{Eamonn Keogh}.} \bibinfo{year}{2016}\natexlab{b}.
\newblock \showarticletitle{Matrix profile ii: Exploiting a novel algorithm and gpus to break the one hundred million barrier for time series motifs and joins}. In \bibinfo{booktitle}{\emph{2016 IEEE 16th international conference on data mining (ICDM)}}. IEEE, \bibinfo{pages}{739--748}.
\newblock


\bibitem[Zimmerman et~al\mbox{.}(2019a)]%
        {ZimmermanEtAl2019Matrix}
\bibfield{author}{\bibinfo{person}{Zachary Zimmerman}, \bibinfo{person}{Kaveh Kamgar}, \bibinfo{person}{Nader~Shakibay Senobari}, \bibinfo{person}{Brian Crites}, \bibinfo{person}{Gareth Funning}, \bibinfo{person}{Philip Brisk}, {and} \bibinfo{person}{Eamonn Keogh}.} \bibinfo{year}{2019}\natexlab{a}.
\newblock \showarticletitle{Matrix profile XIV: scaling time series motif discovery with GPUs to break a quintillion pairwise comparisons a day and beyond}. In \bibinfo{booktitle}{\emph{Proceedings of the ACM Symposium on Cloud Computing}}. \bibinfo{pages}{74--86}.
\newblock


\bibitem[Zimmerman et~al\mbox{.}(2019b)]%
        {ZimmermanEtAl2019Matrixa}
\bibfield{author}{\bibinfo{person}{Zachary Zimmerman}, \bibinfo{person}{Nader~Shakibay Senobari}, \bibinfo{person}{Gareth Funning}, \bibinfo{person}{Evangelos Papalexakis}, \bibinfo{person}{Samet Oymak}, \bibinfo{person}{Philip Brisk}, {and} \bibinfo{person}{Eamonn Keogh}.} \bibinfo{year}{2019}\natexlab{b}.
\newblock \showarticletitle{Matrix profile XVIII: time series mining in the face of fast moving streams using a learned approximate matrix profile}. In \bibinfo{booktitle}{\emph{2019 IEEE International Conference on Data Mining (ICDM)}}. IEEE, \bibinfo{pages}{936--945}.
\newblock


\end{thebibliography}

\end{document}